%% file: main.tex
\documentclass{article}
\usepackage[affil-it]{authblk}
\usepackage[utf8]{inputenc}
\usepackage[T1]{fontenc}
\usepackage{color}
\usepackage[ruled,vlined]{algorithm2e}
\usepackage{amsmath,mathtools, amssymb,amsfonts,amsthm}
\usepackage{pifont}
\usepackage[inline]{enumitem}
\usepackage[style=numeric,maxcitenames=3,maxnames=8, backend=biber, doi=false,isbn=false,url=false,eprint=false,giveninits=true,sorting=none]{biblatex}
\setlist[itemize]{leftmargin=20mm}
\usepackage[skip=0.5ex]{subcaption}
\usepackage{graphicx}
\usepackage{url}
\usepackage{booktabs,tabularx}
\usepackage{siunitx}
\usepackage{multirow}
\usepackage{csquotes}
\usepackage[T1]{fontenc}
\usepackage{lipsum}
\usepackage{commath}
\usepackage{pgf,tikz}
\usepackage{pgfplots}
\DeclareUnicodeCharacter{0301}{\'{e}}
\usetikzlibrary{babel}
\usetikzlibrary[patterns]
\usetikzlibrary{arrows,positioning,shapes} 
\usetikzlibrary{intersections, pgfplots.fillbetween}
\usetikzlibrary{calc} 
\usetikzlibrary{shapes,arrows,positioning,shadows,trees}
\usetikzlibrary{decorations.pathreplacing}
\usetikzlibrary{positioning}
\usetikzlibrary{shapes.misc}
\usetikzlibrary{patterns}
\usetikzlibrary{arrows,positioning,shapes} 
\usetikzlibrary{external}
\usepackage{xcolor,colortbl}
\tikzsetexternalprefix{FIGURES_PDF/}
\newcommand{\cmark}{\ding{51}}%

\newcommand{\bb}[1]{\mathbb{#1}}
\pgfplotsset{grid style={solid,white!90!black}}
\theoremstyle{remark}
\newtheorem*{remark}{Remark}
\usepackage[pdftex,pdfpagelabels,bookmarks,hyperindex,hyperfigures]{hyperref}
\usepackage{cleveref}

\title{Design of borehole resistivity measurement acquisition systems using deep learning}

\author[1]{M. Shahriari}
\author[2]{A. Hazra}
\author[3,2,4]{ D. Pardo}
	
\affil[1]{\footnotesize Software Competence Center Hagenberg GmbH (SCCH), Hagenberg, Austria}
\affil[2]{\footnotesize Basque Center for Applied Mathematics, (BCAM), Bilbao, Spain}
\affil[3]{\footnotesize University of the Basque Country (UPV/EHU), Leioa, Spain}
\affil[4]{\footnotesize  Ikerbasque (Basque Foundation for Sciences), Bilbao, Spain}
\date{\today}

\begin{document}
\maketitle
\begin{abstract}
	Borehole resistivity measurements recorded with logging-while-drilling (LWD) instruments are widely used for characterizing the earth's subsurface properties. They facilitate the extraction of natural resources such as oil and gas. LWD instruments require real-time inversions of electromagnetic measurements to estimate the electrical properties of the earth's subsurface near the well and possibly correct the well trajectory.
	
	Deep Neural Network (DNN)-based methods are suitable for the rapid inversion of borehole resistivity measurements as they approximate the forward and inverse problem \textit{offline} during the training phase and they only require a fraction of a second for the {\em evaluation} (aka prediction). However, the inverse problem generally admits multiple solutions. DNNs with traditional loss functions based on data misfit are ill-equipped for solving an inverse problem. This can be partially overcome by adding regularization terms to a loss function specifically designed for \textit{encoder-decoder} architectures. But adding regularization seriously limits the number of possible solutions to a set of \textit{a priori} desirable physical solutions.  To avoid this, we use a \textit{two-step} loss function without any regularization \cite{Shahriari_loss}. 
	
	In addition, to guarantee an inverse solution, we need a carefully selected measurement acquisition system with a sufficient number of measurements. In this work, we propose a DNN-based iterative algorithm for designing such a measurement acquisition system. We illustrate our DNN-based iterative algorithm via several synthetic examples. Numerical results show that the obtained measurement acquisition system is sufficient to identify and characterize both resistive and conductive layers above and below the logging instrument. Numerical results are promising, although further improvements are required to make our method amenable for industrial purposes.

\end{abstract}

\section{Introduction}
One of the motivations for the geophysical exploration of the earth's subsurface is to facilitate the extraction of natural resources such as oil and gas. Other applications include geothermal exploration and CO$_2$-sequestration. To this end, there exist a wide variety of techniques intended to estimate different subsurface properties. Here, we focus on resistivity measurements \cite{samouelian2005electrical,spies1996electrical}, which intend to determine electrical resistivity (and sometimes dielectric permittivity) of subsurface materials.

We characterize resistivity measurements depending on the acquisition location as: a) \textit{on the surface}, which includes those acquired with controlled source electromagnetic (CSEM) \cite{Bakr,Constable, streich2016controlled, Key, Key1} and magnetotellurics (MT) \cite{Aramberri}; and b) \textit{in the borehole}, often recorded with logging while drilling (LWD) devices \cite{Davydycheva,Ijasan}. In this work, we focus on borehole resistivity measurements recorded with LWD instruments.

In LWD, a well logging tool conveys down into the well borehole, records electromagnetic measurements and transmits the data in real-time to evaluate the formation and subsequently adjust the inclination and azimuth of the well trajectory \cite{alyaev2019decision}. This strategy to determine a well trajectory is known as geosteering and it is crucial for maximizing the extraction of oil and gas.

The main challenge when dealing with geosteering is the need to interpret borehole resistivity measurements rapidly, i.e., we need to invert these measurements in real-time. Traditional inversion methods are often computationally expensive \cite{Tarantola,Puzyrev}. For example, gradient-based methods \cite{Vogel,Tarantola} require simulating the forward problem dozens of times per logging positions. These methods need to evaluate derivatives of measurements with respect to the inversion variables, which is often challenging and time-consuming \cite{Tarantola,Puzyrev}. An alternative to gradient-based techniques is to apply statistics-based methods \cite{Vogel,Tarantola,Watzenig}, which search for a global minima rather than a local one. However, these methods aggravate the problem of high computation time since these techniques often require a large number of forward simulations. Moreover, for each new set of measurements, i.e., for each logging position, one needs to repeat the entire inversion process, which could be computationally intensive. 

These difficulties can be partially overcome by using deep neural networks (DNNs), as shown in recent works \cite{Shahriari_deep_inverse,Shahriari_loss, peyret2019automatic}. While DNNs involve generating a large dataset for the training of the neural network, the biggest advantage of DNN methods is that they approximate the forward and inverse problem \textit{offline}. Once the training process is completed, {\emph{online}} prediction (evaluation) takes a fraction of a second. This makes DNN methods powerful candidates for real-time inversion \cite{Shahriari_deep_inverse, Shahriari_loss}. Unlike gradient-based and statistics-based methods, DNN methods do not require the solution of the forward problem after recording each new set of measurements.

 For two and three-dimensional earth subsurface models, we need to employ numerical techniques such as the finite element method (FEM) \cite{Shahriari, Bakr, Aramberri}, the finite volume (FV) \cite{novo2009three}, or the finite difference method (FDM) \cite{Davydycheva,Davydycheva1,lee2011numerical} to solve the forward problem. Thus, producing a dataset for DNNs may lead to prohibitively expensive computational costs. We can lower the costs by restricting to \num{1}D layered earth models \cite{Loseth}. In multiple realistic scenarios, \num{1}D earth models approximate the geological earth formation reasonably well, and it is often used for borehole resistivity inversion in the oil and gas industry. This technique takes small simulation time (typically, a fraction of a second) due to the availability of a semi-analytical method \cite{Pardo}. In this work, we restrict to 1D earth models for the sake of computational efficiency, although the method proposed here can be easily extended to 2D and 3D, provided sufficient computational resources are available.

 As illustrated in \cite{Shahriari_loss}, DNNs equipped with a traditional loss function based on the data misfit exhibit serious limitations when predicting some physically realistic solutions to the inverse problems. Such limitations occur because inverse problems often admit multiple solutions. The performance of a DNN for solving an inverse problem could be drastically improved with a loss function specifically designed for \textit{encoder-decoder} architectures or a \textit{two-step} loss function \cite{Shahriari_loss}. Regularization terms added to the loss function can guide the solution towards an {\em a priori} \enquote{desirable} physical solution; however, they often hide the fact that other \enquote{physical} solutions may also co-exist \cite{Shahriari_loss}. 

Herein, we consider the two-step loss function based on the measurement misfit described in \cite{Shahriari_loss} without regularization terms. The main advantage of this choice is that we can recover all possible solutions of the inverse problem (in the sense that they satisfy the measurements), and not simply those dictated by the regularization term. This opens the door to analyze how many inverse solutions satisfy all measurements of a given measurement acquisition system. Alternatively, we can enrich the measurement acquisition system until guaranteeing that the inverse solution is unique. This is the approach we take in this work.

The main goal of the present work is to design a measurement acquisition system that guarantees a solution of practical interest. For that purpose, we propose a DNN-based iterative method. We start by considering a large set of borehole logging resistivity measurements from which we first select a single one. Then, we iteratively add a new measurement such that it minimizes the number of existing inverse solutions. To select the new measurement in each step of the iterative algorithm, we employ the \textit{coefficient of determination} (also known as $R^2$ or $R$-squared statistic) \cite{ devore2011probability}. 

In the following, we describe a DNN-based algorithm for designing a measurement acquisition system, and we illustrate its performance with several synthetic examples, analyzing its benefits and limitations.
Improvement of some technical aspects such as optimal data sampling techniques and selection of DNN architecture \cite{Moghadas, Puzyrev} can make the DNN-based inversion method more robust. However, for simplicity, we do not analyze those aspects here. We neither consider noisy measurements, which will be the subject of future work. 

The remainder of this paper is organized as follows. \Cref{sec:def} 
provides an abstract formulation of the problem. \Cref{sec:deeplearningalgo} describes the deep learning algorithm used in this work.  \Cref{sec:database} details the measuring devices, measured components, and data space. \Cref{sec: measacquisys} discusses the algorithm for designing the measurement acquisition system. The numerical implementation is discussed in \Cref{sec:impl}. \Cref{sec:numeric} demonstrates the applicability of our method with some examples. \Cref{sec:discon} is devoted to conclusions and a brief discussion including possible future work.

\section{Problem formulation}\label{sec:def}

\begin{figure}[!h]
	\centering
\begin{tikzpicture}
\node[draw, rectangle, black, fill= orange!20](n1) at (1,-2) {Subsurface properties ${\bf p}$};
\node[draw, rectangle, black, fill= orange!20, below of = n1, node distance = 1 cm] (n2){Measurement system ${\bf s}$};
\node[draw, rectangle, black, fill= orange!20, below of = n2, node distance = 1 cm] (n3){Well trajectory ${\bf t}$};
\node[draw, circle, black , right of = n2, node distance = 4 cm](n4){+};
\node[draw, rectangle, black, fill= green!20, right of = n4, node distance = 4 cm](n5){Measurements ${\bf m}$};
\foreach \a in {n1,  n2, n3}
{\path[-] (\a.east) edge (n4);}
\draw[->, thick] (n4) --(n5) node[midway, above] {$\mathcal{F}$};

\node[draw, rectangle, black, fill= orange!20](n1) at (1,-5) {Measurements ${\bf m}$};
\node[draw, rectangle, black, fill= orange!20, below of = n1, node distance = 1 cm] (n2){Measurement system ${\bf s}$};
\node[draw, rectangle, black, fill= orange!20, below of = n2, node distance = 1 cm] (n3){Well trajectory ${\bf t}$};
\node[draw, circle, black , right of = n2, node distance = 4 cm](n4){+};
\node[draw, rectangle, black, fill= green!20, right of = n4, node distance = 4 cm](n5){Subsurface properties ${\bf p}$};
\foreach \a in {n1,  n2, n3}
{\path[-] (\a.east) edge (n4);}
\draw[->, thick] (n4) --(n5) node[midway, above] {$\mathcal{I}$};
\end{tikzpicture}
	\caption{Schematic diagram of a forward and an inverse problem.}
	\label{fig:problem_types}
\end{figure}
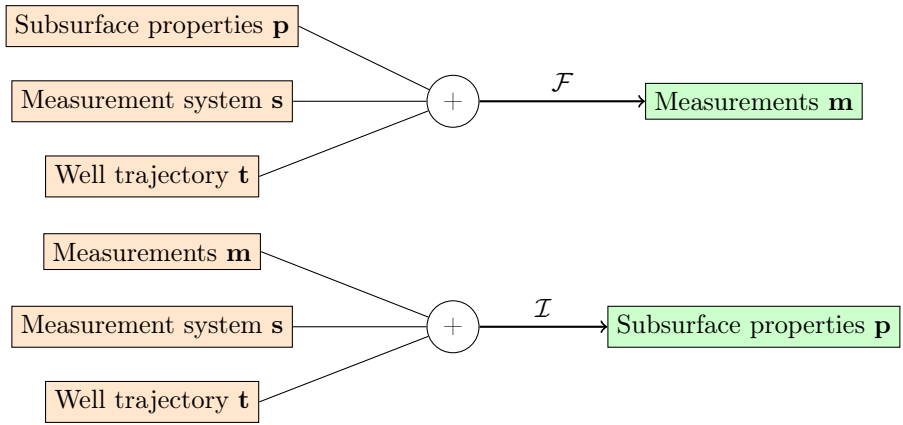

We introduce two different mathematical problems associated with borehole resistivity measurements: forward and inverse (see \Cref{fig:problem_types}).

\paragraph{Forward problem.}
Given an earth model $\mathbf p \in \bb{R}^{n_p}$,  a well trajectory $\mathbf t \in \bb{R}^{n_t}$ and a measurement acquisition system $\mathbf s \in \bb{R}^{n_s}$, the forward problem $\cal F$ delivers the corresponding measurements $\mathbf m({\mathbf s}) \in \bb{R}^{n_m}$; the variables are parameterized by real-valued vectors with $n_{(\cdot)} $ dimensions.
 Mathematically, we have:

\begin{align}
 {\mathcal F}({\mathbf s},{\mathbf t},{\mathbf p})&=  {\mathbf m}({\mathbf s}) \quad \text{(Forward)}.
\end{align}
The forward operator $\cal F$ is governed by partial differential equations (PDEs), in this case, Maxwell's equations in the frequency domain with the corresponding boundary conditions governing the electromagnetic wave propagation phenomena \cite{Shahriari}.

\paragraph{Inverse problem.}
Given a set of measurements $\mathbf m$ obtained with a specific measurement acquisition system $\mathbf s$ and logging trajectory $\mathbf t$, the inverse operator $\cal{I}$ determines
a \textit{plausible} earth model $\mathbf p$, in the sense that it satisfies the recorded measurements. The earth model is characterized by a material subsurface distribution. We have:

\begin{align}
 \mathcal I({\mathbf s},{\mathbf t},{\mathbf m}({\mathbf s}))&= {\mathbf p}, \quad \text{(Inverse)}
\end{align}


An inverse problem often exhibits multiple solutions \cite{Tarantola,Vogel} since different earth models may satisfy the recorded measurements.

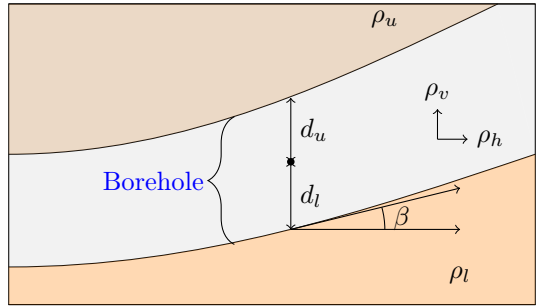
\begin{figure}[ht]
	\centering
	\begin{tikzpicture}[scale=1.0]
	\draw (0,0) rectangle (7,4);
	\draw[name path=A1] (0,0) -- (7,0);
	\draw[name path=A2] (0,4) -- (7,4);
	\draw[ name path=line1] (0,2) cos (6.5,4);
	\draw[ name path=line2] (0,0.5) cos (7,2);
	\fill[fill=gray,  opacity=0.1] (6.5,4) -- (7,4) -- (7,2) --(6.5,4);
	\tikzfillbetween[of=line2 and A1]{orange, opacity=0.3};
	\tikzfillbetween[of= line2 and line1]{gray, opacity=0.1};
	\tikzfillbetween[of= line1 and A2]{brown, opacity=0.3};
	\draw [decorate,decoration={brace,amplitude=10pt}]
	(3,0.8) -- (3,2.5) node [blue,midway,left,xshift=-8pt] {Borehole};
	\fill[black] (3.75, 1.9) circle (0.05cm);
	\draw[<->] (3.75, 1.9) -- (3.75,2.75) node [midway, right] {$d_u$};
	\draw[<->] (3.75,1.9) -- (3.75,1.0) node [midway, right] {$d_l$};
	\draw[->] (3.75,1.0) -- (6,1.0);
	\draw[->] (3.75,1.0) -- (6,1.55);
	\draw (5,1.0) arc (0:15.5:1.1 cm) node[midway, right] {$\beta$};
	\draw [->] (5.7,2.2) -- (6.1,2.2) node [right] {$\rho_h$};
	\draw [->] (5.7,2.2) -- (5.7,2.6) node [above] {$\rho_v$};
	\node (rho_u) at (5,3.8) {$\rho_{u}$};
	\node (rho_l) at (6,0.4) {$\rho_{l}$};
	\end{tikzpicture}
	\caption{1D media and a trajectory. The black circle indicates the current position of the midpoint of the logging instrument.}
	\label{fig:1D}
\end{figure}

\paragraph{Earth parametrization.} We restrict to a \num{1}D layered model of the earth. This assumption is common for geosteering applications \cite{Key,Ijasan} since it allows to simulate a problem in a fraction of a second using a semi-analytic method \cite{Loseth,Shahriari}. We further assume the earth model to be a three-layer medium \cite{Pardo}, as illustrated in \Cref{fig:1D}. This medium is generally characterized by seven parameters, namely: 

\begin{enumerate}[label=(\alph*)]
	\item Horizontal resistivity $\rho_h$ of the layer containing the mid-point of the logging instrument.  
	\item Vertical resistivity $\rho_v$ of of the layer containing the mid-point of the logging instrument. Often, instead of two different resistivities,  $\rho_v$ is replaced by anisotropy factor $a$, i.e. $\frac{\rho_v}{\rho_h}$. 
	\item The resistivity of the layer located below the current layer $\rho_l$.
	\item The resistivity of the layer located above the current layer $\rho_u$.
	\item Vertical distance from the current logging position to the upper bed boundary $d_u$.
	\item Vertical distance from the current logging position to the lower bed boundary $d_l$.
	\item The dip angle ($\beta$) of the formation.
\end{enumerate}

 In this work, for simplicity, we assume earth model $E_5 := \{ \rho_v,~\rho_l,~\rho_u,~d_u,~d_l\}$, i.e., we assume isotropic formations $\rho_h=\rho_v$ and formation dip angle $\beta=0$.
 However, our algorithm for designing measurement acquisition systems can be applied to other earth parameterizations.

\section{Deep learning  loss function}\label{sec:deeplearningalgo}
We briefly describe here the deep learning approach adopted for this work. The interested reader can find a more detailed discussion on other deep learning algorithms in books (e.g \cite{goodfellow2016deep}) and review articles ( e.g. \cite{Higham}).

Our target is to approximate the inverse operator $\mathcal{I}:~ \bb{R}^{n_s} \times \bb{R}^{n_t} \times \bb{R}^{n_m}  \mapsto \bb{R}^{n_p} $ using a DNN. 
After the model architecture is created, we optimize a \textit{loss function} to train our DNN. 
The simplest loss function is based on data misfit and optimized in $l_1$ or $l_2$ norm, given by:
\begin{align}
\label{eq:lossmisfit}
\mathcal{I}_{\alpha^\ast}&:=\arg \min_{\alpha } \enVert{\mathcal{I}_{\alpha}(\mathbf s,\mathbf t,\mathbf m)-\mathbf p},
\end{align}
where $\alpha$ is the set of weights associated to the DNN.

However, due to the non-uniqueness of the solution in inverse problems, this loss function produces an average of various plausible solutions. Hence, it may produce incorrect results in the sense that the forward simulation of the average of several solutions may not satisfy the measurements \cite{Shahriari_loss}. 

We can overcome this limitation by optimizing a loss function based on the misfit of measurements \cite{jin2019using}, given by:
\begin{align}
\label{eq:lossmisfitm}
\mathcal{I}_{\alpha^\ast}:=\arg \min_{\alpha} \enVert{(\mathcal{F}\circ\mathcal{I}_{\alpha})(\mathbf s,\mathbf t,\mathbf m)-\mathbf m},
\end{align}
where $\alpha$ is the set of weights associated to the DNN.

Although the above \textit{loss function} produces accurate results, it has some critical limitations \cite{Shahriari_loss}. First, the requirement of evaluating the forward problem multiple times during training may result in a prohibitively high computational time. Second, although sophisticated and GPU-compatible libraries like \textit{Tensorflow} \cite{abadi2016tensorflow} are available for training of a DNN, the forward problem is generally solved in a CPU. This produces a communication bottleneck between CPU and GPU. We can remove this bottleneck by making the forward solver GPU-compatible, but this involves additional implementation challenges.

A suitable alternative is to solve the following optimization problem \cite{Shahriari_loss}:
\begin{align}
\label{eq:lossencdec}
 ({\cal F}_{\alpha^\ast} ,\mathcal{I}_{\beta^\ast}):=\arg \min_{\alpha,\beta}\big( &\enVert{(\mathcal{F}_\alpha\circ\mathcal{I}_{\beta})(\mathbf s,\mathbf t,\mathbf m)-\mathbf m} + \enVert{\mathcal{F}_{\alpha}(\mathbf s,\mathbf t,\mathbf p)-\mathbf m} \big),
\end{align}
where $\alpha$ and $\beta$ are the set of weights associated to the DNNs which we use to approximate the forward function and inverse operator respectively.

The above loss function consists of two terms. The first one constitutes an encoder-decoder architecture \cite{Badrinarayanan} and ensures that $\mathcal{I}_{\beta}$ is an inverse of $\mathcal{F}_\alpha$. The second one guarantees that the forward DNN approximates the measurements.

We can also split the previous minimization problem into two steps \cite{Shahriari_loss}.
In the first step, we approximate the forward operator:
\begin{align}
   {\cal F}_{\alpha^\ast} (\mathbf{s}) := \arg \min_{\alpha}  \enVert{{\cal F}_\alpha (\mathbf{s}, \mathbf{t},\mathbf{p}) - \mathbf{m}({\mathbf s})}.
       \label{eq:1twostep}
\end{align}
Next, we determine the inverse operator:
\begin{align}
   {\cal I}_{\beta^\ast} (\mathbf{s}):= \arg \min_{\beta} \enVert{ ({\cal F}_{\alpha^\ast} \circ {\cal I}_\beta) (\mathbf{s}, \mathbf{t},\mathbf{m}({\mathbf s})) - \mathbf{m}({\mathbf s})}.
    \label{eq:2twostep}
\end{align}
We use this two-step approach for our deep learning algorithm.

\begin{remark}
	In the above expressions, $\enVert{\cdot}$ denotes some suitable matrix norm. For details, see \cite{Shahriari_loss}.
\end{remark}
\section{Databases with multiple measurements} \label{sec:database}

\subsection{Borehole resistivity measurements} \label{subsec:meas}
\paragraph{Logging instrument.}
Modern borehole resistivity instruments can measure all possible (nine) couplings  of the magnetic field $H_{ij}$, where $i$  indicates the orientation of the transmitter and $j$ stands for the receiver orientation. We obtain different measurements combining one or more $H_{ij}$. The measurements further depend on the number and locations of the transmitters and receivers along the logging instrument. We consider two different types of logging instruments: 
\begin{enumerate}
	\item \textit{Conventional LWD}. We consider an instrument that incorporates a pair of receivers and three pairs of transmitters, as illustrated in \Cref{fig:lwd}. We refer to them as short-spaced, medium-spaced, and long-spaced depending on the distance to the receiver. In this case, each measurement is specified by a pair of transmitters, a specific operator frequency, and the pair of receivers (\Cref{tab:toolsofmeas}). 
	\begin{figure}[ht]
		\centering
		\begin{tikzpicture}[scale=1.0]
		\node (layers) at (0,0)[scale=0.80]{
			\input{Figures/LWD_schem.tex} 
		};
		\end{tikzpicture}
		\caption{Conventional LWD. \textbf{Tx\textsubscript{i,j}}, \textbf{j}=1,2 denote the transmitters and \textbf{i}=1,2,3 specifies the spacing between the  transmitters. \textbf{Rx\textsubscript{1}}, \textbf{Rx\textsubscript{2}} stands for the two receivers.}
		\label{fig:lwd}
	\end{figure}
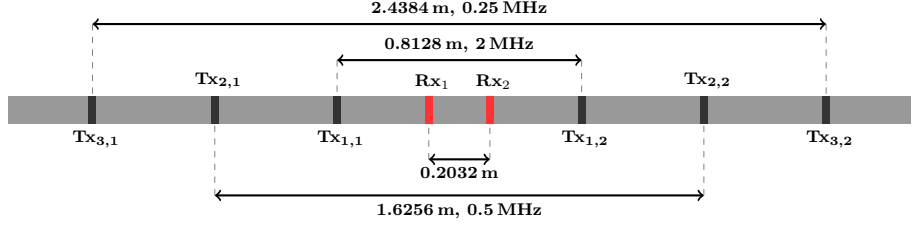
	\item \textit{A deep azimuthal instrument}. We consider an instrument that incorporates one transmitter that can operate at two different frequencies and two associated receivers, as depicted in \Cref{fig:deepazim}. Using a deep azimuthal instrument, each measurement is specified by the transmitter with an operating frequency and the corresponding receiver (\Cref{tab:toolsofmeas}).
	\begin{figure}[ht]
		\centering
		\begin{tikzpicture}[scale=1.0]
		\node (layers) at (0,0)[scale=0.80]{
			\input{Figures/deep_azim_schem.tex} 
		};
		\end{tikzpicture}
		\caption{Deep azimuthal instrument. It has one transmitter \textbf{Tx} operating at two different frequencies and two receivers \textbf{Rx\textsubscript{1}}, \textbf{Rx\textsubscript{2}}.}
		\label{fig:deepazim}
	\end{figure}
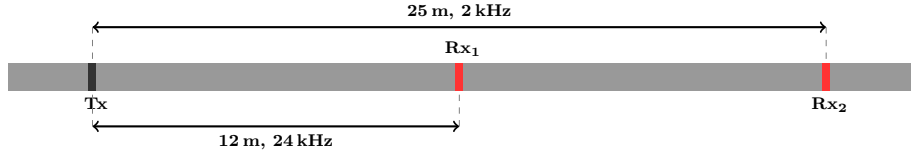
	\begin{table}
		\begin{center}
			\begin{tabular}{l|l|l|l|l|l|r}
				\toprule
				Measurement Systems & $n_T$ & $n_R$ & Spacing & $p_T~[\si{\m}]$& $p_R~[\si{\m}]$ & $\nu~[\si{\kHz}]$ \\
				\midrule
				\multirow{3}{*}{Conventional LWD} & \multirow{3}{*}{\num{6}} &\multirow{3}{*}{\num{2}} & short & \num{\pm 0.4064} & \num{\pm 0.1016} & 2000\\
				&  & & medium &\num{\pm 0.8128} & \num{\pm 0.1016} & 500\\
				&  & & long & \num{\pm 1.2192} &\num{\pm 0.1016}  & 250\\
				\midrule
				\multirow{2}{*}{Deep azimuthal instrument} & \multirow{2}{*}{\num{1}} &\multirow{2}{*}{\num{2}} & short & \num{-12.0} & \num{0.0} & \num{24}\\
				&&& long & \num{-12.0} & \num{13.0} & \num{2}\\
				\bottomrule
			\end{tabular}
		\end{center}
		\caption{ Different measurement systems and their configurations. We list here the number of transmitters $n_T$ and receivers $n_R$, the positions of the transmitters ($p_T$) and the receivers ($p_R$), and the frequency of the emitted pulse ($\nu$).}
		\label{tab:toolsofmeas}
	\end{table}
\end{enumerate}

\paragraph{Measured component.} 
We record certain components of the electromagnetic field with different instruments, as listed in \Cref{tab:modesofmeas}. Measurement logs vary depending on the number of receivers and transmitters \cite{Shahriari_deep_inverse}. 

Let $m_c$ be a measured component. When using a conventional LWD instrument, we record $m_c$ at both receivers (and denote it as $m_c^1$ and $m_c^2$), and postprocess them to compute the attenuation and phase difference as follows:

\begin{align*}
\ln \frac{m^1_c}{m^2_c} = \underbrace{\ln \frac{\envert{m^1_c}}{\envert{m^2_c}}}_{\times 20 \log(e) =: \text{attenuation }(dB)}+i \underbrace{\left( ph({m}^1_{c})-ph({m}^2_{c})\right)}_{\times \dfrac{180}{\pi} =:\text{phase difference (degree)}},
\end{align*}
where $ph(\cdot)$ and $\envert{\cdot}$ are the modulus and phase of a complex number. For an azimuthal instrument, $m_c^2 = 1$.

We measure attenuation and phase difference for the two logging instruments listed in \Cref{tab:toolsofmeas} in combination with every measurement listed in \Cref{tab:modesofmeas}. In total, this amounts to \num{45} different cases. In this work, we propose an algorithm to select a few of these measurements for inversion. 
\begin{table}
	\begin{center}
		\begin{tabular}{l|l|l|l}
			\toprule
			Name & Measured Component & LWD & Deep Azimuthal\\
			\midrule
			zz & $H_{zz}$ & \cmark & \cmark\\
			xx & $H_{xx}$ & \cmark & \cmark\\
			yy & $H_{yy}$ & \cmark & \cmark\\
			xxyyzz+ & $H_{xx}+H_{yy}+H_{zz}$ & \cmark & \cmark\\
			Geosignal & $\displaystyle\frac{H_{zz}-H_{zx}}{H_{zz}+H_{zx}}$ & \cmark & \cmark \\
			Symmetrized directional & $\displaystyle\frac{H_{zz}+H_{zx}}{H_{zz}-H_{zx}}\cdot \frac{H_{zz}-H_{xz}}{H_{zz}+H_{xz}}$ & \cmark & \cmark \\
			Antisymmetrized directional & $\displaystyle\frac{H_{zz}+H_{zx}}{H_{zz}-H_{zx}}\cdot \frac{H_{zz}+H_{xz}}{H_{zz}-H_{xz}}$ & \cmark & \cmark \\
			Harmonic resistivity & $\displaystyle\frac{H_{xx}+H_{yy}}{2~H_{zz}}$ & \cmark & \cmark\\
			Harmonic anisotropy & $\displaystyle\frac{H_{xx}}{H_{yy}}$ & \cmark & \cmark\\
			\bottomrule
		\end{tabular}
	\end{center}
	\caption{ Different measurement components and their definitions. For all $H_{ij}$, $i$ and $j$ indicate the orientations of transmitters and receivers, respectively.}
	\label{tab:modesofmeas}
\end{table}
\subsection{Databases for the earth subsurface properties}
\paragraph{Data normalization.} Data normalization is generally applied as part of data preparation for deep learning. The purpose is to rescale the numerical values in the dataset within a common or at least comparable range. We rescale the relevant geophysical variables into two categories using either a linear or a log-linear rescaling.
In both cases, we rescale the values of the variables to the interval $\intcc{0.5,1.5}$. In the log-linear case, the linear rescaling is preceded by taking the logarithm of the given variable (see \cite{Shahriari_loss} for details).  \Cref{tab:type_of_variables} describes the variables, their domain, and corresponding rescaling. 

\begin{table}[!htp]
	\centering
	\begin{tabular}{l|crl}
		\toprule
		Geophysical Variables   & Category  & Domain & Rescaling    \\
		\midrule
		Angles, attenuations,  & linear & $\mathbb{R}^n$  & ${\cal R}_{lin}({\bf x})$    \\
		phases, and geosignals & &    &   \\
		\midrule
		Apparent resistivities, & log-linear  & $(a,\infty)^n$   & ${\cal R}_{lin}({\cal R}_{\ln}({\bf x}))$   \\
		resistivities, and distances &    &   $a>0$ &    \\
		\bottomrule
	\end{tabular}
	\caption{Two categories of geophysical variables: we apply a different normalization to each of them.}
	\label{tab:type_of_variables}
\end{table}

\paragraph{Data space.}
We consider a piecewise 1D layered model of the earth, as depicted in \Cref{fig:1D}. Moreover, we assume isotropic formations and zero dip angle $\beta$ of the formation. After that, we select  random samples of material properties in the logarithmic scale within the specified ranges mentioned in \Cref{tab:matprop} and generate a dataset of \num{300000} samples.
\begin{table}[ht!]
	\begin{center}
		\begin{tabular}{lSS}
			\toprule
			Material properties & {Range} & {Log(Range)} \\
			\midrule
			$\rho_u$ & \SIrange{1}{e3}{\ohm \m} & \SIrange{0}{3}{} \\
			$\rho_l$ &   \SIrange{1}{e3}{\ohm \m}&\SIrange{0}{3}{} \\
			$\rho_v$ &  \SIrange{1}{e3}{\ohm \m}&\SIrange{0}{3}{} \\
			$d_u$ & \SIrange{e-2}{10}{\m} & \SIrange{-2}{1}{} \\
			$d_l$ & \SIrange{e-2}{10}{\m} & \SIrange{-2}{1}{} \\
			\bottomrule
		\end{tabular}
	\end{center}
	\caption{Relevant material properties of earth subsurface and their range of values.}
	\label{tab:matprop}
\end{table}

\paragraph{Dataset.}
We select \SI{80}{\percent} of the data for training. The remaining data are divided equally for validation and test.
We formally define the training, validation, and test datasets:
\begin{align*}
D_{\textnormal{train}} &:=\{(\mathbf{\tilde{p}}_i,\mathbf{\tilde{m}}_i({\mathbf s}),\mathbf{\tilde{t}}_i)\}_{i=1}^{n_{\textnormal{train}}},\\
D_{\textnormal{test}} &:=\{(\mathbf{\tilde p}_i,\mathbf{\tilde m}_i({\mathbf {s}}),\mathbf{\tilde t}_i)\}_{i=n_{\textnormal{train}}+1}^{n_{\textnormal{train}}+n_{\textnormal{test}}},\\
D_{\textnormal{validation}} &:=\{(\mathbf{\tilde p}_i,\mathbf{\tilde m}_i({\mathbf {s}}),\mathbf{\tilde t}_i)\}_{i=n_{\textnormal{train}}+n_{\textnormal{test}}+1}^{n_{\textnormal{train}}+n_{\textnormal{test}}+n_{\textnormal{validation}}},
\end{align*}
where $n_{\textnormal{train}}$, $n_{\textnormal{validation}}$, $n_{\textnormal{test}}$ are the number of training, validation, and test samples, respectively.

Hence, using \Cref{eq:1twostep,eq:2twostep}, the forward function and inverse operator are defined as the minimizers of the following problems:
\begin{align}
{\cal F}_{\alpha^\ast} (\mathbf{s}) := \arg \min_{\alpha}  \sum_{i=1}^{n_{\textnormal{train}}} \envert {{\cal F}_\alpha (\mathbf{\tilde s}, \mathbf{\tilde{t}}_i,\mathbf{\tilde{p}}_i) - \mathbf{\tilde{m}}_i({\mathbf s})},\\
{\cal I}_{\beta^\ast} (\mathbf{s}):= \arg \min_{\beta} \sum_{i=1}^{n_{\textnormal{train}}} \envert{{\cal F}_{\alpha^\ast} \circ {\cal I}_\beta (\mathbf{\tilde{s}}, \mathbf{\tilde{t}}_i,\mathbf{\tilde{m}}_i({\mathbf s})) - \mathbf{\tilde{m}}_i({\mathbf s})}.
\end{align}

\begin{remark}
	In the above expressions, $\envert{\cdot}$ denotes some suitable vector norms. For details, see \cite{Shahriari_loss}.
\end{remark}

\section{Algorithm for designing the measurement acquisition system} \label{sec: measacquisys}
We define a measurement acquisition system $\mathbf{s}:=\{{s}_i\}_{i=1}^{n}$ as a set of measurements recorded by a logging instrument. Each measurement $s_i$ records a quantity that corresponds to a specific component of the electromagnetic fields and a particular mode of operation of a logging instrument.
When the system is properly trained, we have:
\begin{equation}
{\cal F}({\bf s},{\bf t},{\cal I}_{\beta^\ast}({\bf s})) \approx {\cal F}({\bf s},{\bf t},{\cal I}({\bf s})) =  {\bf m} ({\bf s}).
\label{forward_condition}
\end{equation}
However, the correlation of inverse operator $\cal{I}(\bf s)$ and its approximation ${\cal I}_{\beta^\ast}({\bf s})$ may stay below an optimal threshold value, which implies that our DNN provides a different output of the inverse operator than the actual one.



In our iterative algorithm, we denote the measurement acquisition system at $i$-th iteration by $\mathbf{s}^i$. We start with a single measurement $s_1$. Therefore, our initial measurement acquisition system 
$\mathbf{s}^0=\{s_1\}$. In our first iteration, we train a DNN based on this measurement and obtain an estimation of ${\cal I}_{\beta^*}$. We form the set ${\bf s}^{ \textnormal{all}}=\{s_i\}_{i=1}^{n_{s}}$ with all the available measurements and iterate over every measurement from ${\bf s}^{ \textnormal{all}}$ to calculate how accurately the model fits the data when applied to the validation dataset $D_{\textnormal{validation}}$. We measure this using $R^2$ statistics, defined as \cite{devore2011probability} : 
	\begin{align}
	R^2 ({\mathbf s}^0, s_i)&:= 1-\frac{SS_{\textnormal{res}}}{SS_{\textnormal{tot}}},
	\end{align}
	where
	\begin{equation}
	\begin{split}
	SS_{\textnormal{res}}({\bf s}^0, s_i)&:= \sum^{n_{\textnormal{validation}}}_{j=1} \left({\cal F}(s_i,t_j,p_j)-{\cal F}(s^i,t_j,{\cal I}_{\beta^\ast}({\bf s}^{0},t_j, {\bf m}_j ({\bf s}^{0})))\right)^2 \\
	SS_{\textnormal{tot}}({\bf s}^0, s_i)&:= \sum^{n_{\textnormal{validation}}}_{j=1} \left({\cal F}(s_i,t_j,p_j)-\overline{{\cal F}(s_i,t_j, p_j)}\right)^2\\
	\overline{{\cal F}(s_i,t_j,p_j)} &:= \frac{1}{n_{\textnormal{validation}}}\sum^{n_{\textnormal{validation}}}_{j=1} {\cal F}(s_i,t_j,p_j)
	\end{split}
	\label{eq:r2stat}
	\end{equation}
After computing ${\cal I}_{\beta^*}$, we calculate $R^2 ({\bf s}^0, s_i) $ for every $s_i \in \bf s^{\textnormal{all}}$ using (\ref{eq:r2stat}). We select the worst correlated measurement $i^*:=\arg \min_i R^2 ({\bf s}^0, s_i)$ and form a new set of measurement ${\bf s}^1 := {\bf s}^0 \cup s_{i^\ast}$. Iterating this procedure, we obtain \Cref{algo:adaptive}. We stop the algorithm when the $R^2$ coefficient of $i^*$ is above a certain threshold level $r_a$ (in our case, \num{0.8}), which indicates that all measurements are properly approximated.

\begin{figure}[ht!]
	\centering 
	\includegraphics[width=0.8\textwidth] {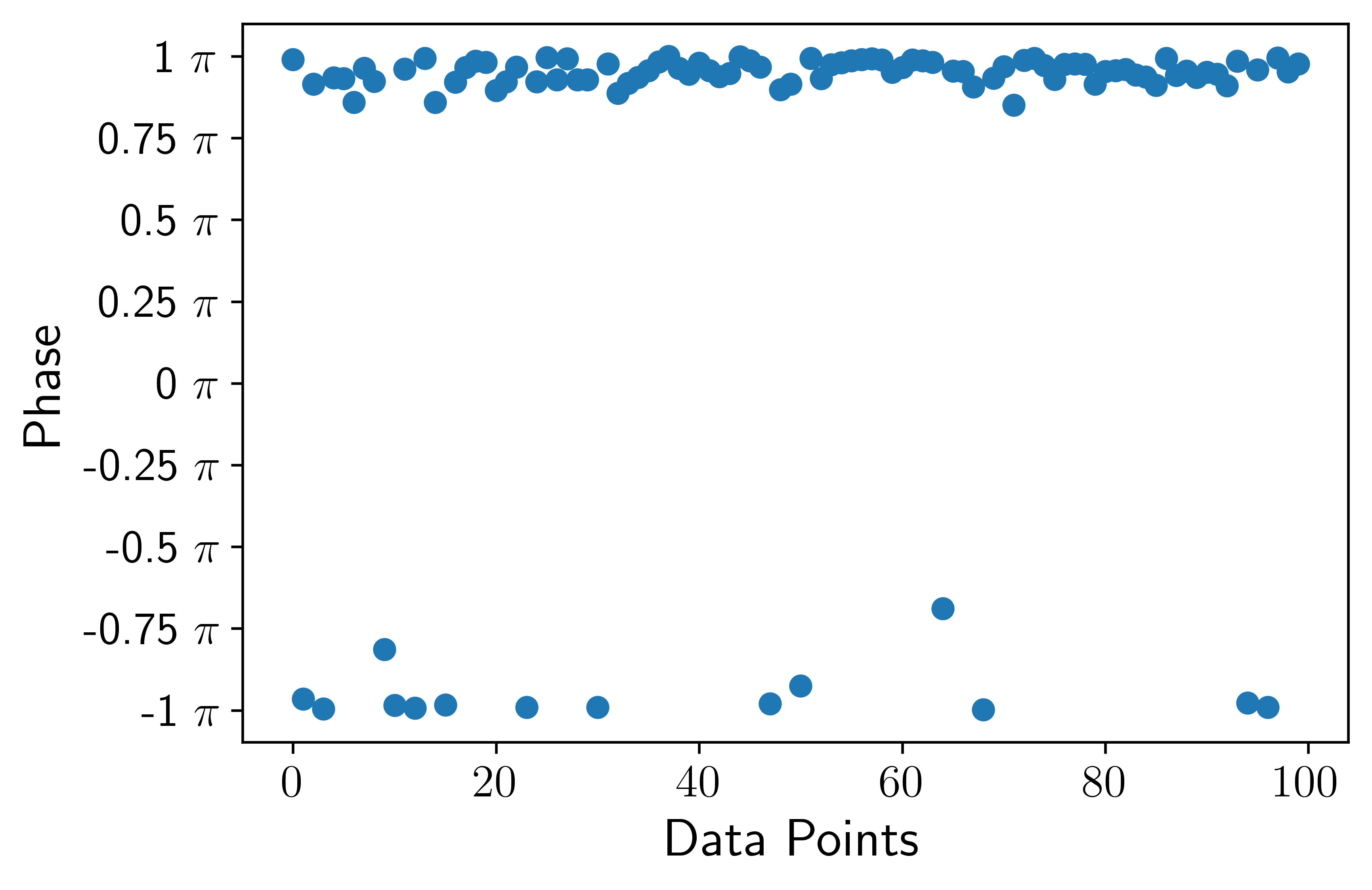}
	\caption{Phase discontinuity. Phase difference can produce an artificial discontinuity. }
	\label{fig:phaseprob}
\end{figure}

\begin{remark}
	We generally take an average of $R^2$ values associated with both attenuation and phase difference. However, some measurements produce phase differences close to $-\pi$ or $\pi$. As any phase $\phi$ is equivalent to a phase value $\phi+2\pi$, the phase can exhibit an artificial discontinuity. For illustration, we select \num{100} samples of a specific measurement -- $yy$ component with the short-spaced deep azimuthal instrument-- and plot the phase difference in \Cref{fig:phaseprob}. As the phase at the bottom is possibly close to a point near the top of the figure, a DNN approximation of the phase difference of this dataset would result in an erroneous approximation. We identified such measurements and ignore the $R^2$ values of phase data associated with them. In those cases, we only consider $R^2$ associated with attenuation data.
\end{remark}

\begin{algorithm}[H]
\SetAlgoLined
\KwIn{${\bf s}^{\textnormal{all}}=\{s_i\}_{i=1}^{n_{s}}$}
\KwOut{${\bf{s}}^k,~{\cal I}_{\beta^\ast}$}
 Select ${\bf s}^0$\;
 Train a DNN to obtain ${\cal I}_{\beta^\ast}$ using ${\bf s}^0$\;
 \While{$1<=k<=n_{s-1}$}{
   $i^\ast := \arg \min_i R^2 ({\bf s}^{k-1}, s_i)$\;
 \eIf{$R^2 ({\bf s}^{k-1}, {s}_{i^\ast}) < r_a$}{
 ${\bf s}^k={\bf s}^{k-1} \cup {s}_{i^\ast}$\; 
  Train a DNN to obtain ${\cal I}_{\beta^\ast}$ using ${\bf s}^k$\;
 }{
  exit\;
 }
 $k=k+1$;
 }
 \caption{Selection of measurement acquisition system.}
 \label{algo:adaptive}
\end{algorithm}

\section{Implementation} \label{sec:impl}
We solve our forward problem employing a semi-analytic method \cite{Loseth,Pardo}. The method is implemented in an {\em{in-house}} Fortran 90 code \cite{Shahriari}. Using the earth model $E_5$ and \num{45} different measurements, we produce a large dataset (\textit{ground truth}) of \num{300000} samples. Computations take almost two days to produce such dataset using a personal computer.

We use Tensorflow 2.0 \cite{tf2,abadi2016tensorflow} and Keras \cite{chollet2015} libraries to build our DNN model architecture for training. We employ the two-step loss function described in \Cref{eq:1twostep,eq:2twostep}. First, we train a DNN to produce the forward approximate ${\cal F}_{\alpha^\ast}$ following \Cref{eq:1twostep}, and then use the trained DNN and \Cref{eq:2twostep} to produce the approximate inverse operator ${\cal I}_{\beta^\ast}$. We employ a simpler DNN architecture to approximate ${\cal F}$ than the one to approximate ${\cal I}$ since the forward function $\cal F$ is well-posed and continuous, while the inverse operator $\cal{I}$ is not even well-defined.  The full architecture is described in \cite{Shahriari_loss}. We use an NVIDIA Quadro GV100 GPU for training.

To reduce the computational cost of \Cref{algo:adaptive}, we first execute it using \num{30000} samples. Once we obtain our final measurement acquisition system, we train our DNN with the entire dataset. The algorithm needs seven iterations with this reduced dataset and almost \num{44} hours of processing time to obtain the final measurement acquisition system. In the last iteration, we spend \num{53} hours for training using the entire dataset composed of \num{300000} samples. The details of processing time is given in \Cref{tab:comp_time}. We can afford these \num{96} hours, i.e., \num{4} days of processing time for entire simulation as it is \textit{offline}. Once we obtain the trained DNN, we need a fraction of a second to evaluate the inverse solution.

\begin{table}[!htp]
	\centering
	\begin{tabular}{r|rccc}
		\toprule
		problem   & data size & iterations  & epochs & processing time [\si{\hour}]   \\
		\midrule
		Forward and inverse & \num{30000} & \num{7}  & \num{1200} & \num{43.77} \\
		Forward & \num{300000} & \num{1}  & \num{1200} & \num{18.69} \\
		Inverse & \num{300000} & \num{1}  & \num{3000} & \num{34.41} \\
		\bottomrule
	\end{tabular}
	\caption{Training times of the DNN.}
	\label{tab:comp_time}
\end{table}


\section{Numerical Results}\label{sec:numeric}

In this section, we describe our iterative algorithm for designing the measurement acquisition system. At each iteration, we evaluate the performance of the DNN-based inversion with an updated measurement acquisition system. For evaluation of the DNN approximation, different varieties of cross-plots of DNN-predicted measurement against \textit{ground truth} provide us with important information \cite{Shahriari_loss}. We focus here on two types of cross-plots that are crucial for the assessment of DNN-based inversions: \\

\begin{tabular}{c r c l}
	Cross-plots 1: & $\mathcal{F} \circ \mathcal{I}$ & vs &  $\mathcal{F} \circ \mathcal{I}_{\beta^\ast}$\\
	Cross-plots 2: & $\mathcal{I}$ & vs & $\mathcal{I}_{\beta^\ast}$
\end{tabular}
\\
Cross-plots of type 1 depict how well the composition of predicted inverse operator and forward function approximates the original measurements $\mathbf {m}$. These cross-plots indicate the quality of the inverse approximation. When the correlation is high, we can state that the DNN training has been successful. On the other side, cross-plots of type 2 also reflect on the non-uniqueness of $\cal I$. It is possible to obtain low-correlated cross-plots of type 2 due to non-uniqueness of $\cal I$ while cross-plots 1 exhibit high correlation.

\subsection{Optimization of measurement acquisition system}
We start our training with the coaxial measurement with short-spaced LWD instrument because they provide a fair assessment of the resistivity of the formation near the well. This measurement is essential to characterize $\rho_h$, which is a basic quantity needed to obtain a good inversion result. 

\begin{figure}[hb!]
	\centering 
	
	\underline{\textbf{Iteration \num{1}}}
	
	\medskip 
	
	\begin{subfigure}[t]{0.48\linewidth}
		\centering
		\subcaptionbox*{Coaxial measurement with the short-spaced LWD instrument}{\includegraphics[width=1.0\linewidth,trim={1.0cm 0.0cm 1.0cm 0.0cm},clip] {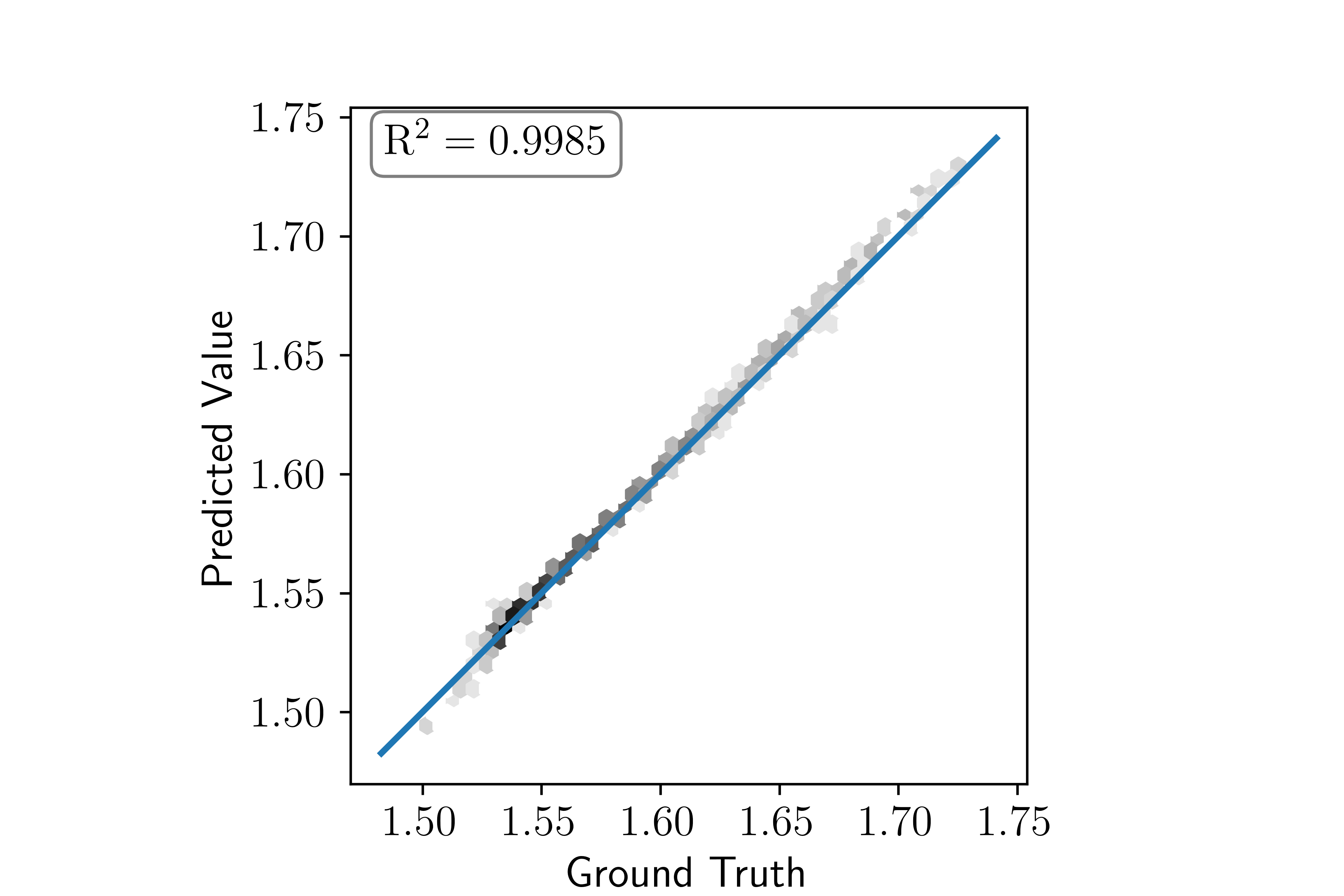}}
	\end{subfigure}
	\hfill
	\begin{subfigure}[t]{0.48\linewidth}
		\centering
		\subcaptionbox*{Symmterized directional measurement with the short-spaced deep azimuthal instrument} {\includegraphics [width=1.0\linewidth, trim={1.0cm 0.0cm 1.0cm 0.0cm}, clip] {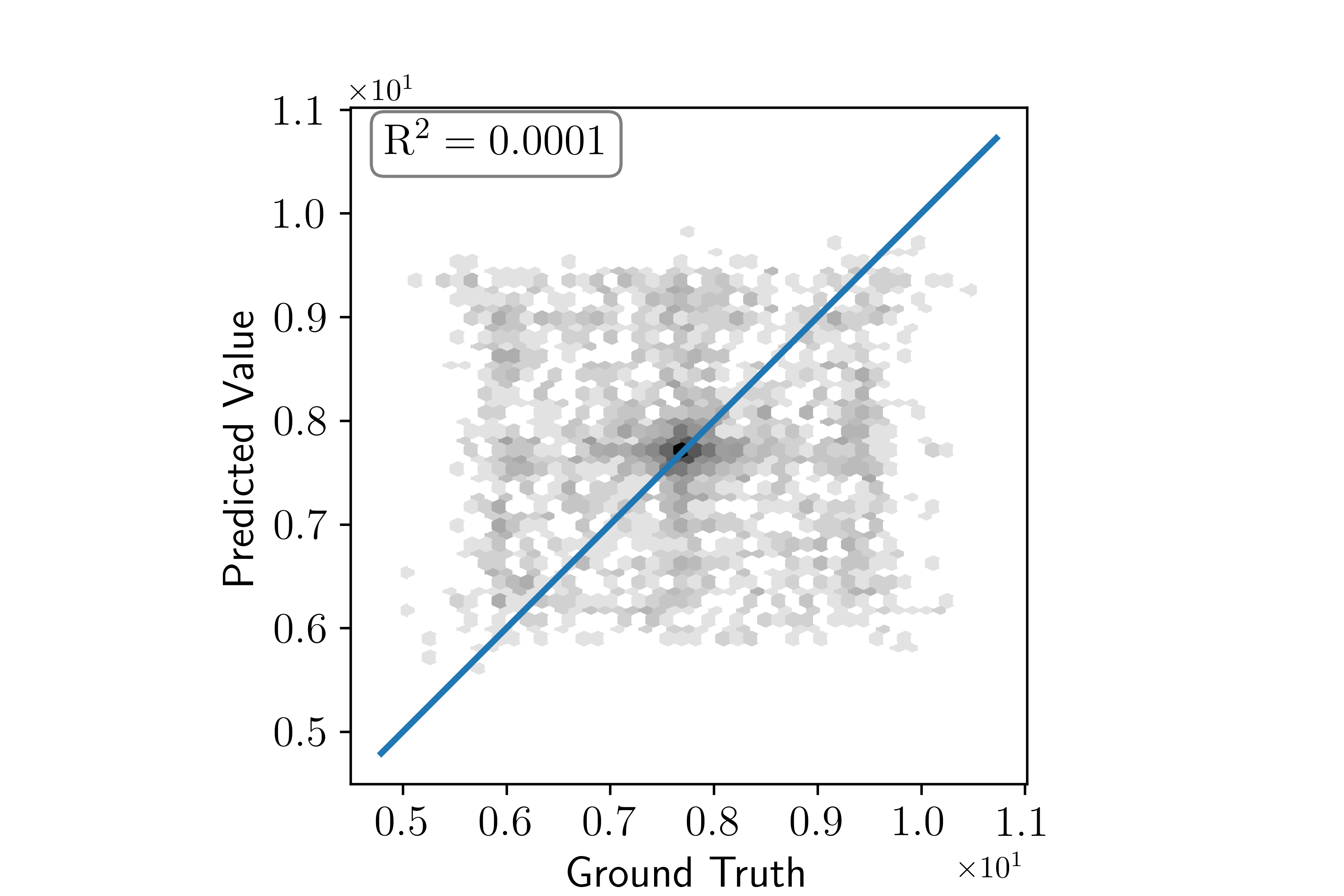}}
	\end{subfigure}	
	\\
	
	\medskip 
	
	\underline{\textbf{Iteration \num{2}}}
	
	\medskip 
	
	\begin{subfigure}[t]{0.48\linewidth}
		\centering
		\subcaptionbox*{Symmterized directional measurement with the short-spaced deep azimuthal instrument}{\includegraphics [width=1.0\linewidth, trim={1.0cm 0.0cm 1.0cm 0.0cm}, clip] {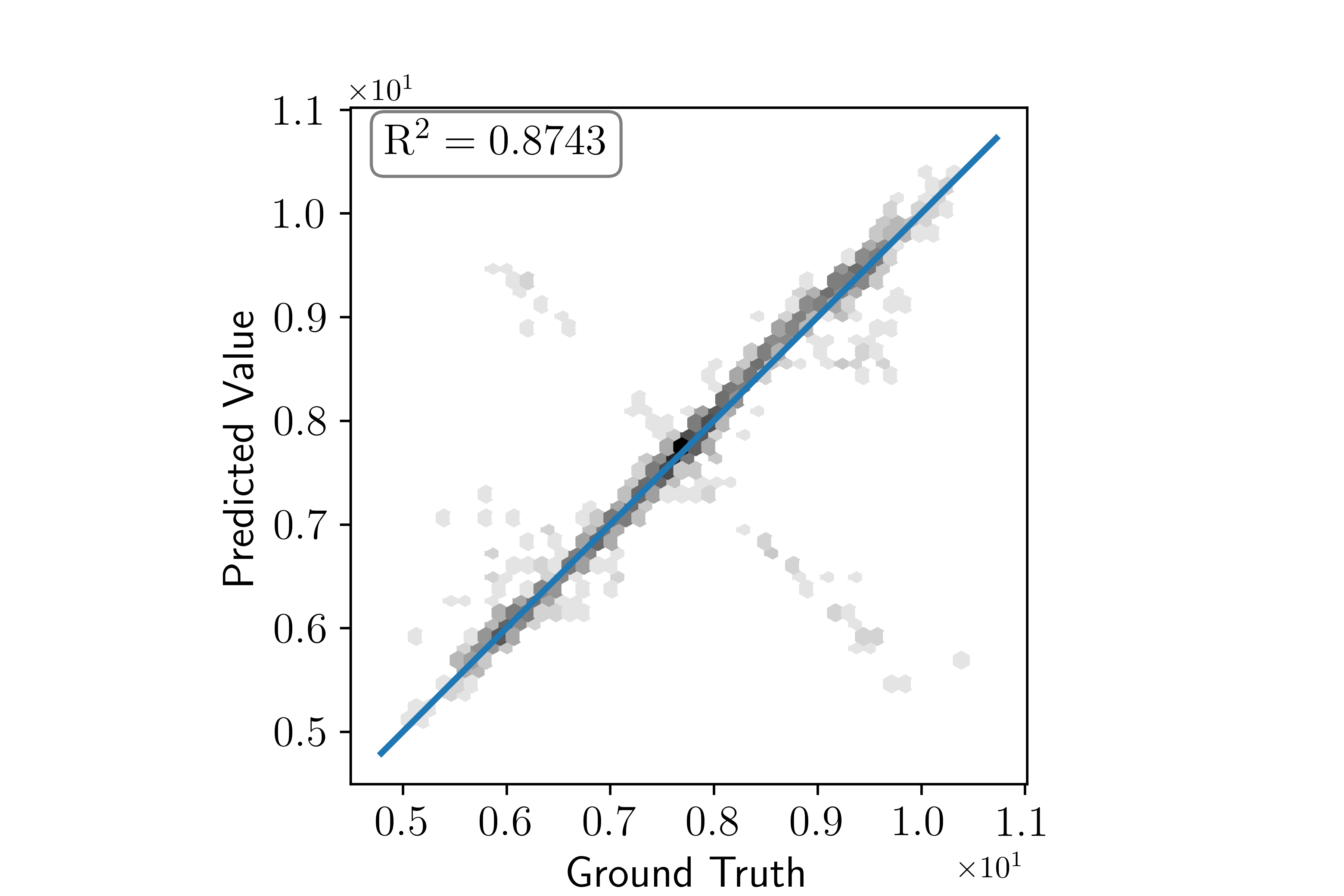}}
	\end{subfigure}
	\hfill
	\begin{subfigure}[t]{0.48\linewidth}
		\centering
		\subcaptionbox*{Symmetrized directional measurement with the short-spaced LWD instrument} {\includegraphics[width=1.0\linewidth, trim={1.0cm 0.0cm 1.0cm 0.0cm}, clip] {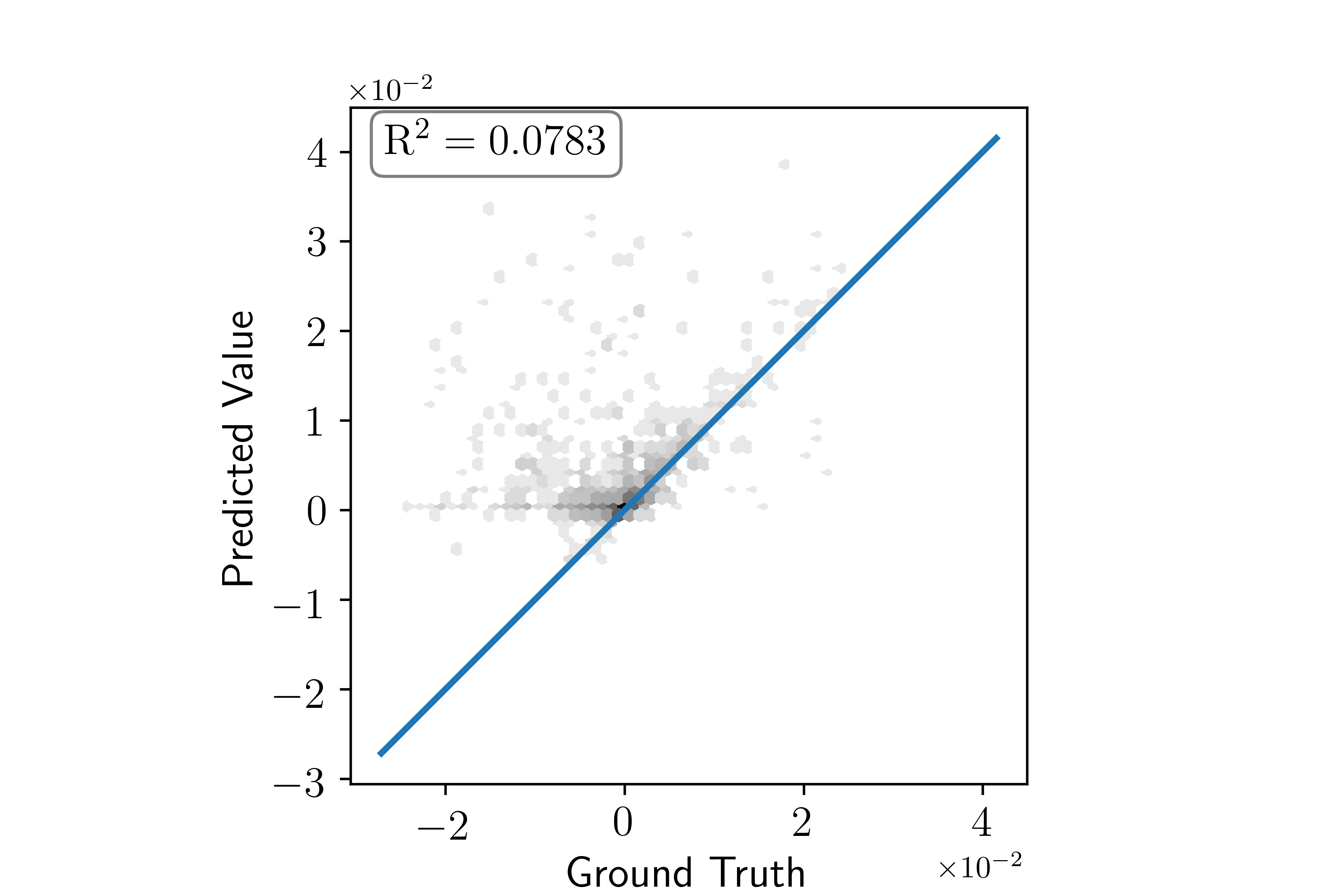}}
	\end{subfigure}
	\caption{Cross-plots of type 1. Predicted value vs. ground truth for attenuation -- iterations \numrange[range-phrase= -- ]{1}{2}. First column corresponds to the last added measurement and second column is associated with the measurement with the worst $R^2$ value (after averaging the $R^2$ values of attenuation and phase).}
	\label{fig:cross-plots11}
\end{figure}

\paragraph{\it{\bf {Iteration 1.}}} In this iteration, we obtain the first approximation of ${\cal I}_{\beta^\ast}$ using a measurement acquisition system with the coaxial measurement with the short-spaced LWD instrument. After that, we calculate the $R^2$ factor of all other measurements. The algorithm selects the worst correlated measurement, which corresponds to the symmetrized directional measurement with the short-spaced deep azimuthal instrument, as depicted in \Cref{fig:cross-plots11}. We only display attenuation in the figures for brevity. However, the algorithm considers $R^2$ values of both phase difference and attenuation except for those measurements where phase differences show an artificial discontinuity.

This selection is consistent with the known physics of borehole resistivity logging instruments since the symmetrized directional measurement in a horizontal well can differentiate between the top and bottom, while co-axial measurements are unable to make such a distinction. Thus, a directional measurements is needed to distinguish between the upper and lower layers.

\begin{figure}[b!]	
 \centering
 \hspace{0.5cm}\underline{\textbf{Iteration \num{1}}}
 
 \medskip
\begin{subfigure}[t]{0.32\linewidth}
\centering

\hspace{.6cm}$\rho_h$

\medskip

\includegraphics[width=1.2\linewidth,trim={1.0cm 0.0cm 1.0cm 0.0cm},clip]{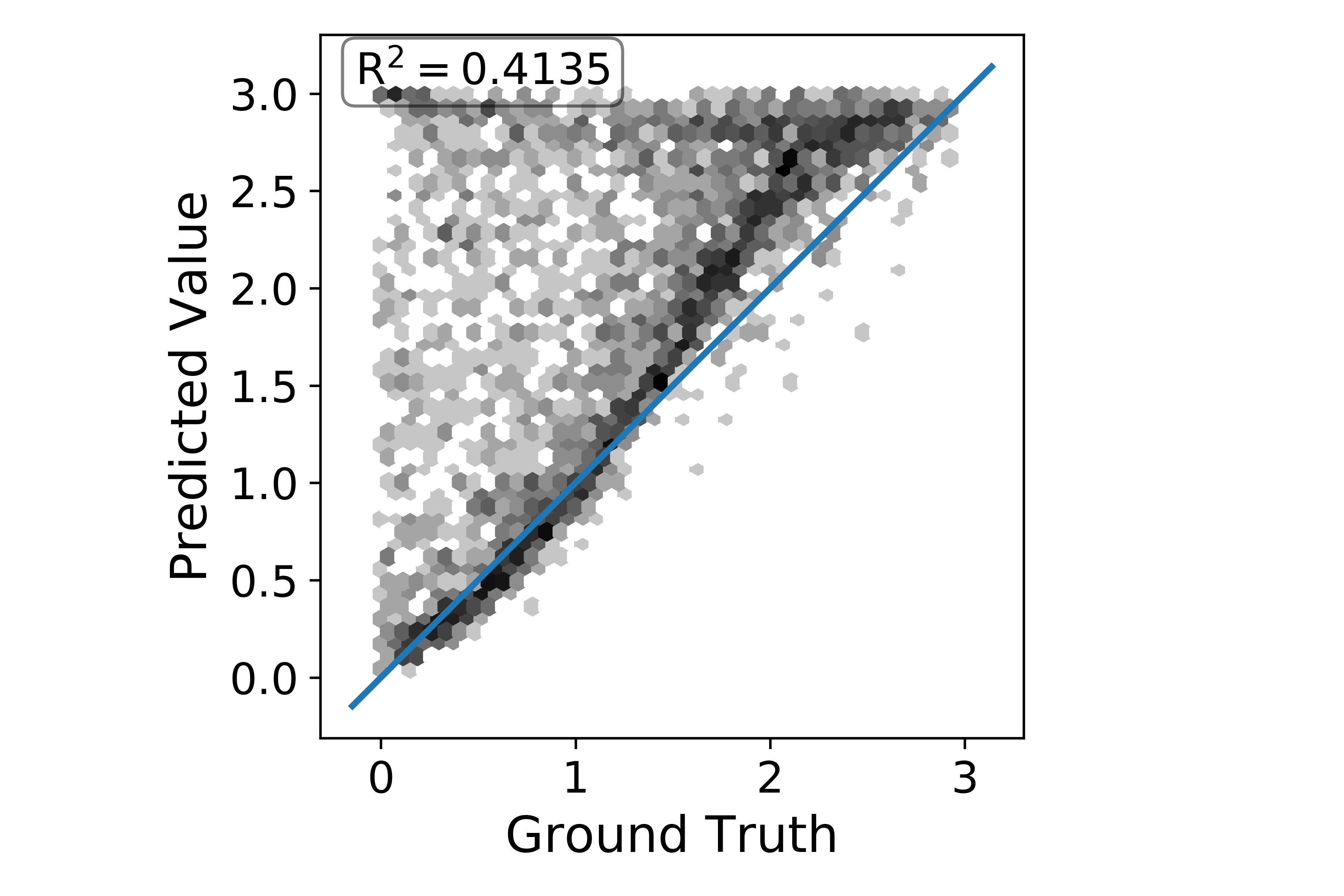}
\end{subfigure}
\begin{subfigure}[t]{0.32\linewidth}
\centering	
\hspace{.6cm}$\rho_u$

\medskip

\includegraphics[width=1.2\linewidth,trim={1.0cm 0.0cm 1.0cm 0.0cm}, clip] {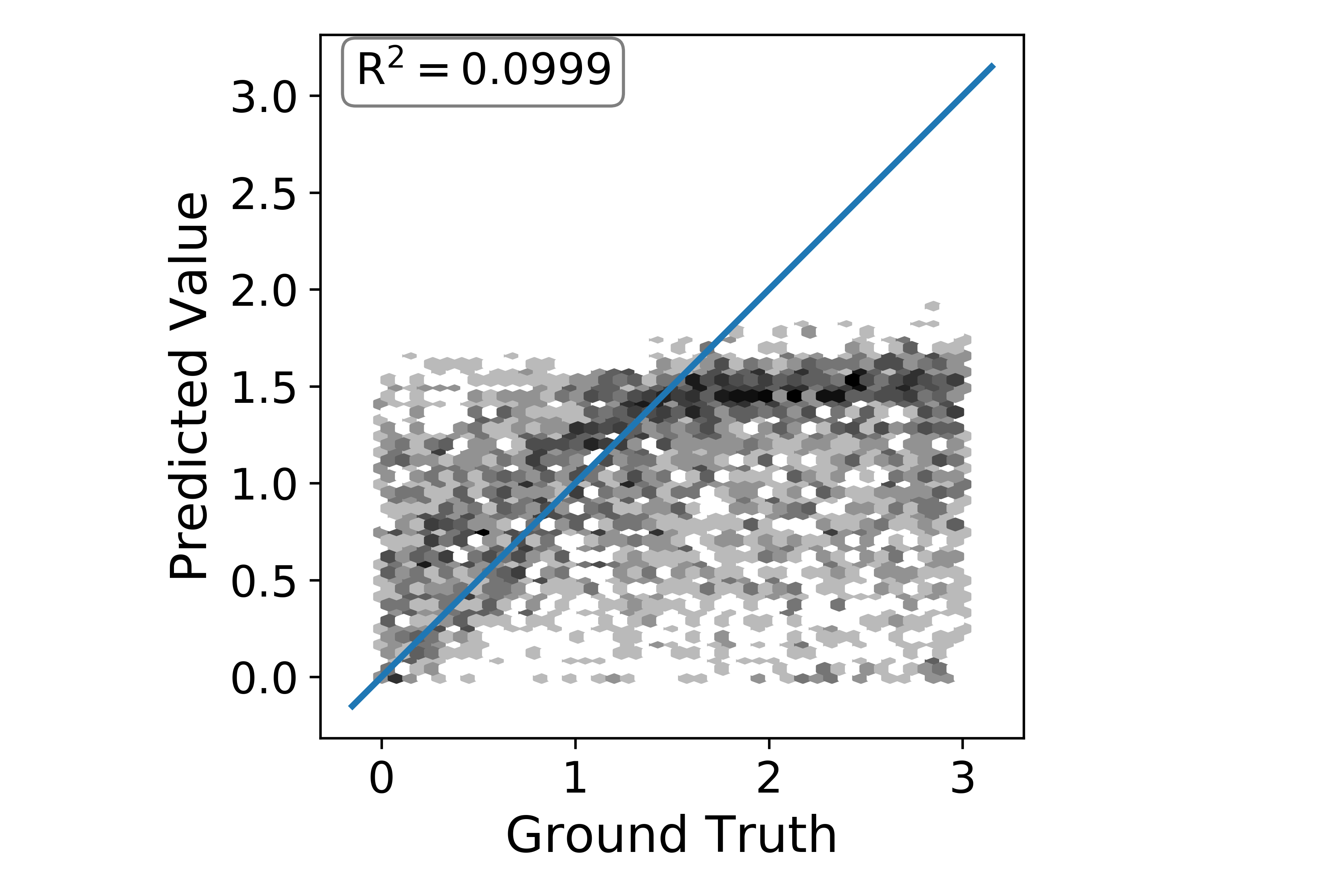}
\end{subfigure}
\begin{subfigure}[t]{0.32\linewidth}
\centering
\hspace{.6cm} $d_u$

\medskip

\includegraphics[width=1.2\linewidth,trim={1.0cm 0.0cm 1.0cm 0.0cm},clip]{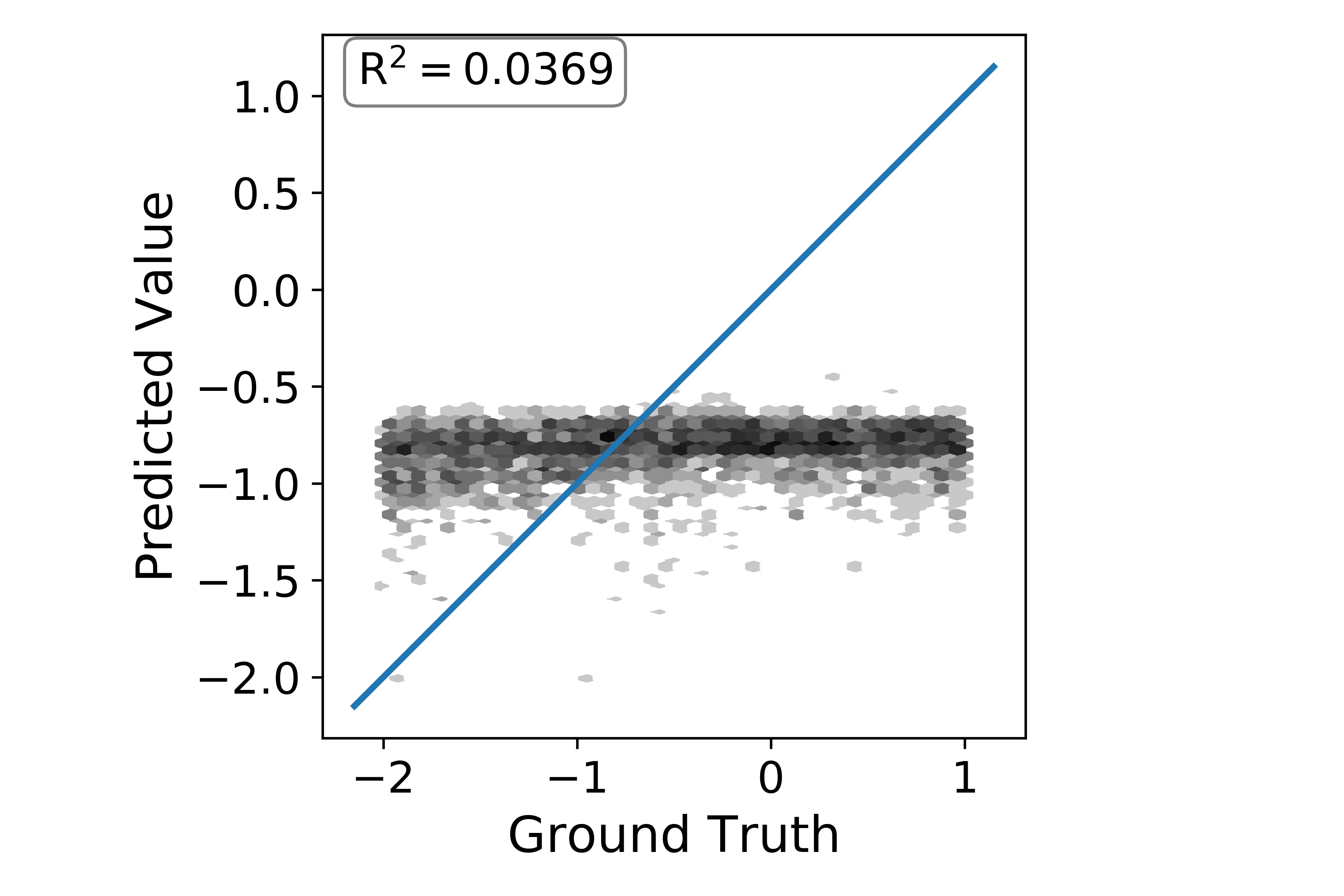}
\end{subfigure}\\

\medskip

\hspace{0.5cm}\underline{\textbf{Iteration \num{2}}}

\medskip

\begin{subfigure}[t]{0.32\linewidth}
	\centering
	\includegraphics[width=1.2\linewidth,trim={1.0cm 0.0cm 1.0cm 0.0cm},clip]{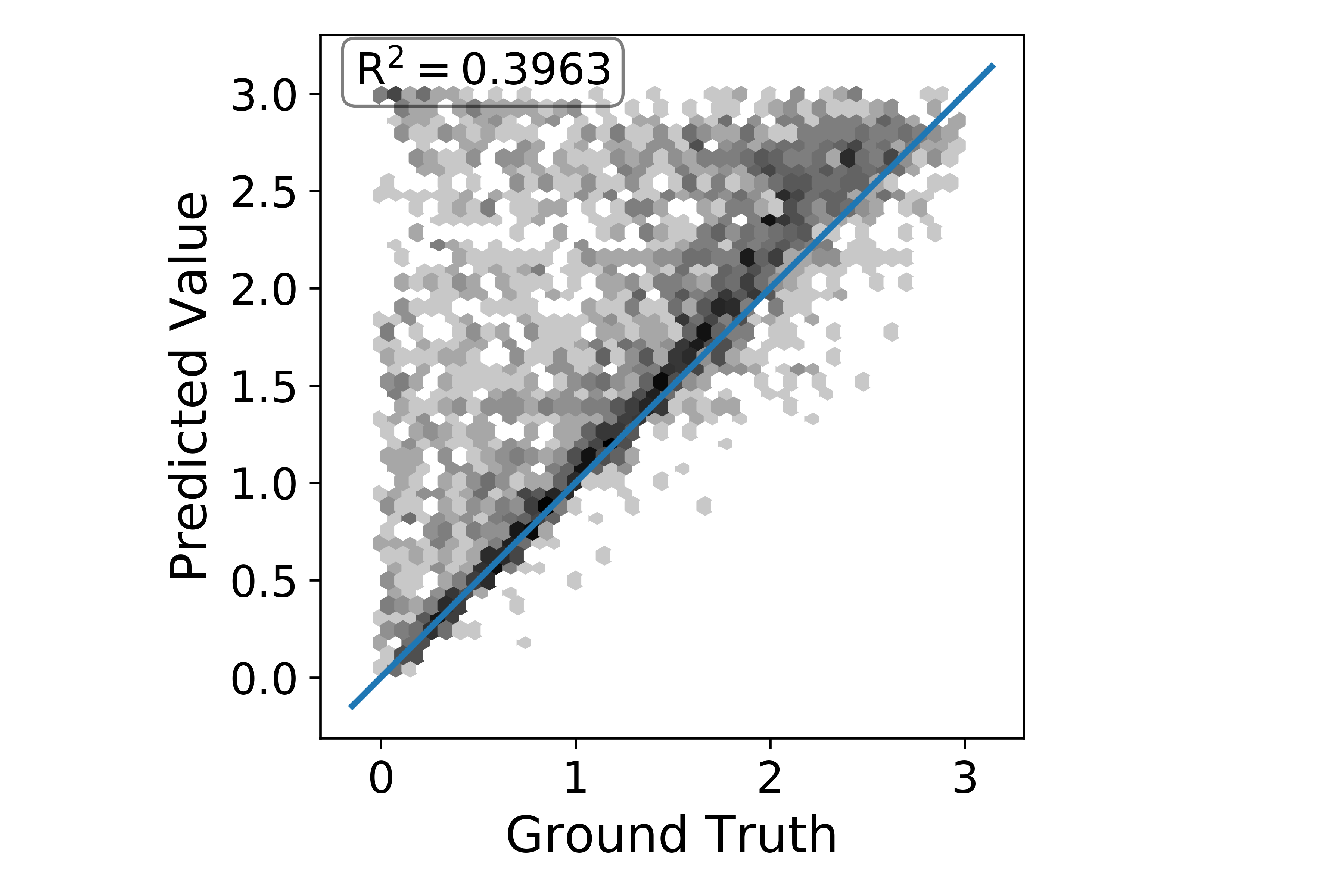}
\end{subfigure}
\begin{subfigure}[t]{0.32\linewidth}
	\centering	
	\includegraphics[width=1.2\linewidth,trim={1.0cm 0.0cm 1.0cm 0.0cm}, clip] {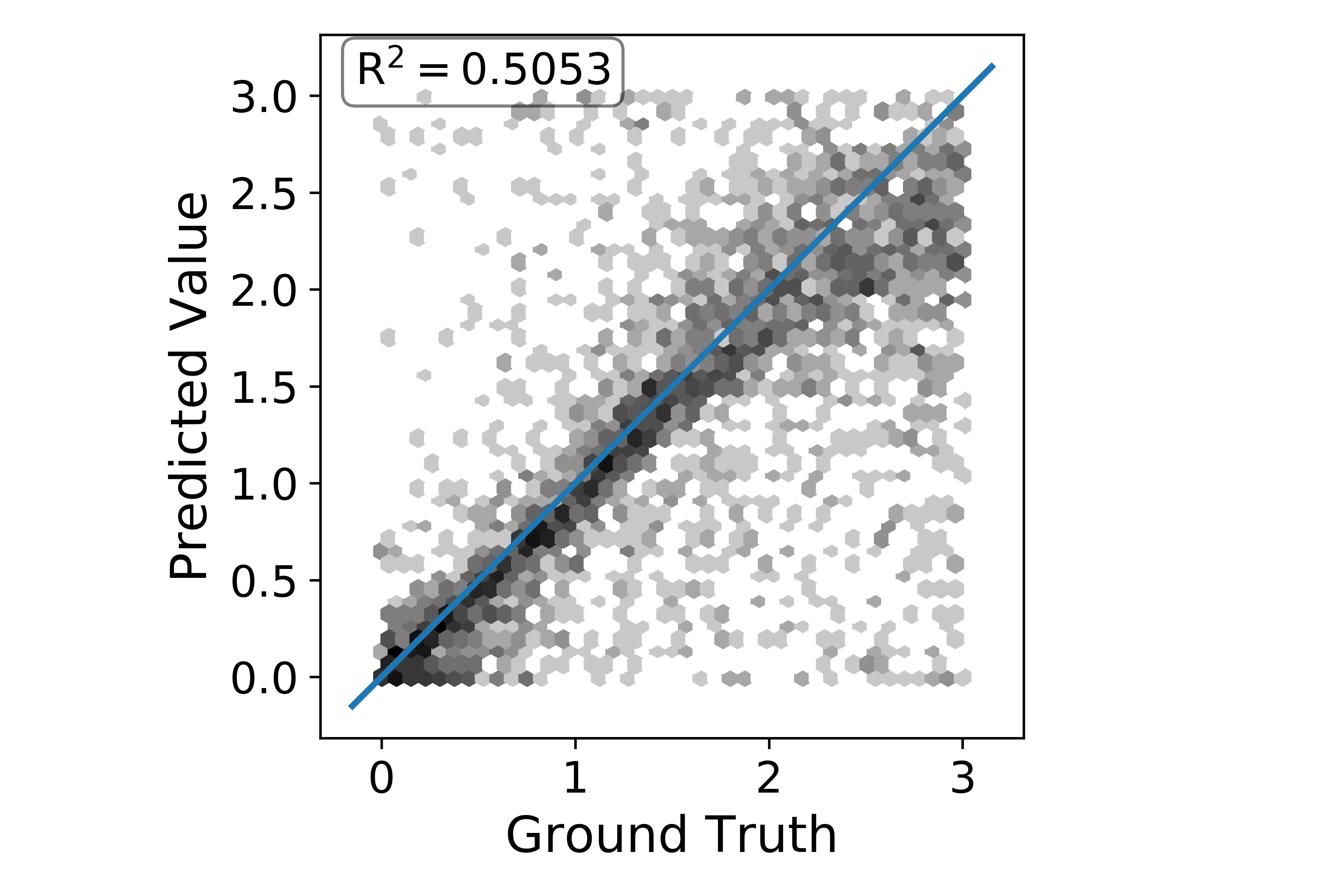}
\end{subfigure}
\begin{subfigure}[t]{0.32\linewidth}
	\centering
	\includegraphics[width=1.2\linewidth,trim={1.0cm 0.0cm 1.0cm 0.0cm},clip]{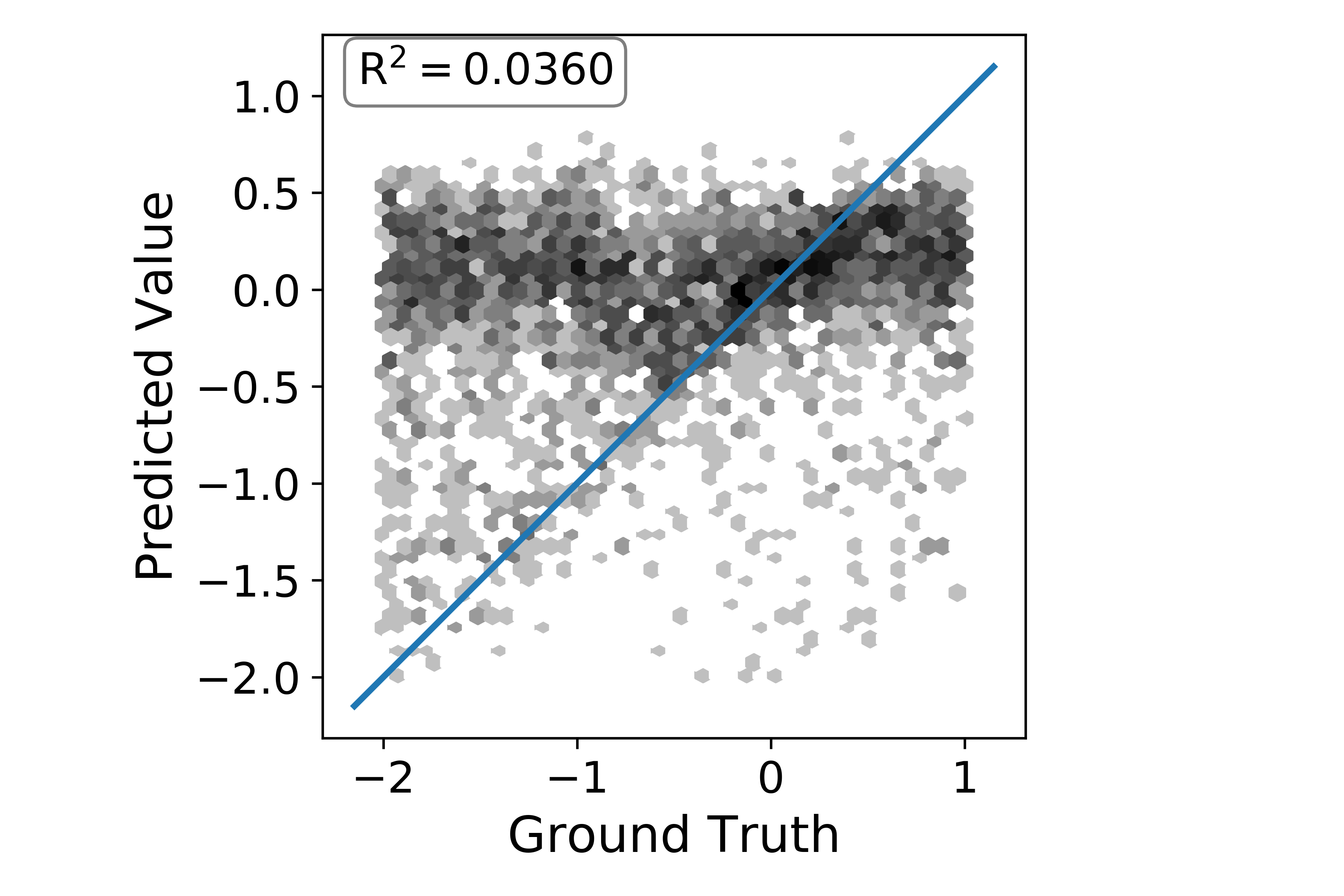}
\end{subfigure}\\
\medskip
\hspace{0.5cm}\underline{\textbf{Iteration \num{3}}}\\
\medskip
\begin{subfigure}[t]{0.32\linewidth}
	\centering
	\includegraphics[width=1.2\linewidth,trim={1.0cm 0.0cm 1.0cm 0.0cm},clip]{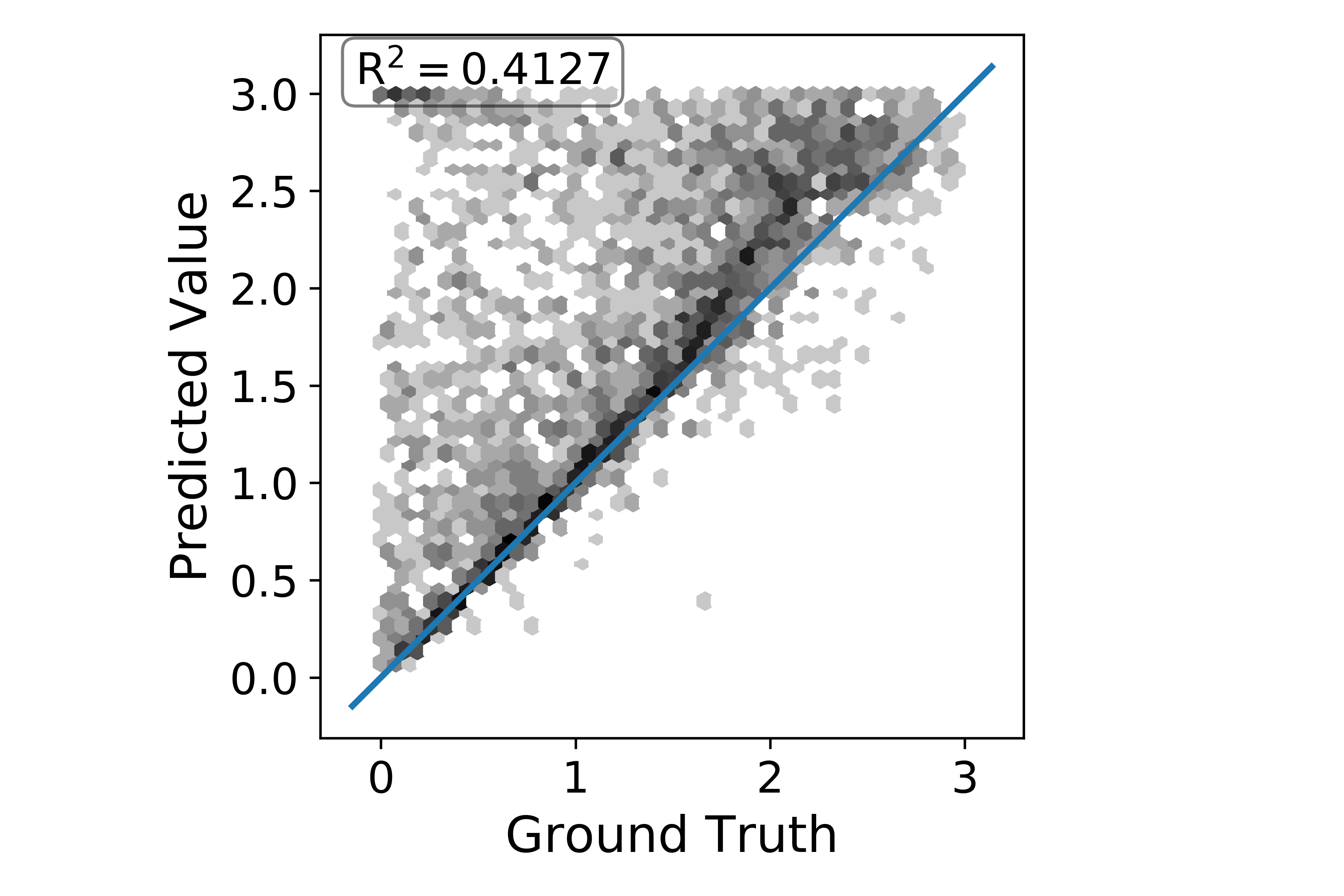}
\end{subfigure}
\begin{subfigure}[t]{0.32\linewidth}
	\centering	
	\includegraphics[width=1.2\linewidth,trim={1.0cm 0.0cm 1.0cm 0.0cm}, clip] {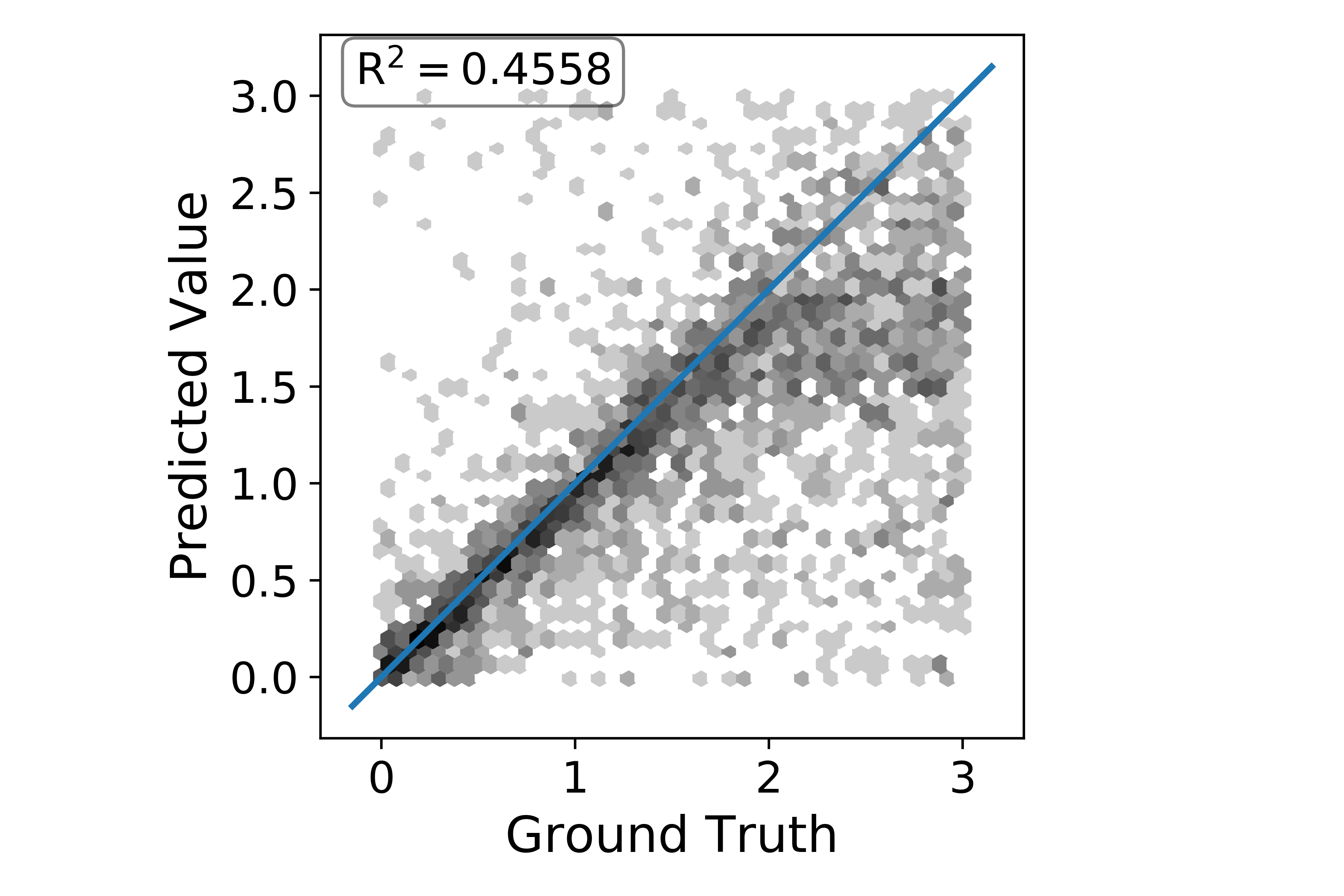}
\end{subfigure}
\begin{subfigure}[t]{0.32\linewidth}
	\centering
	\includegraphics[width=1.2\linewidth,trim={1.0cm 0.0cm 1.0cm 0.0cm},clip]{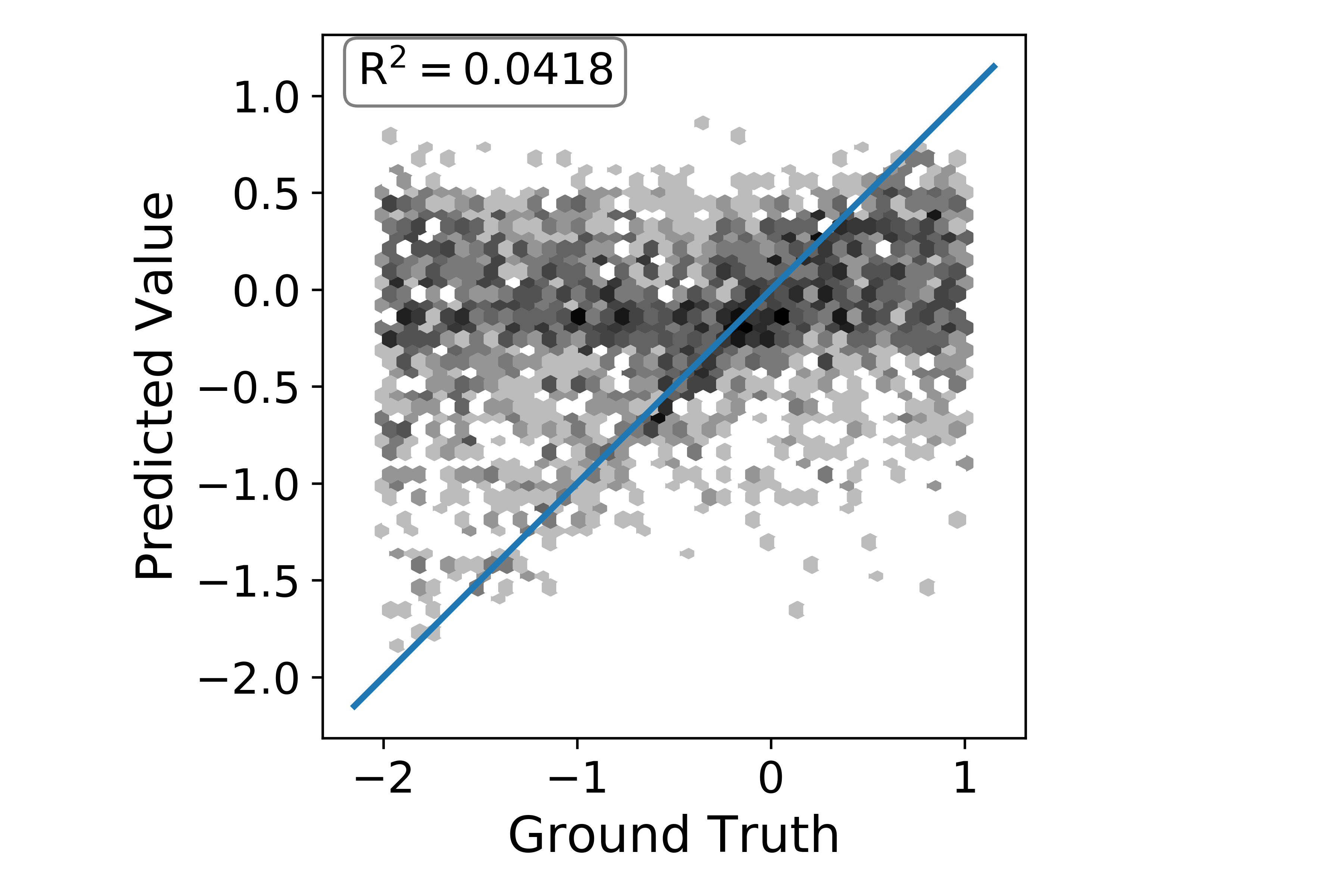}
\end{subfigure}\\
\medskip
\hspace{0.5cm}\underline{\textbf{Iteration \num{4}}}\\
\medskip
\begin{subfigure}[t]{0.32\linewidth}
	\centering
	\includegraphics[width=1.2\linewidth,trim={1.0cm 0.0cm 1.0cm 0.0cm},clip]{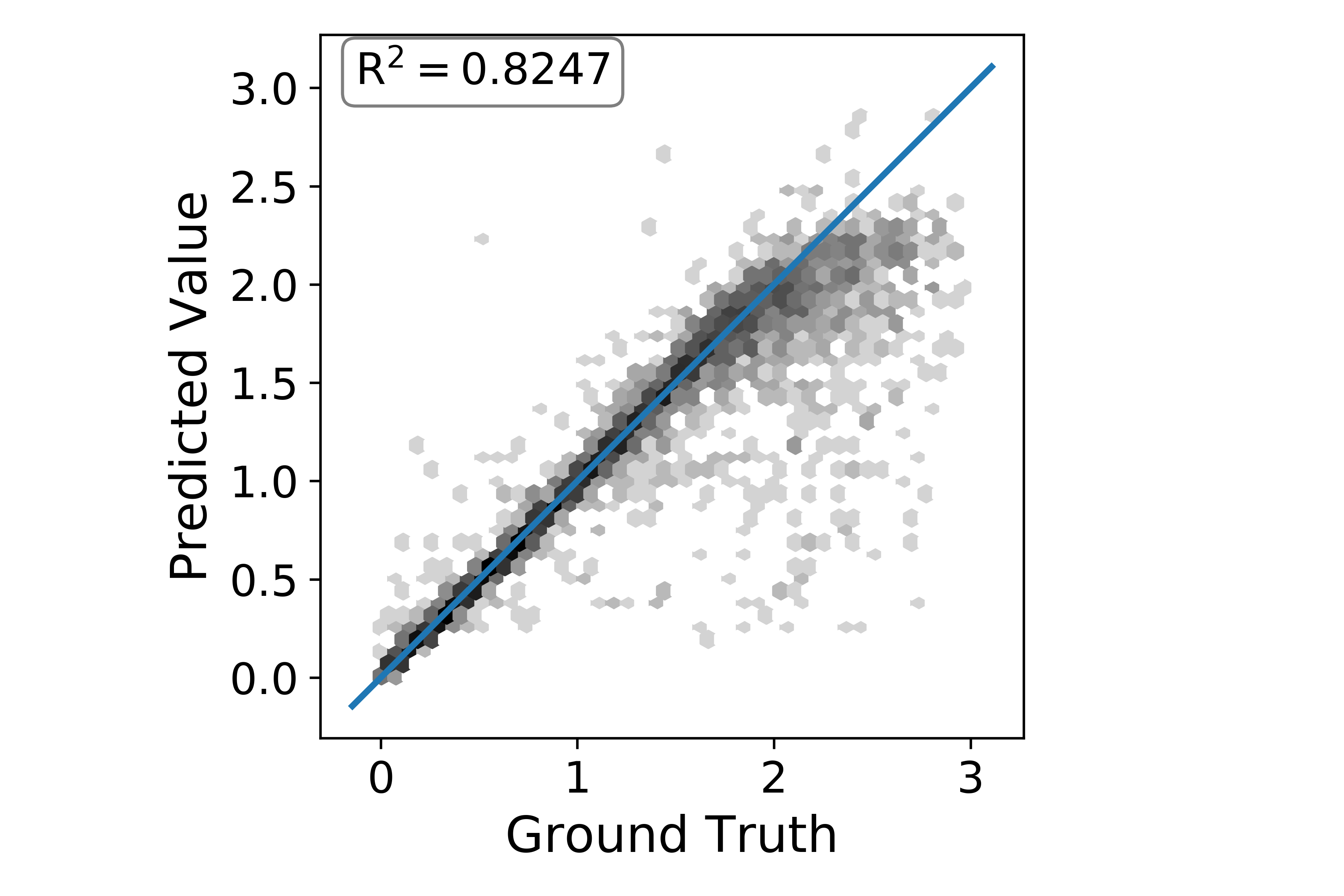}
\end{subfigure}
\begin{subfigure}[t]{0.32\linewidth}
	\centering	
	\includegraphics[width=1.2\linewidth,trim={1.0cm 0.0cm 1.0cm 0.0cm}, clip] {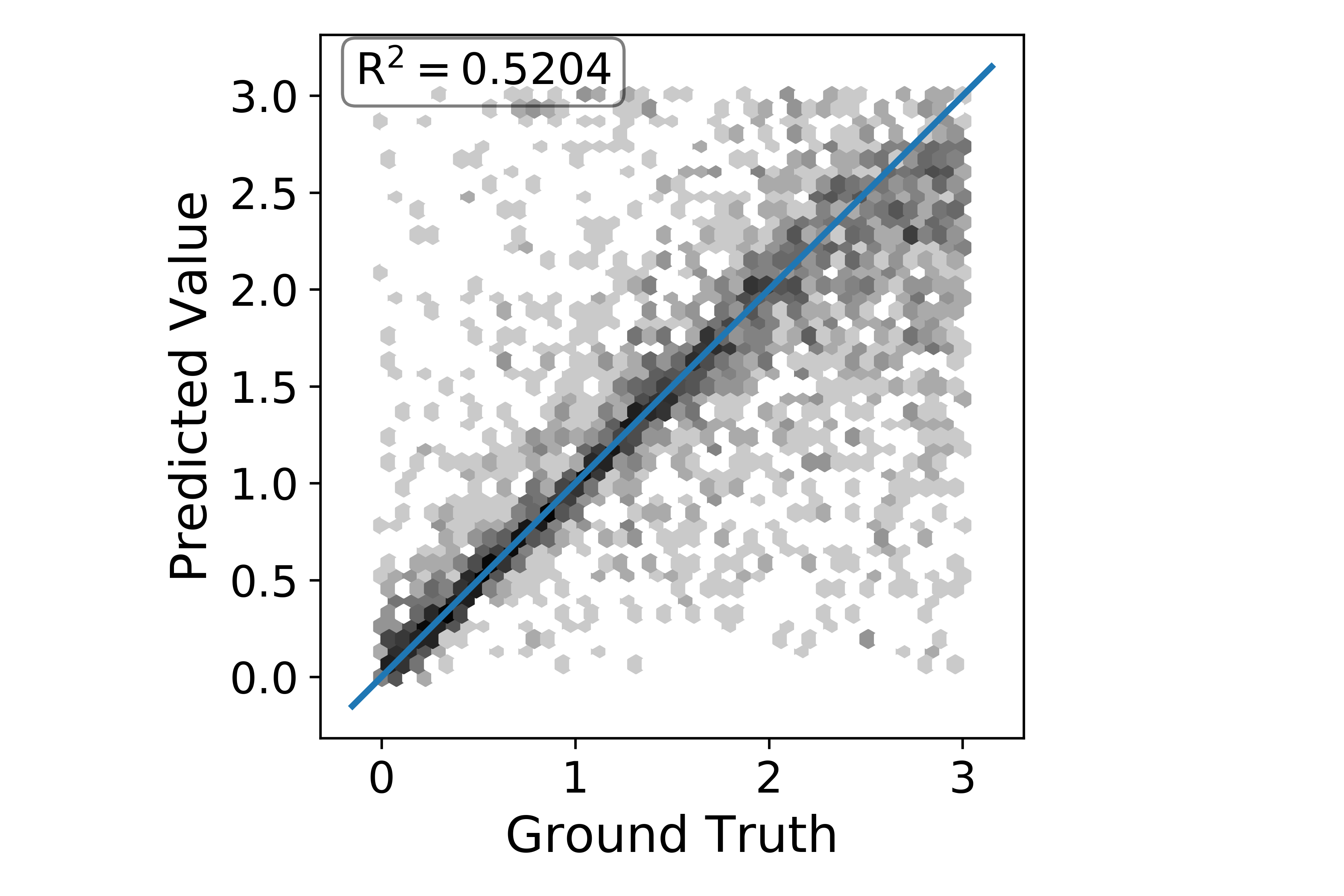}
\end{subfigure}
\begin{subfigure}[t]{0.32\linewidth}
	\centering
	\includegraphics[width=1.2\linewidth,trim={1.0cm 0.0cm 1.0cm 0.0cm},clip]{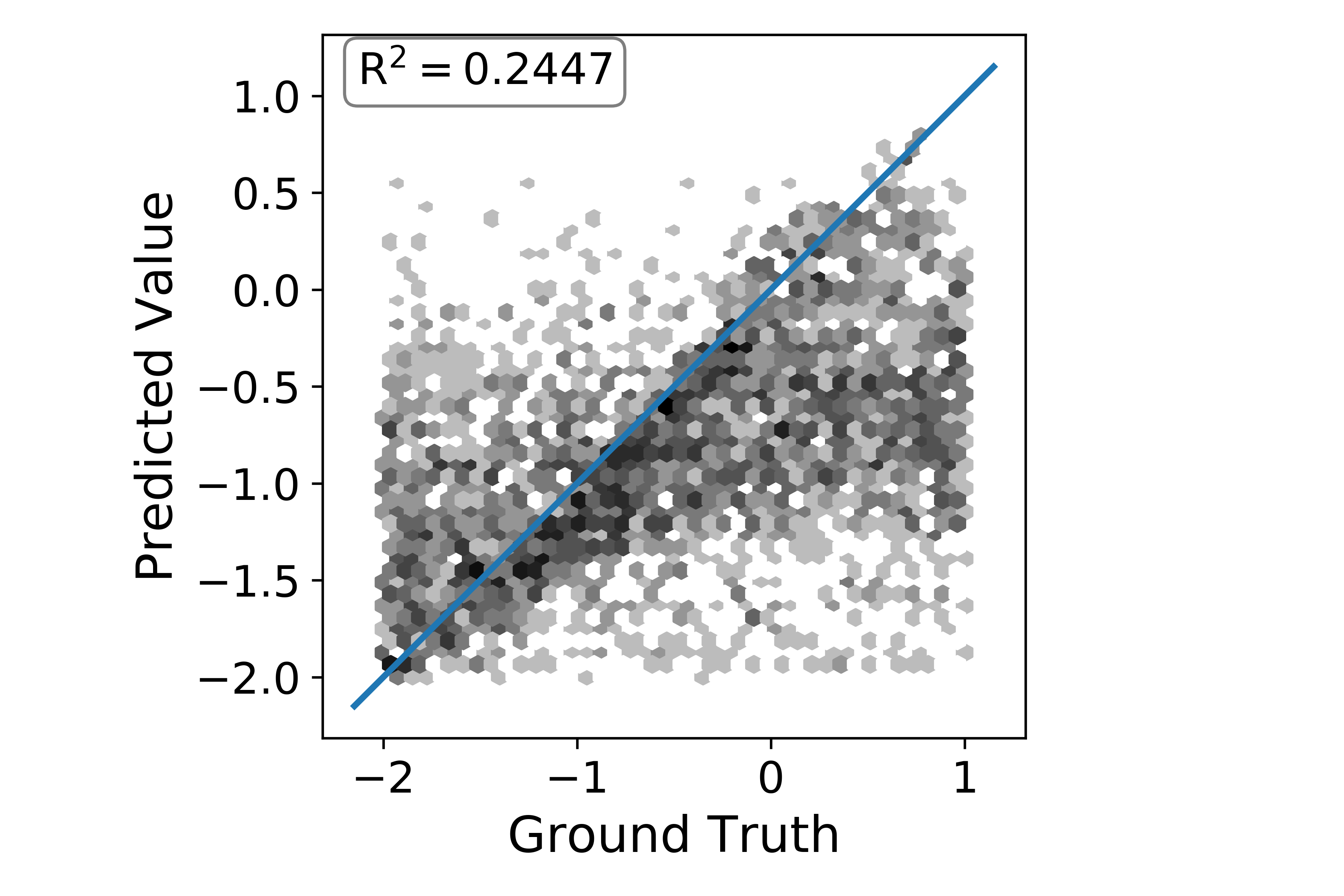}
\end{subfigure}
\caption{ Predicted value vs ground truth for three subsurface material properties. Horizontal resistivity $\rho_h$, resistivity of layer located above the current layer $\rho_u$, and vertical distance from the current logging position to the upper boundary $d_u$ .}
\label{fig:cross-plot21}
\end{figure}

\begin{figure}[b!]	
 	\centering
 	\hspace{0.5cm}\underline{\textbf{Iteration \num{5}}}
 	
 	\medskip
 	\begin{subfigure}[t]{0.32\linewidth}
 		\centering
 		\hspace{.6cm}$\rho_h$
 		\medskip
 		\includegraphics[width=1.2\linewidth,trim={1.0cm 0.0cm 1.0cm 0.0cm},clip]{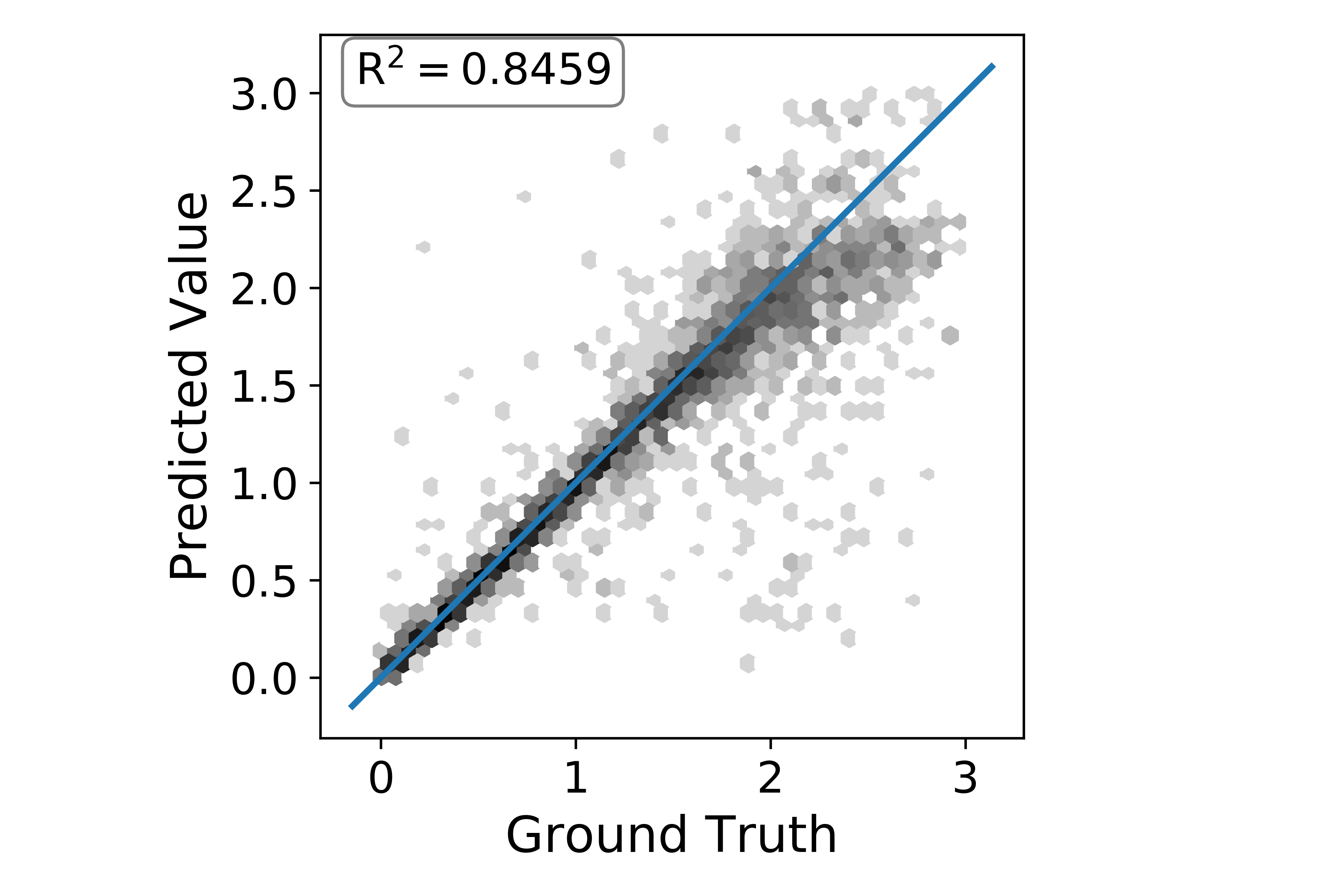}
 	\end{subfigure}
 	\begin{subfigure}[t]{0.32\linewidth}
 		\centering		
 		\hspace{.6cm}$\rho_u$
 		\medskip
 		\includegraphics[width=1.2\linewidth,trim={1.0cm 0.0cm 1.0cm 0.0cm}, clip] {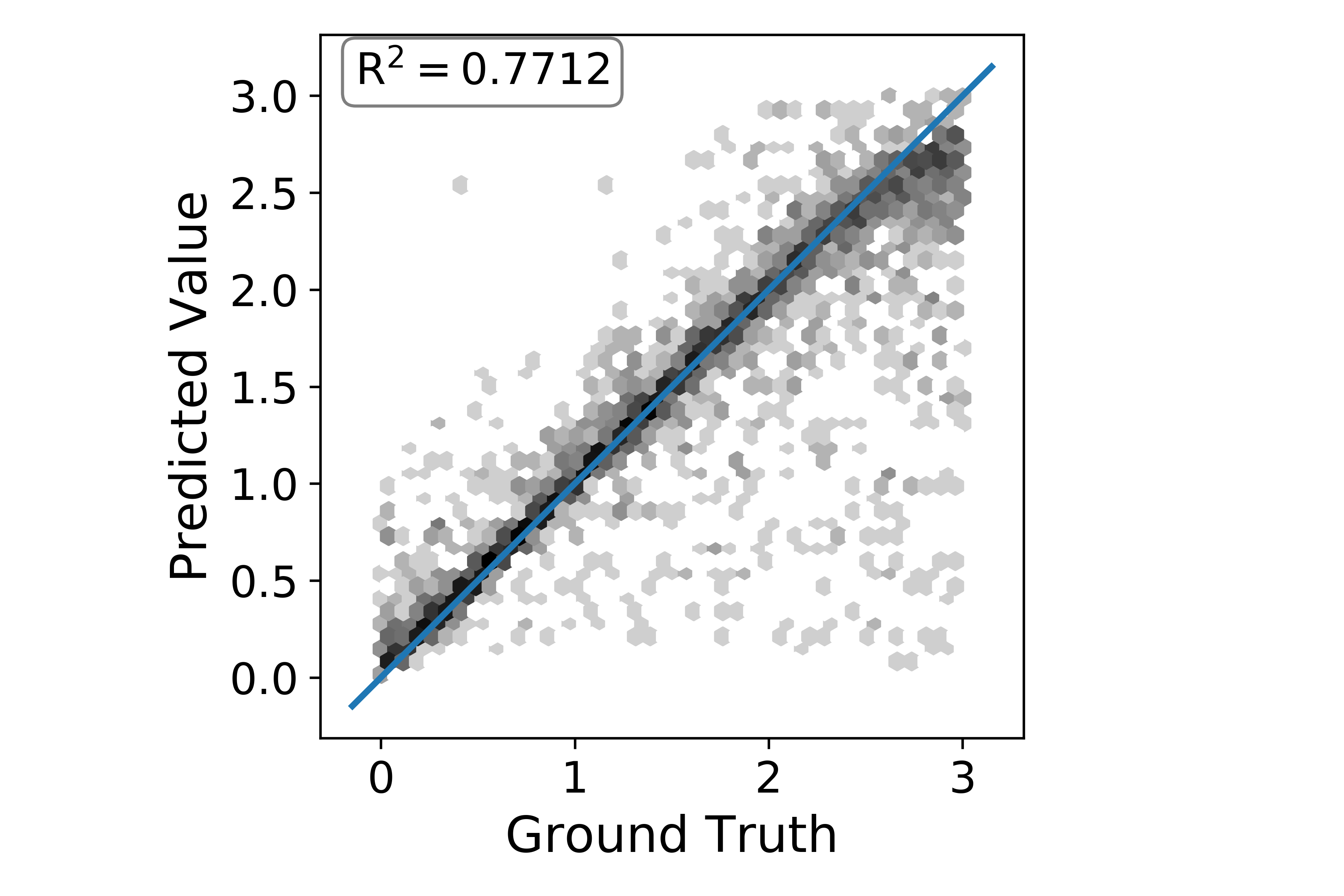}
 	\end{subfigure}
 	\begin{subfigure}[t]{0.32\linewidth}
 		\centering
 		\hspace{.6cm} $d_u$
 		\medskip
 		\includegraphics[width=1.2\linewidth,trim={1.0cm 0.0cm 1.0cm 0.0cm},clip]{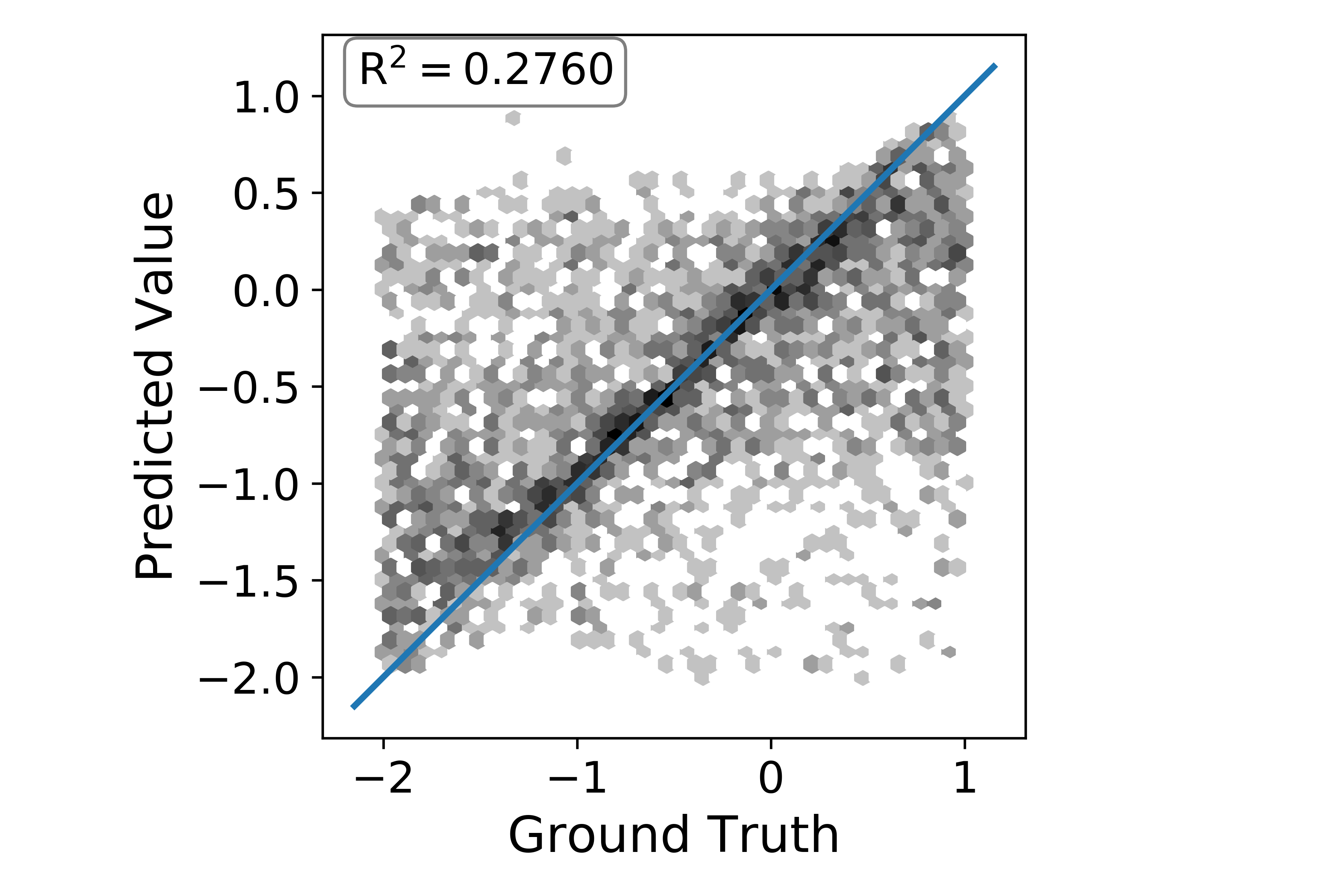}
 	\end{subfigure}\\
 	
 	\medskip
 	
 	\hspace{0.5cm}\underline{\textbf{Iteration \num{6}}}
 	
 	\medskip
 	
 	\begin{subfigure}[t]{0.32\linewidth}
 		\centering
 		\includegraphics[width=1.2\linewidth,trim={1.0cm 0.0cm 1.0cm 0.0cm},clip]{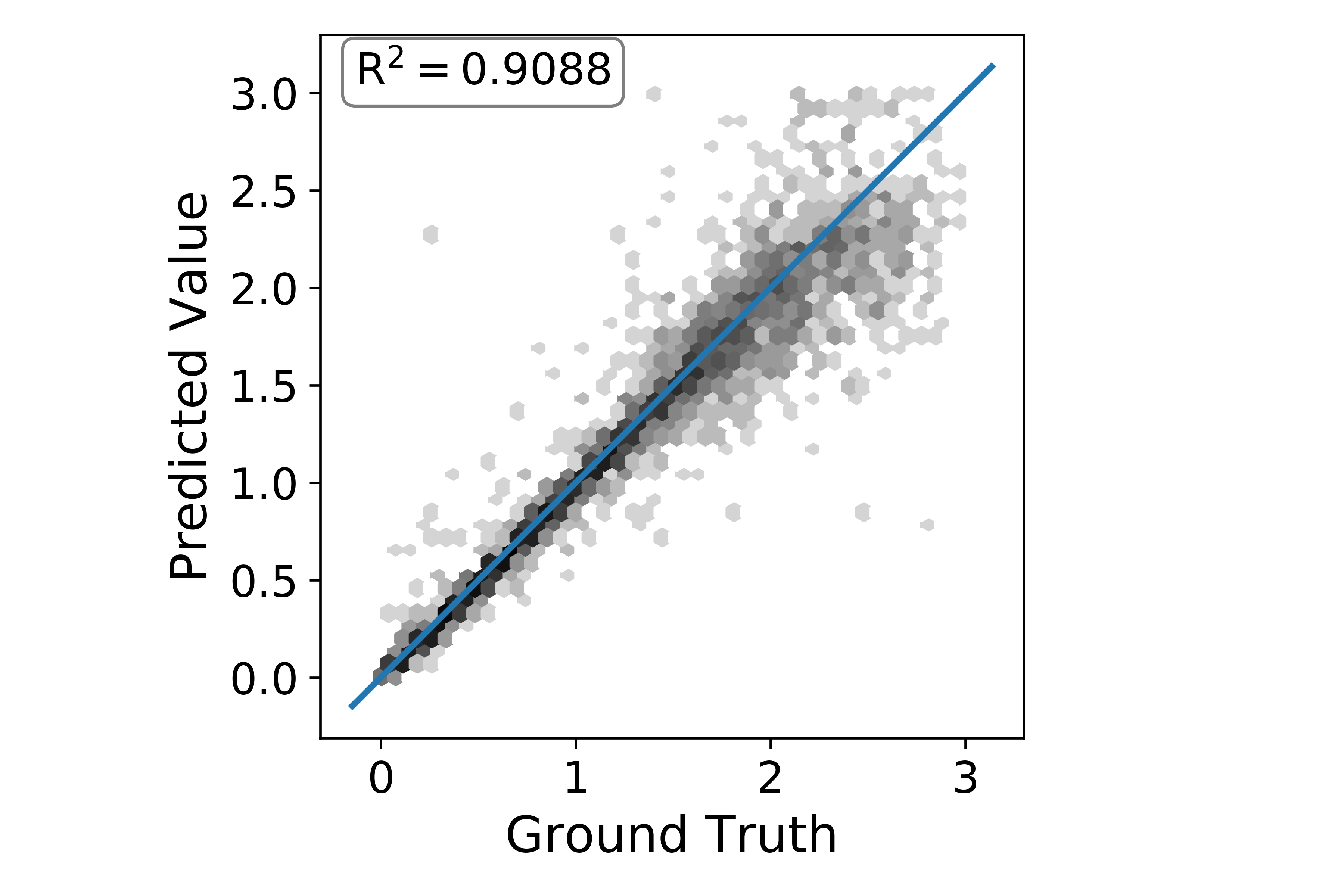}
 	\end{subfigure}
 	\begin{subfigure}[t]{0.32\linewidth}
 		\centering	
 		\includegraphics[width=1.2\linewidth,trim={1.0cm 0.0cm 1.0cm 0.0cm}, clip] {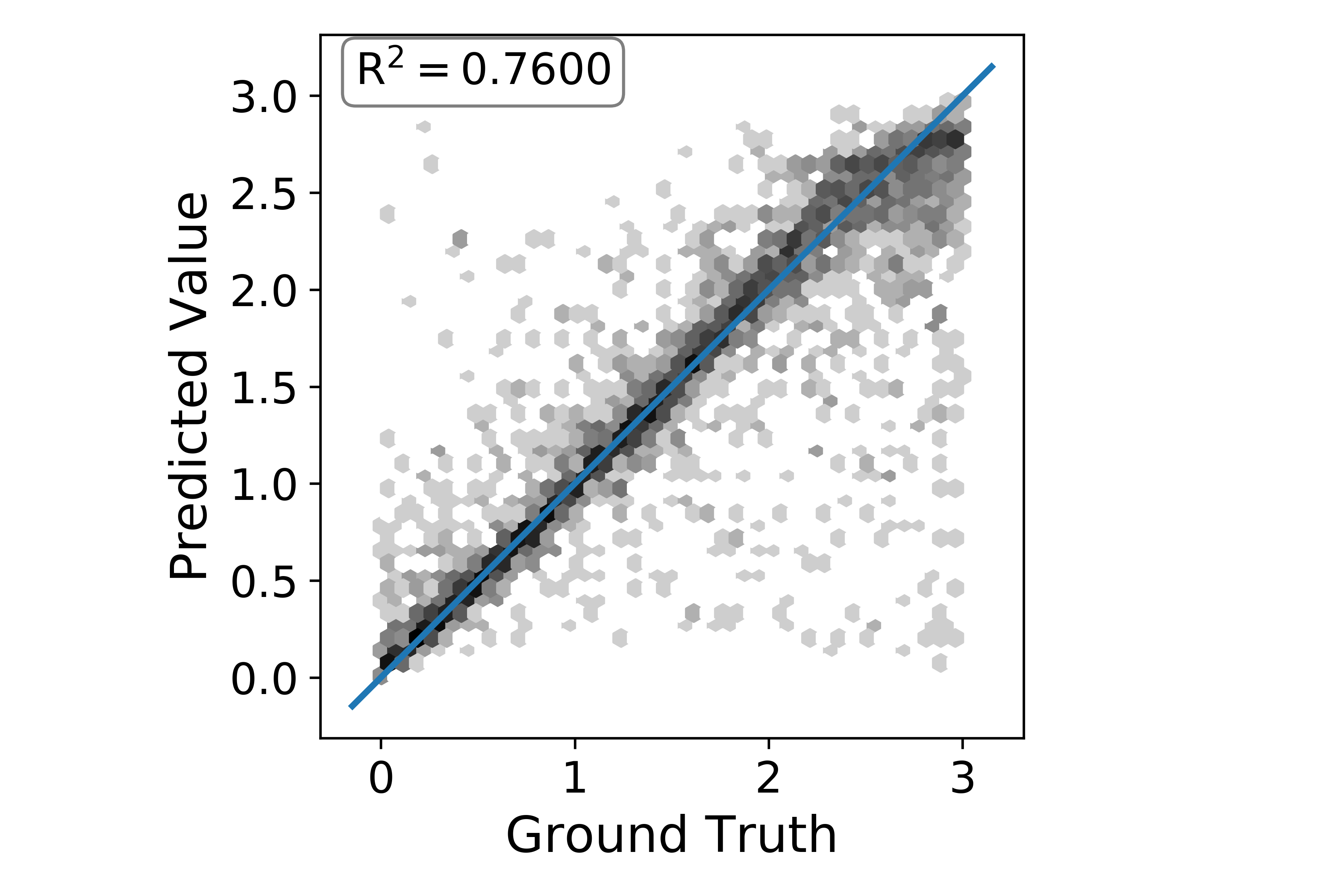}
 	\end{subfigure}
 	\begin{subfigure}[t]{0.32\linewidth}
 		\centering
 		\includegraphics[width=1.2\linewidth,trim={1.0cm 0.0cm 1.0cm 0.0cm},clip]{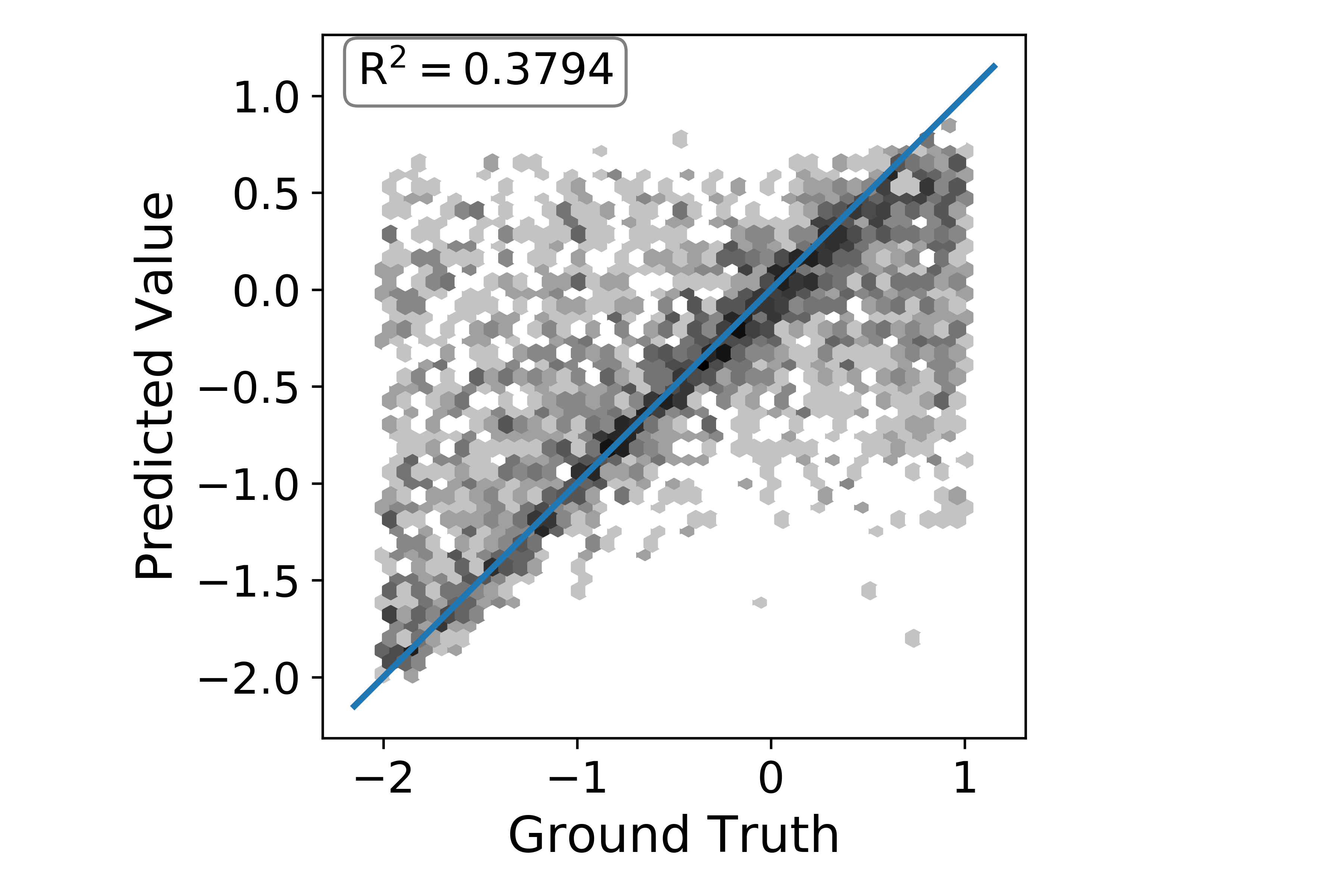}
 	\end{subfigure}\\
 	\medskip
 	\hspace{0.5cm}\underline{\textbf{Iteration \num{7}}}\\
 	\medskip
 	\begin{subfigure}[t]{0.32\linewidth}
 		\centering
 		\includegraphics[width=1.2\linewidth,trim={1.0cm 0.0cm 1.0cm 0.0cm},clip]{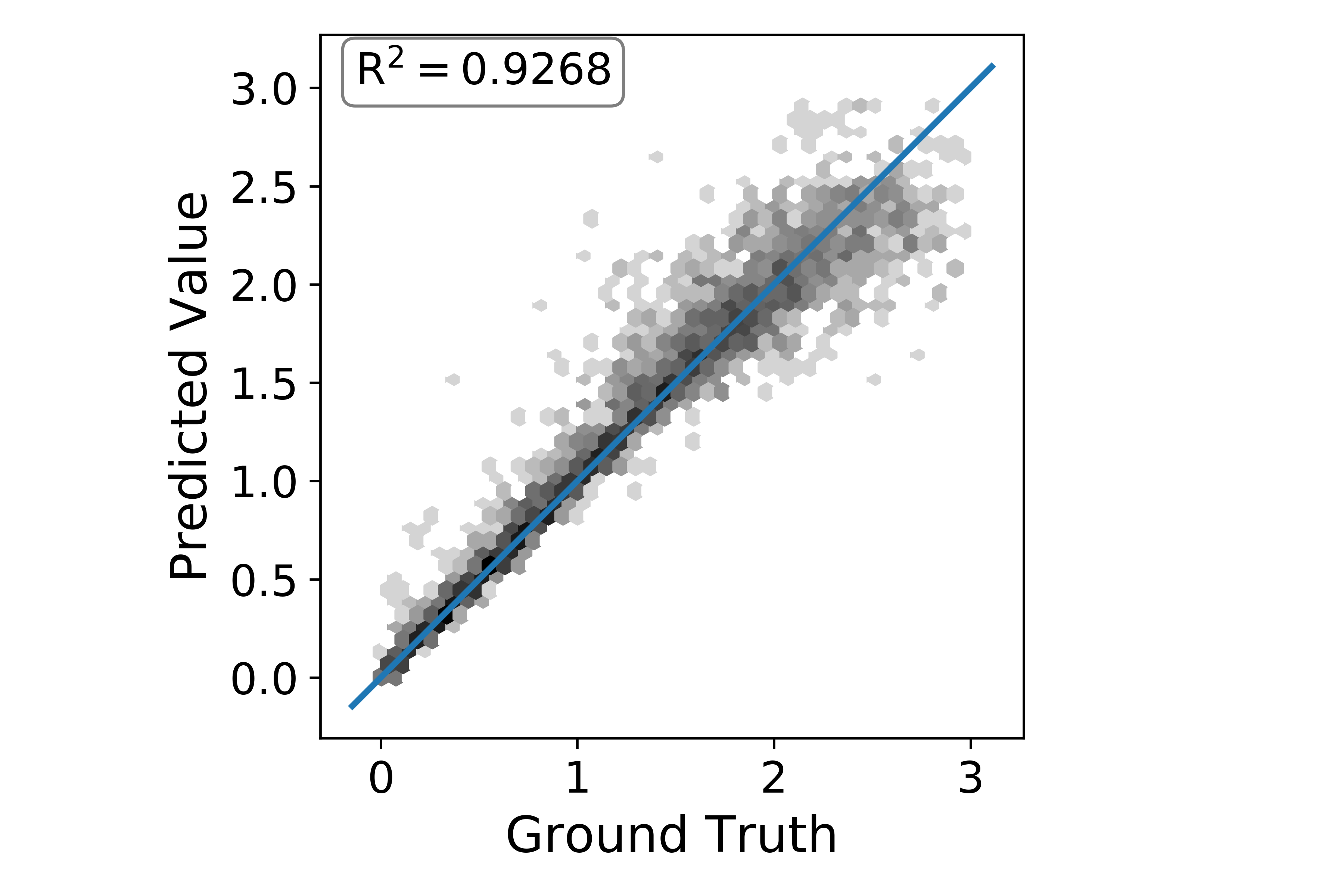}
 	\end{subfigure}
 	\begin{subfigure}[t]{0.32\linewidth}
 		\centering	
 		\includegraphics[width=1.2\linewidth,trim={1.0cm 0.0cm 1.0cm 0.0cm}, clip] {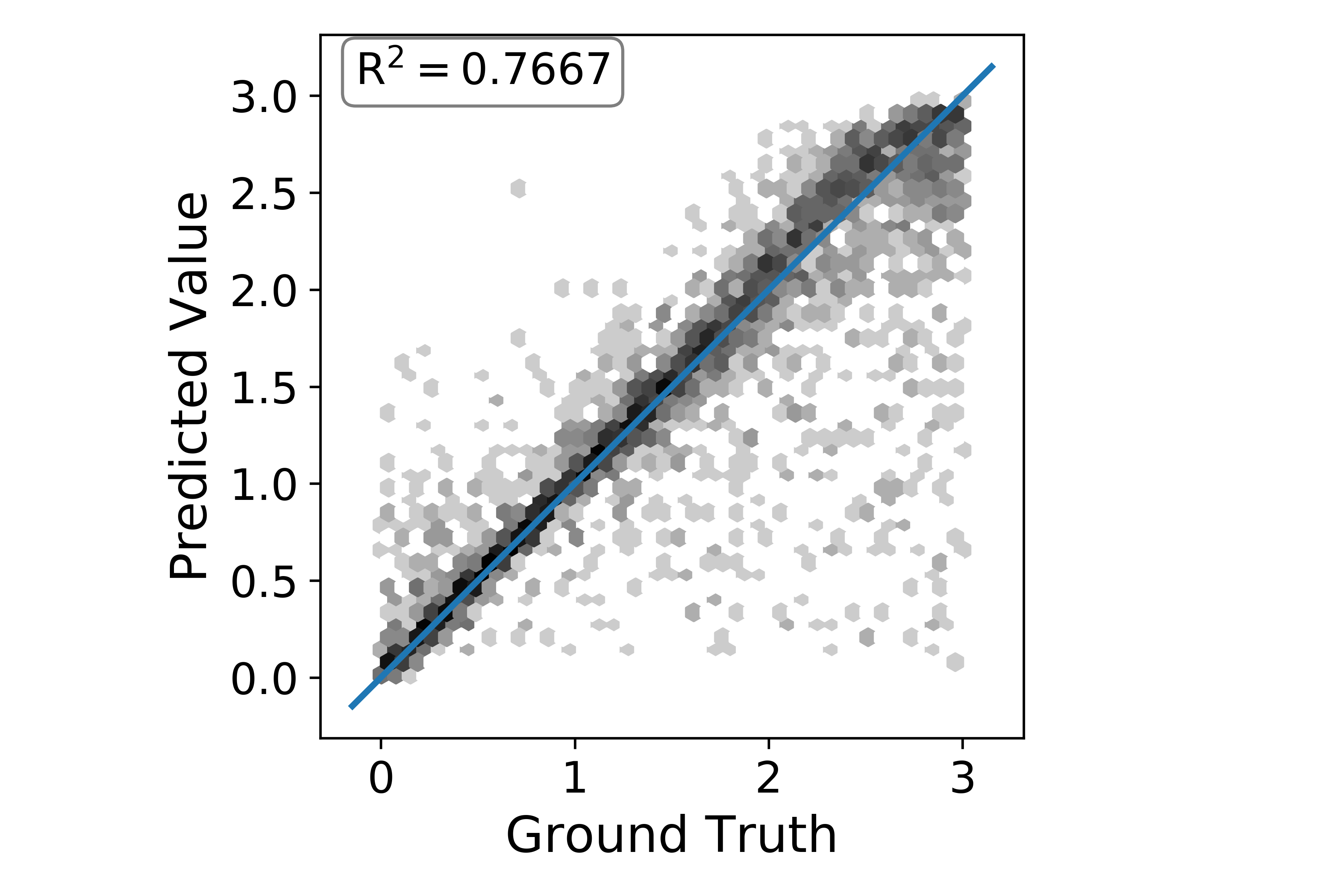}
 	\end{subfigure}
 	\begin{subfigure}[t]{0.32\linewidth}
 		\centering
 		\includegraphics[width=1.2\linewidth,trim={1.0cm 0.0cm 1.0cm 0.0cm},clip]{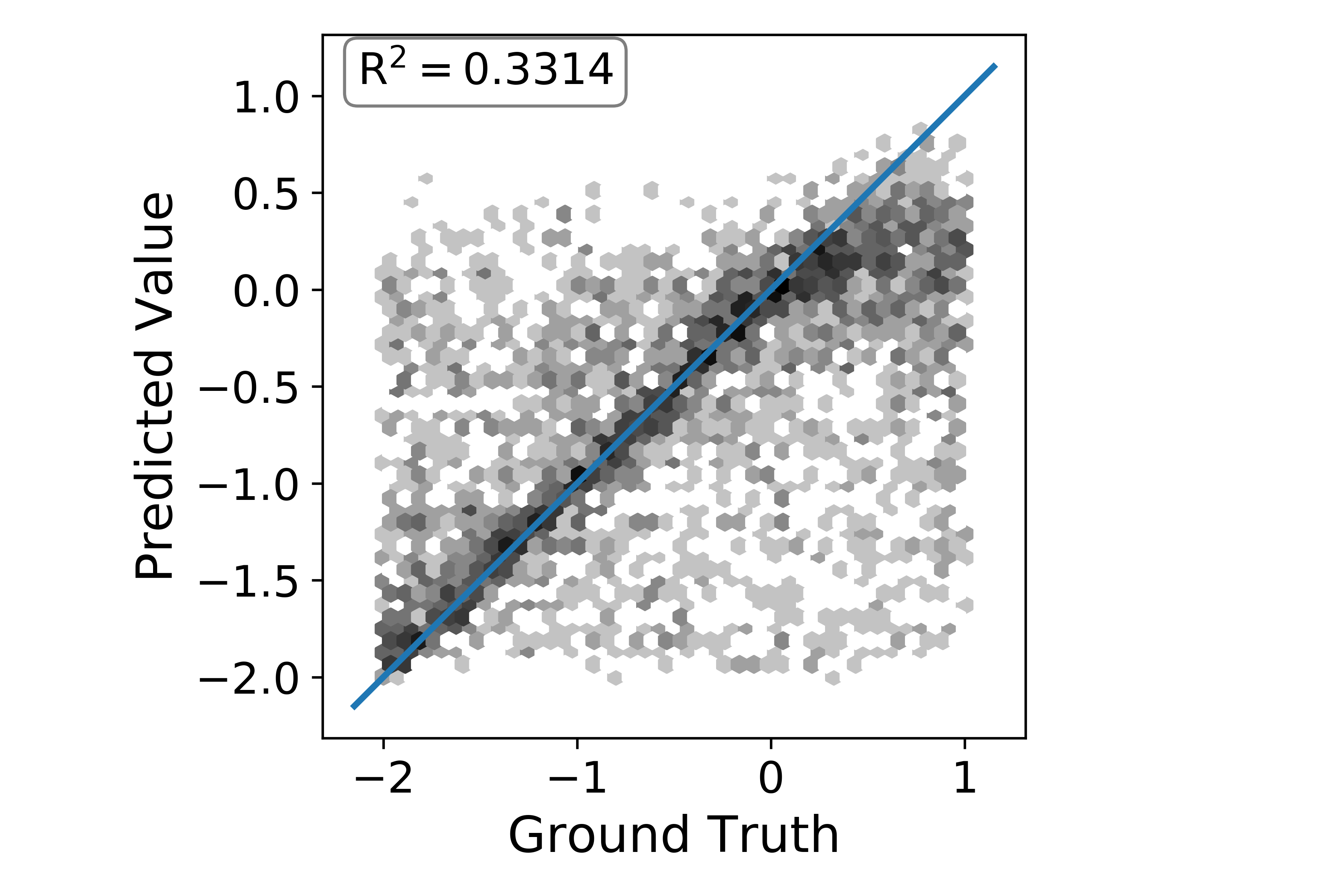}
 	\end{subfigure}
 \medskip
 \hspace{0.5cm}\underline{\textbf{Final Iteration}}\\
 \medskip
 \begin{subfigure}[t]{0.32\linewidth}
 	\centering
 	\includegraphics[width=1.2\linewidth,trim={1.0cm 0.0cm 1.0cm 0.0cm},clip]{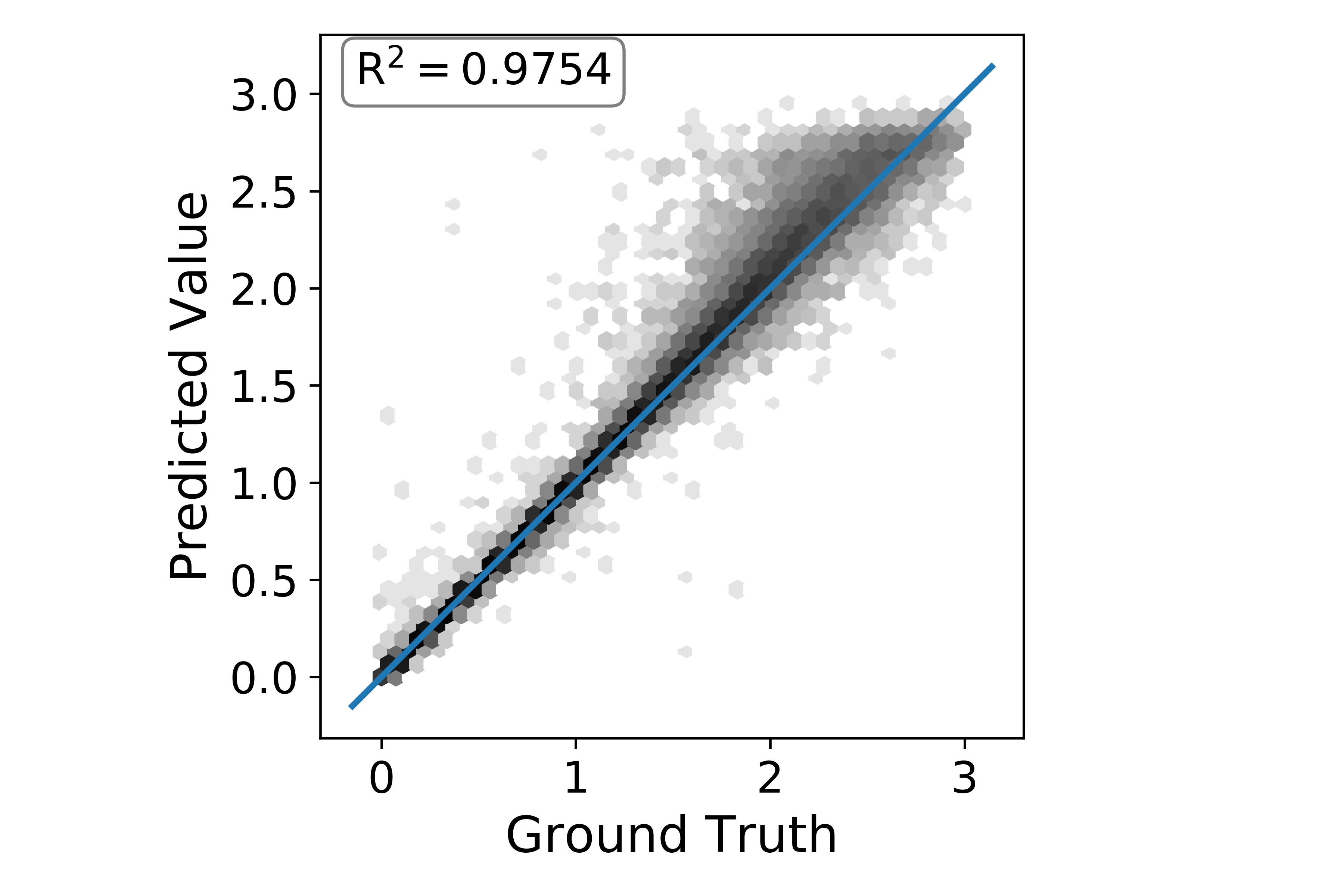}
 \end{subfigure}
 \begin{subfigure}[t]{0.32\linewidth}
 	\centering	
 	\includegraphics[width=1.2\linewidth,trim={1.0cm 0.0cm 1.0cm 0.0cm}, clip] {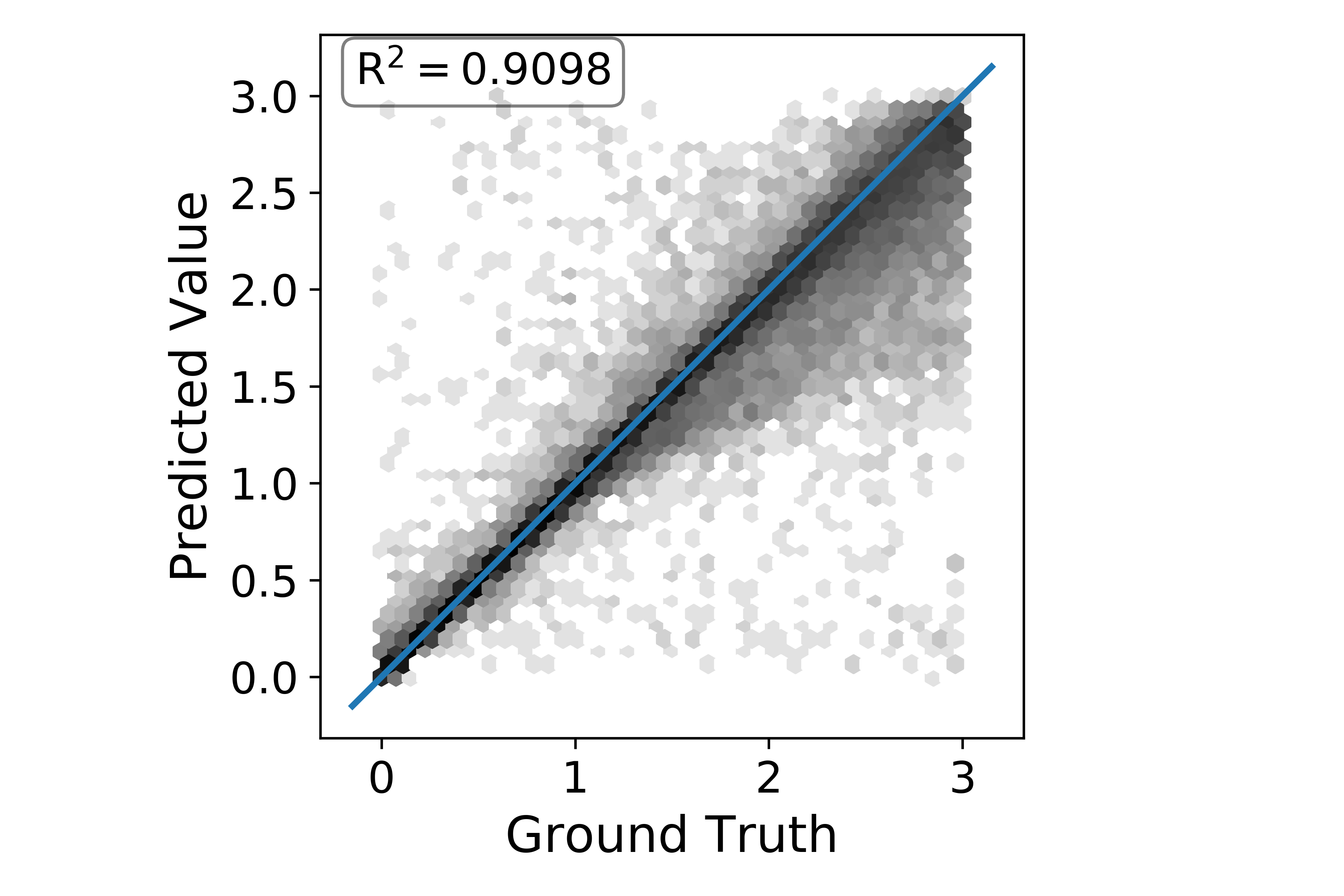}
 \end{subfigure}
 \begin{subfigure}[t]{0.32\linewidth}
 	\centering
 	\includegraphics[width=1.2\linewidth,trim={1.0cm 0.0cm 1.0cm 0.0cm},clip]{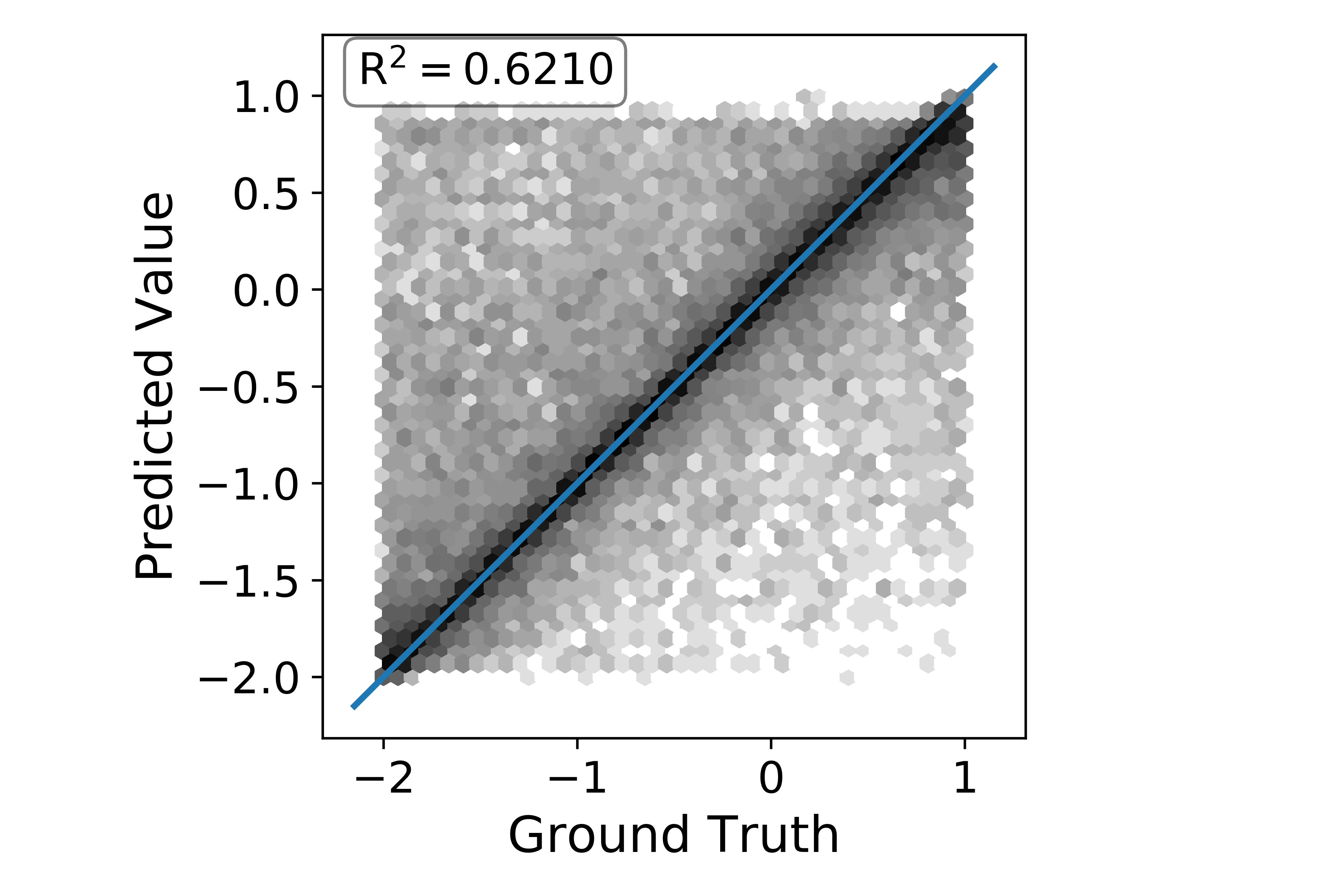}
 \end{subfigure}
 	\caption{ Predicted value vs ground truth for three subsurface material properties: horizontal resistivity $\rho_h$, resistivity of layer located above the current layer $\rho_u$, and vertical distance from the current logging position to the upper boundary $d_u$  for different iterations.}
 	\label{fig:cross-plot22}
 \end{figure}

We observe the predicted values for subsurface material properties for each iteration in the corresponding row of \Cref{fig:cross-plot21,fig:cross-plot22}. We select three properties $\rho_h,~\rho_u,~d_u$ for illustration. The poor correlation in the first iteration hints at the inadequacy of a single measurement acquisition system. 
\paragraph{\it{\bf {Iteration 2.}}} In this iteration, the dataset with the worst correlated measurement from the previous iteration, i.e., the symmetrized directional measurement with short-spaced deep azimuthal instrument, is selected. After training, the algorithm finds that the worst correlated measurement corresponds to the symmetrized directional measurement with short-spaced LWD instrument.
\paragraph{\it{\bf {Iteration 3.}}} 
The algorithm selects the symmetrized directional measurement with the short-spaced LWD instrument and train the DNN in this iteration. Using this trained DNN, the worst correlated measurement corresponds to the geosignal measurement with the short-spaced LWD instrument. 
\paragraph{\it{\bf {Iteration 4.}}} In this iteration, the algorithm adds the geosignal measurement with the short-spaced LWD instrument and trains the DNN. Using this trained DNN, the worst correlated measurement corresponds to the coaxial measurement with the long-spaced deep azimuthal instrument.
\begin{figure}[ht!]
	\centering 
	
	\underline{\textbf{Iteration \num{3}}}
	
	\medskip
	\begin{subfigure}[t]{0.48\linewidth}
		\centering
		\subcaptionbox*{Symmetrized directional measurement with the short-spaced LWD instrument} {\includegraphics [width=1.0\linewidth, trim={1.0cm 0.0cm 1.0cm 0.0cm}, clip] {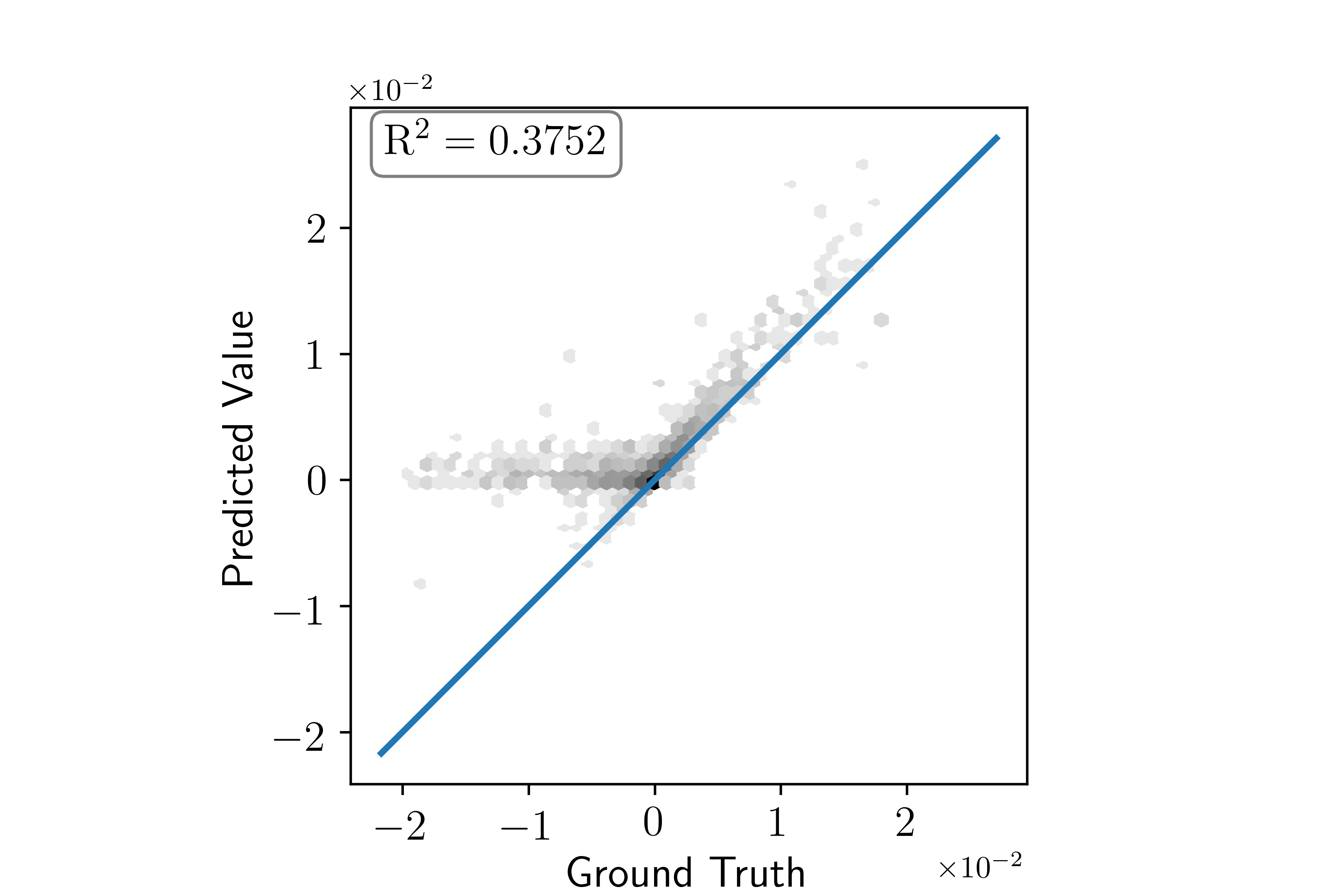} }
	\end{subfigure}
	\hfill
	\begin{subfigure}[t]{0.48\linewidth}
		\centering
		\subcaptionbox*{Geosignal measurement with the short-spaced LWD instrument} {\includegraphics [width=1.0\linewidth, trim={1.0cm 0.0cm 1.0cm 0.0cm}, clip] {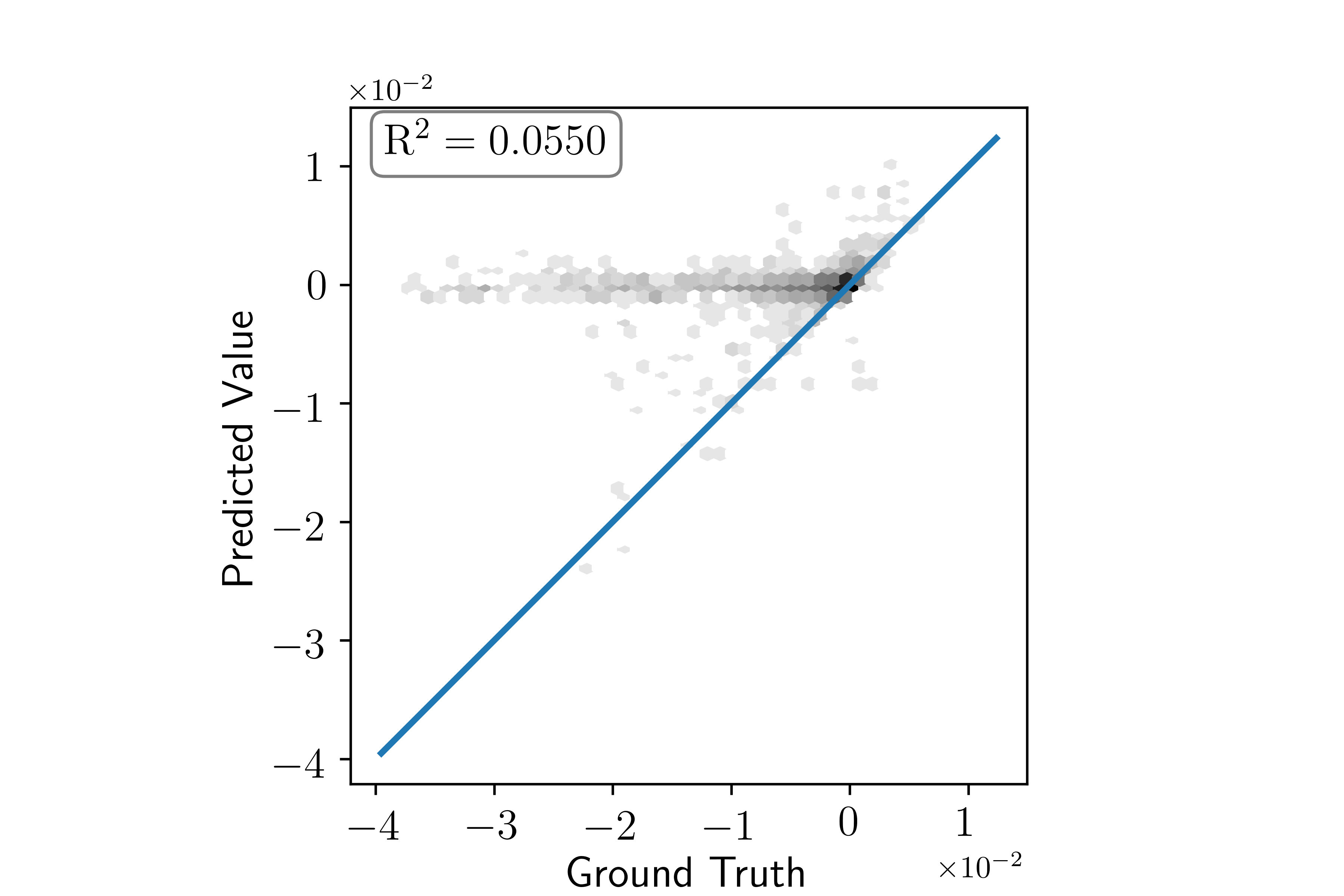}}
	\end{subfigure}
\\
	\medskip
	
	\underline{\textbf{Iteration \num{4}}}
	
	\medskip
	\begin{subfigure}[t]{0.48\linewidth}
		\centering
		\subcaptionbox*{Geosignal measurement with the short-spaced LWD instrument} {\includegraphics [width=1.0\linewidth, trim={1.0cm 0.0cm 1.0cm 0.0cm}, clip] {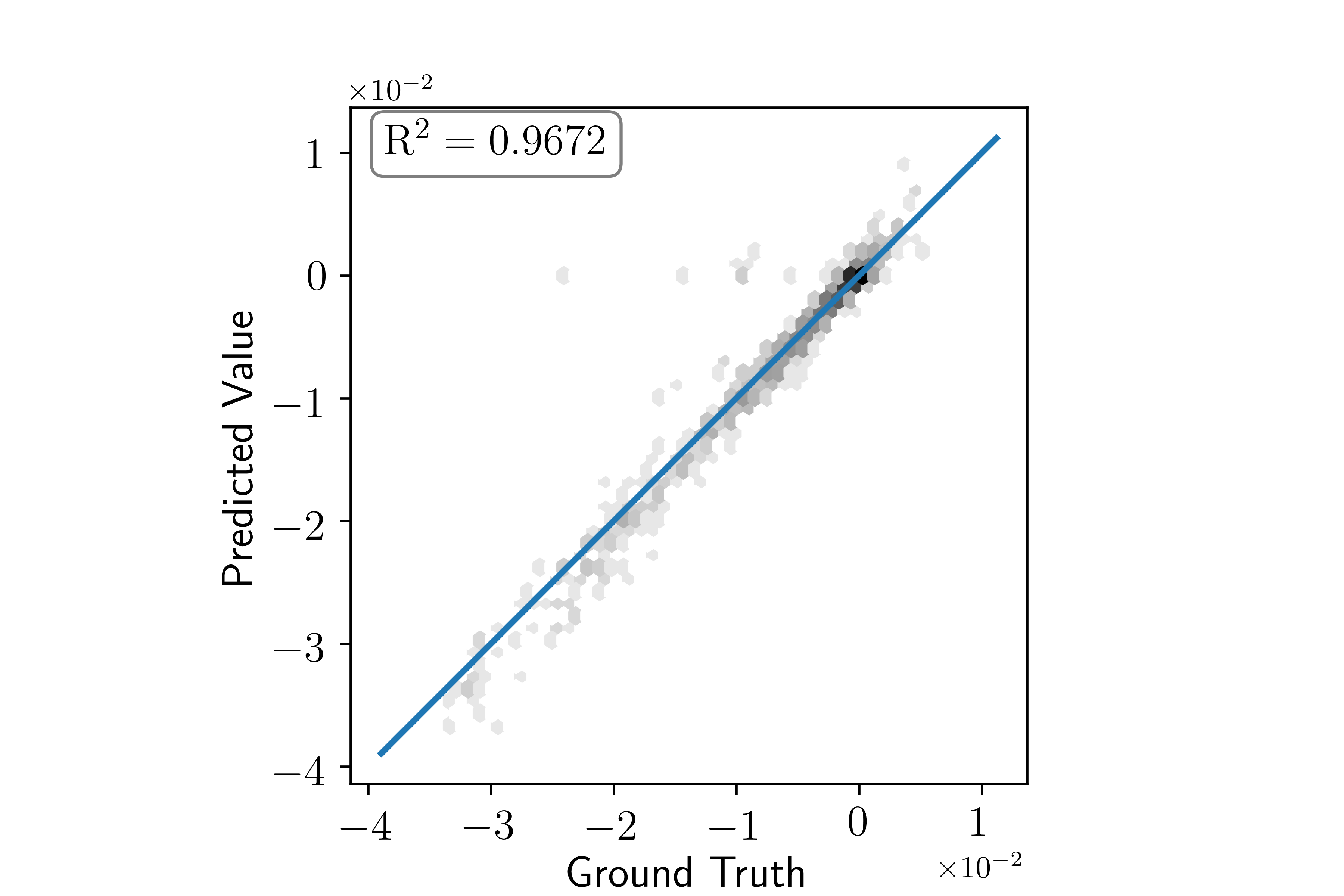} }
	\end{subfigure}
	\hfill
	\begin{subfigure}[t]{0.48\linewidth}
		\centering
		\subcaptionbox*{Coaxial measurement with the long-spaced deep azimuthal instrument} {\includegraphics [width=1.0\linewidth, trim={1.0cm 0.0cm 1.0cm 0.0cm}, clip] {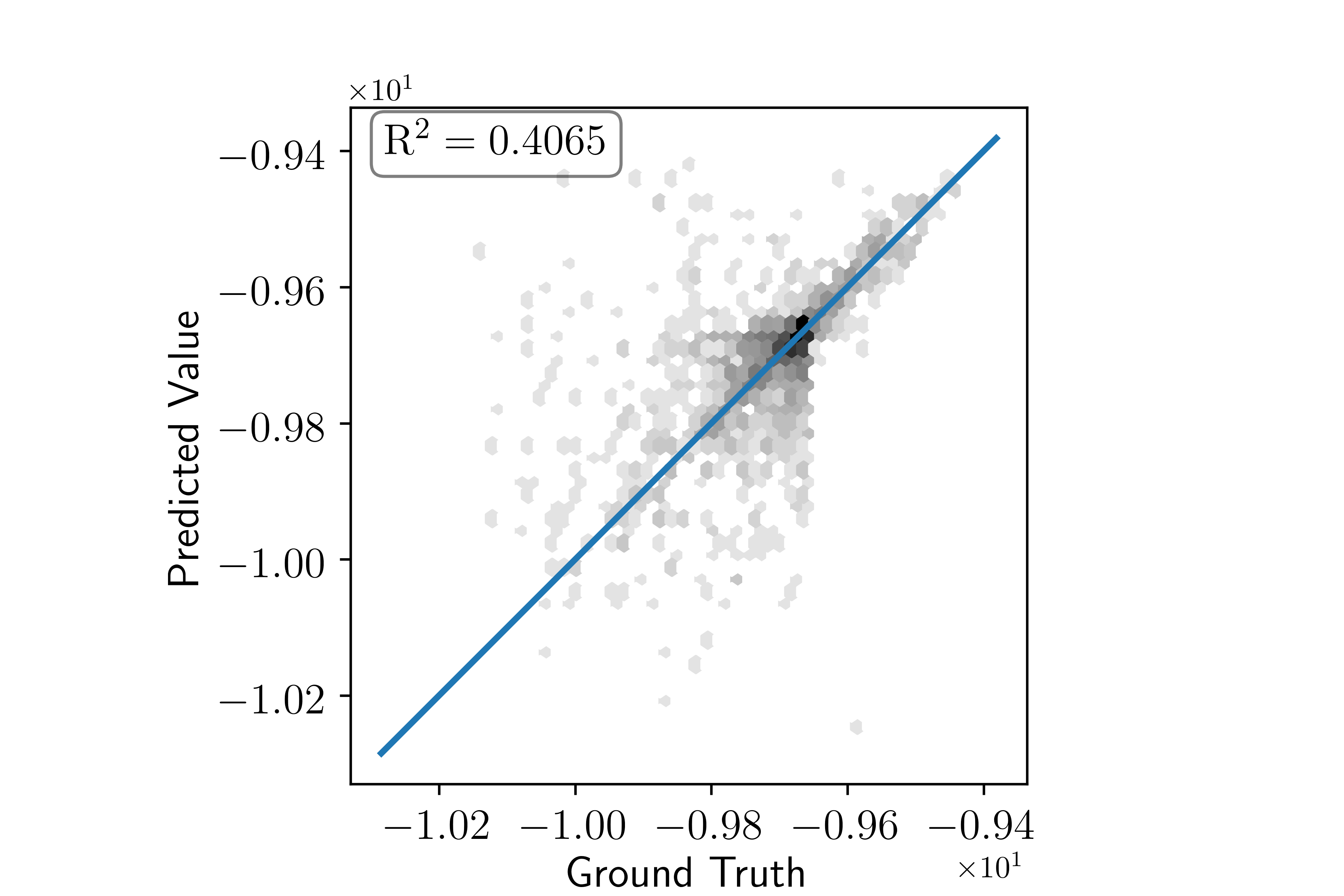}}
	\end{subfigure}
	\caption{Cross-plots of type 1. Predicted value vs. ground truth for attenuation--iterations \numrange[range-phrase= -- ]{3}{4}. First column corresponds to the last added measurement and second column is associated with the measurement with the worst $R^2$ value (after averaging the $R^2$ values of attenuation and phase).}
	\label{fig:cross-plots12}
\end{figure}

\paragraph{\it{\bf {Iteration 5.}}} The algorithm selects the coaxial measurement with the long-spaced deep azimuthal instrument for further training the DNN. We find that the worst averaged correlation corresponds to the symmetrized directional measurement with the long-spaced LWD instrunent.

\begin{figure}[ht!]
	\centering 
	
	\underline{\textbf{Iteration \num{5}}}
	
	\medskip
	\begin{subfigure}[t]{0.48\linewidth}
		\centering
		\subcaptionbox*{Coaxial measurement with the long-spaced deep azimuthal instrument} {\includegraphics [width=1.0\linewidth, trim={1.0cm 0.0cm 1.0cm 0.0cm}, clip] {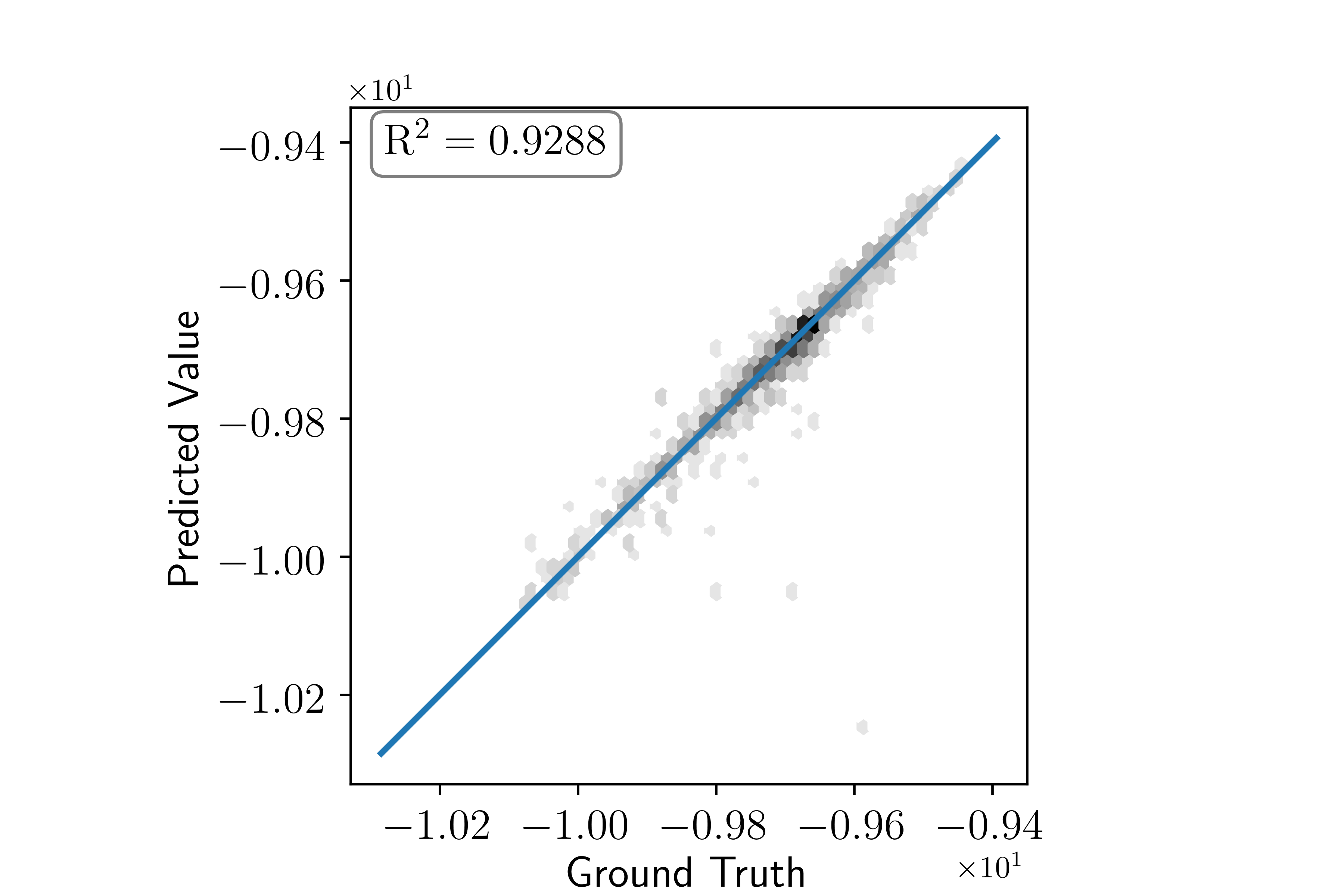} }
	\end{subfigure}
	\hfill
	\begin{subfigure}[t]{0.48\linewidth}
		\centering
		\subcaptionbox*{Symmterized directional measurement with the long-spaced LWD instrument} {\includegraphics [width=1.0\linewidth, trim={1.0cm 0.0cm 1.0cm 0.0cm}, clip] {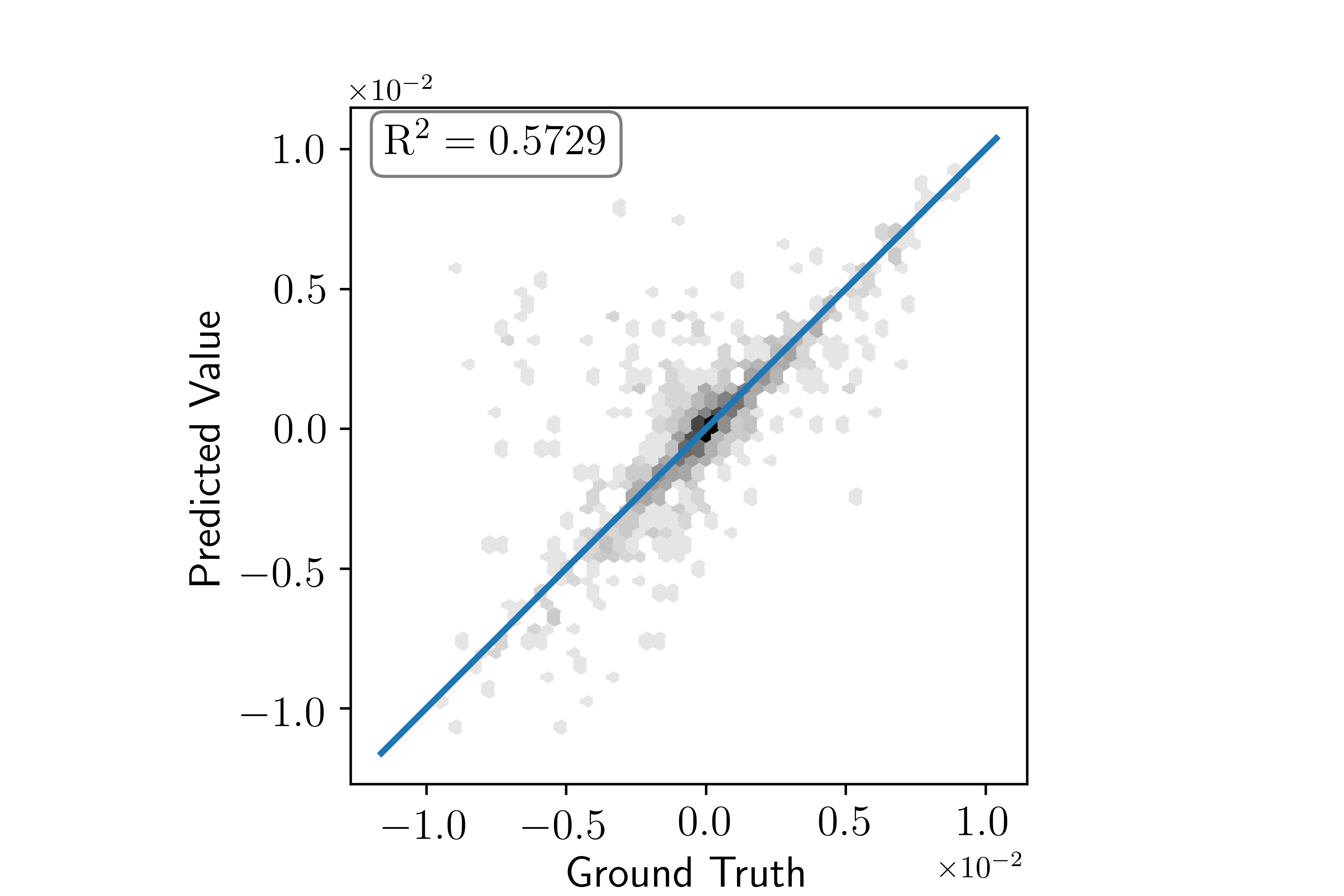}}
	\end{subfigure}
	\\
	
	\medskip
	
	\underline{\textbf{Iteration \num{6}}}
	
	\medskip
	\begin{subfigure}[t]{0.48\linewidth}
		\centering
		\subcaptionbox*{Symmterized directional measurement with the long-spaced LWD instrument}{\includegraphics [width=1.0\linewidth, trim={1.0cm 0.0cm 1.0cm 0.0cm}, clip] {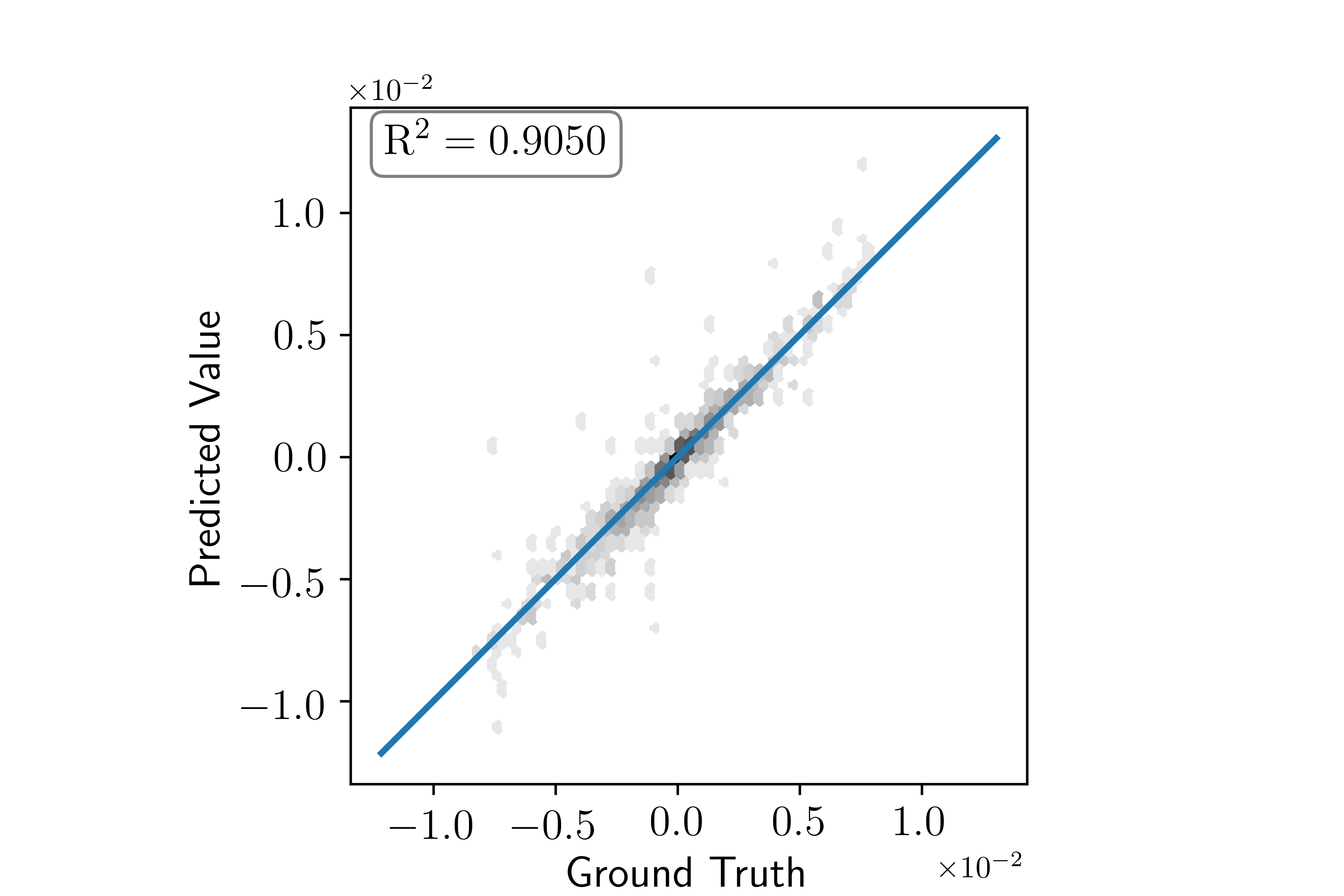}}
	\end{subfigure}
	\hfill
	\begin{subfigure}[t]{0.48\linewidth}
		\centering
		\subcaptionbox*{yy-measurement with the short-spaced LWD instrument} {\includegraphics [width=1.0\linewidth, trim={1.0cm 0.0cm 1.0cm 0.0cm}, clip] {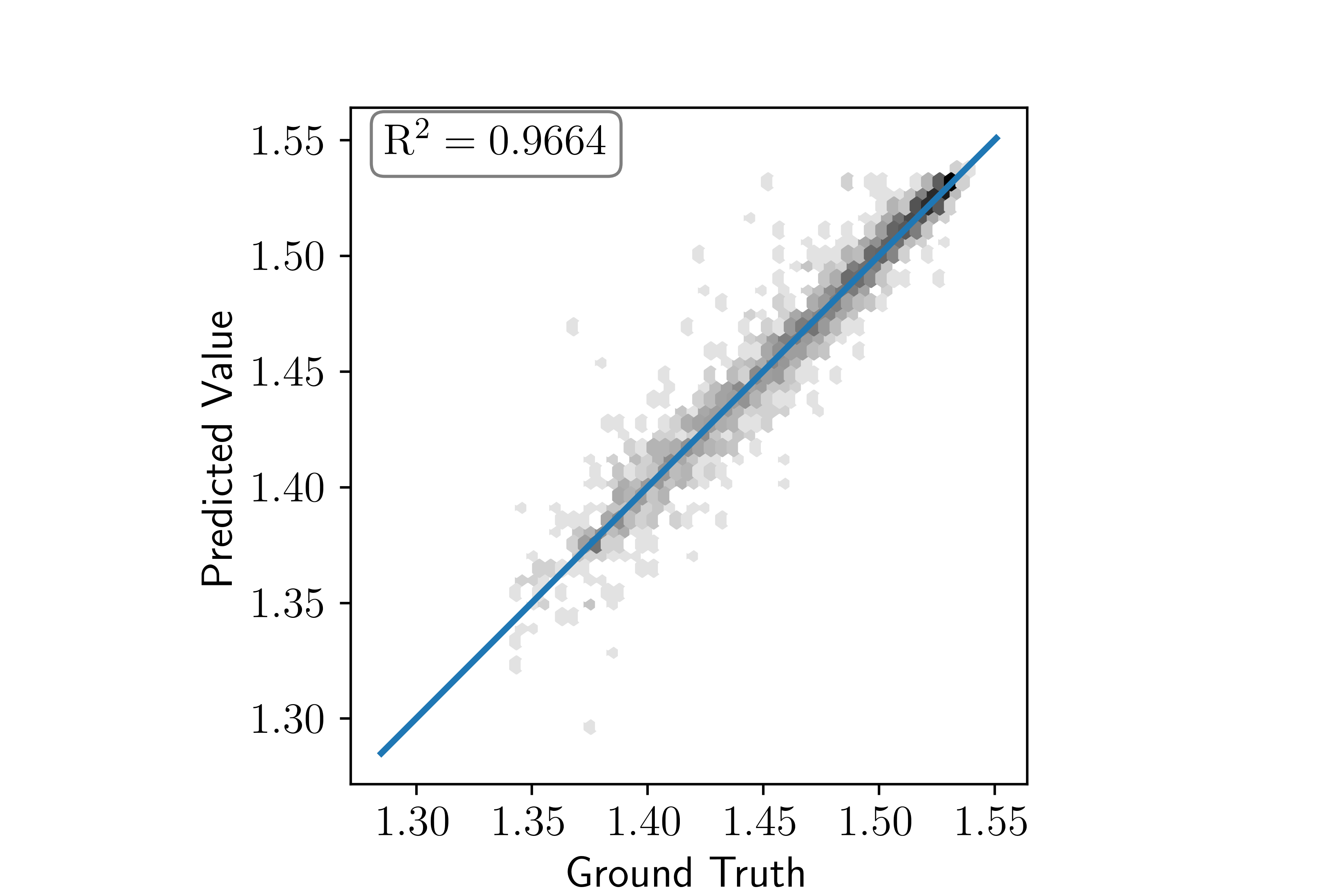} }
	\end{subfigure}
	\caption{Cross-plots of type 1. Predicted value vs. ground truth for attenuation--iterations \numrange[range-phrase= -- ]{5}{6}. First column corresponds to the last added measurement and second column is associated with the measurement with the worst $R^2$ value (after averaging the $R^2$ values of attenuation and phase).}
	\label{fig:cross-plots13}
\end{figure}

\paragraph{\it{\bf {Iteration 6.}}} 
In this iteration, the dataset with the symmetrized directional measurement using the long-spaced LWD instrument is added for further training. Using this trained DNN, we find that the worst averaged correlation corresponds to the yy-measurement using the short-spaced LWD instrument. Although the $R^2$ value for attenuation is high for the yy-measurement using the short-spaced LWD instrument (right panel of the third row of \Cref{fig:cross-plots13}), low $R^2$ value of phase difference makes this measurement the worst correlated one.

\paragraph{\it{\bf {Iteration 7.}}} 
In this iteration, the yy-measurement with the short-spaced LWD instrument is added for further training. Using this trained DNN, we find that the worst averaged correlation corresponds to the xxyyzz-measurement using the short-spaced deep azimuthal instrument. The correlation improves over the previous iteration and the average $R^2$ value of the worst correlation is above \num{0.8}. At this point, we stop adding measurements and fix our final measurement acquisition system.

\paragraph{\it{\bf {Iteration 8.}}} 

In this iteration, we retrain our DNN using our last upgraded acquisition system with the entire dataset, and we observe that the average $R^2$ value of the worst correlation is above \num{0.9}.
\begin{figure}[ht!]
	\centering 
	
	\medskip
	
	\underline{\textbf{Iteration \num{7}}}
	
	\medskip
	\begin{subfigure}[t]{0.48\linewidth}
		\centering
		\subcaptionbox*{yy-measurement with the short-spaced LWD instrument}{\includegraphics [width=1.0\linewidth, trim={1.0cm 0.0cm 1.0cm 0.0cm}, clip] {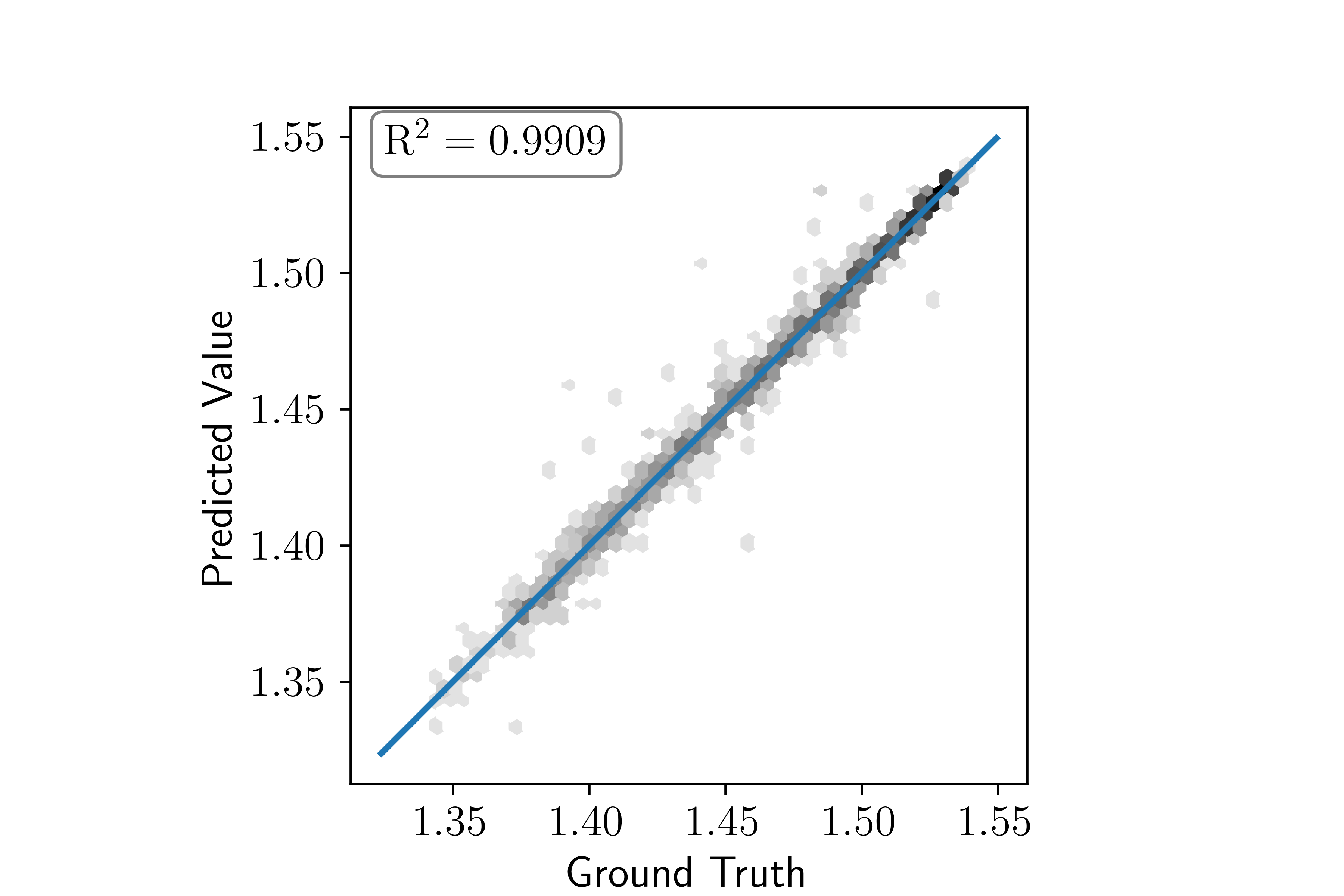}}
	\end{subfigure}
	\hfill
	\begin{subfigure}[t]{0.48\linewidth}
		\centering
		\subcaptionbox*{xxyyzz-measurement with the short-spaced deep azimuthal instrument} {\includegraphics [width=1.0\linewidth, trim={1.0cm 0.0cm 1.0cm 0.0cm}, clip] {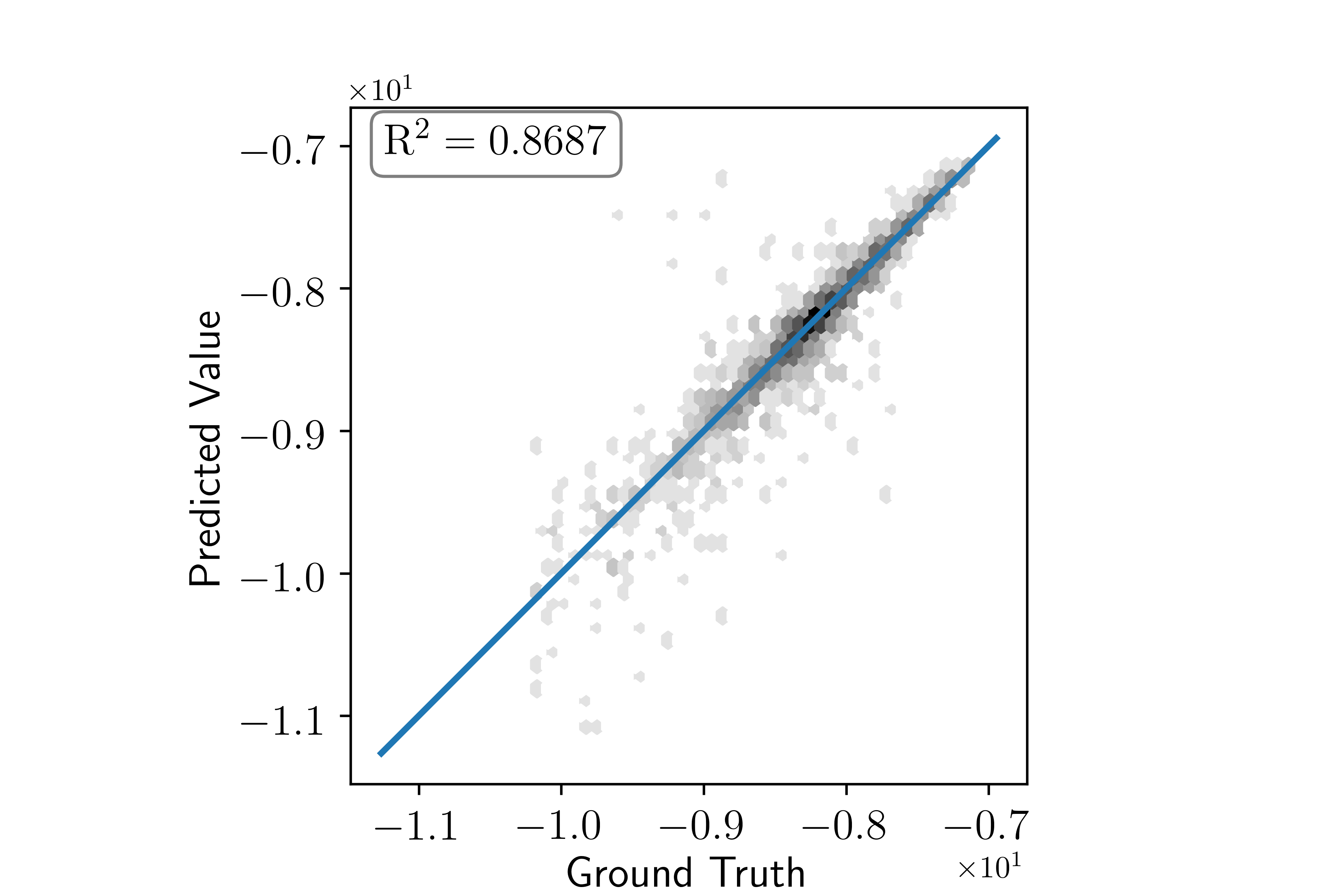} }
	\end{subfigure}
	\\
	
	\medskip
	
	\underline{\textbf{Final Iteration}}
	
	\medskip
	\begin{subfigure}[t]{0.48\linewidth}
		\centering
		\subcaptionbox*{yy-measurement with the short-spaced LWD instrument}{\includegraphics [width=1.0\linewidth, trim={1.0cm 0.0cm 1.0cm 0.0cm}, clip] {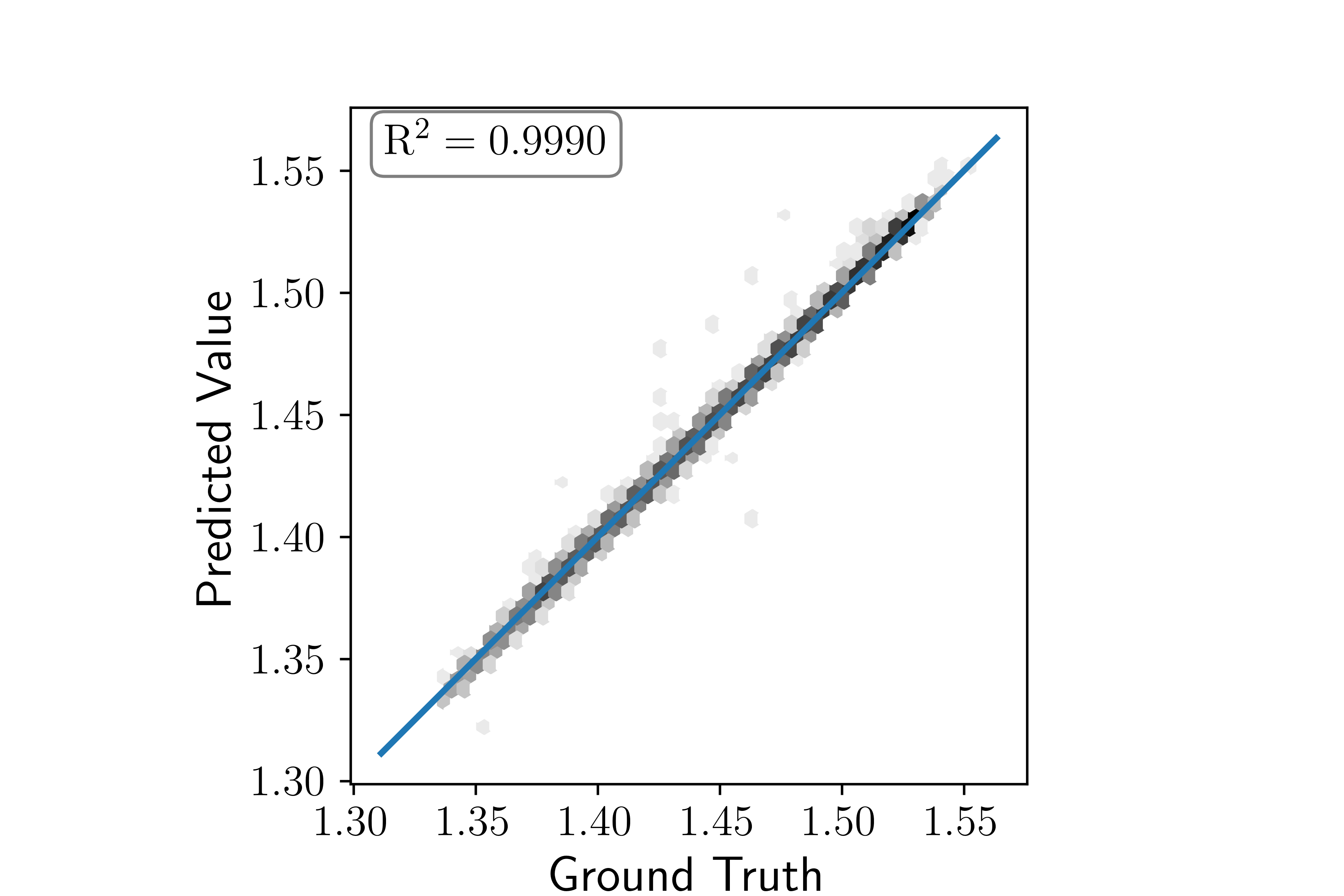}}
	\end{subfigure}
	\hfill
	\begin{subfigure}[t]{0.48\linewidth}
		\centering
		\subcaptionbox*{xxyyzz-measurement with the short-spaced deep azimuthal instrument} {\includegraphics [width=1.0\linewidth, trim={1.0cm 0.0cm 1.0cm 0.0cm}, clip] {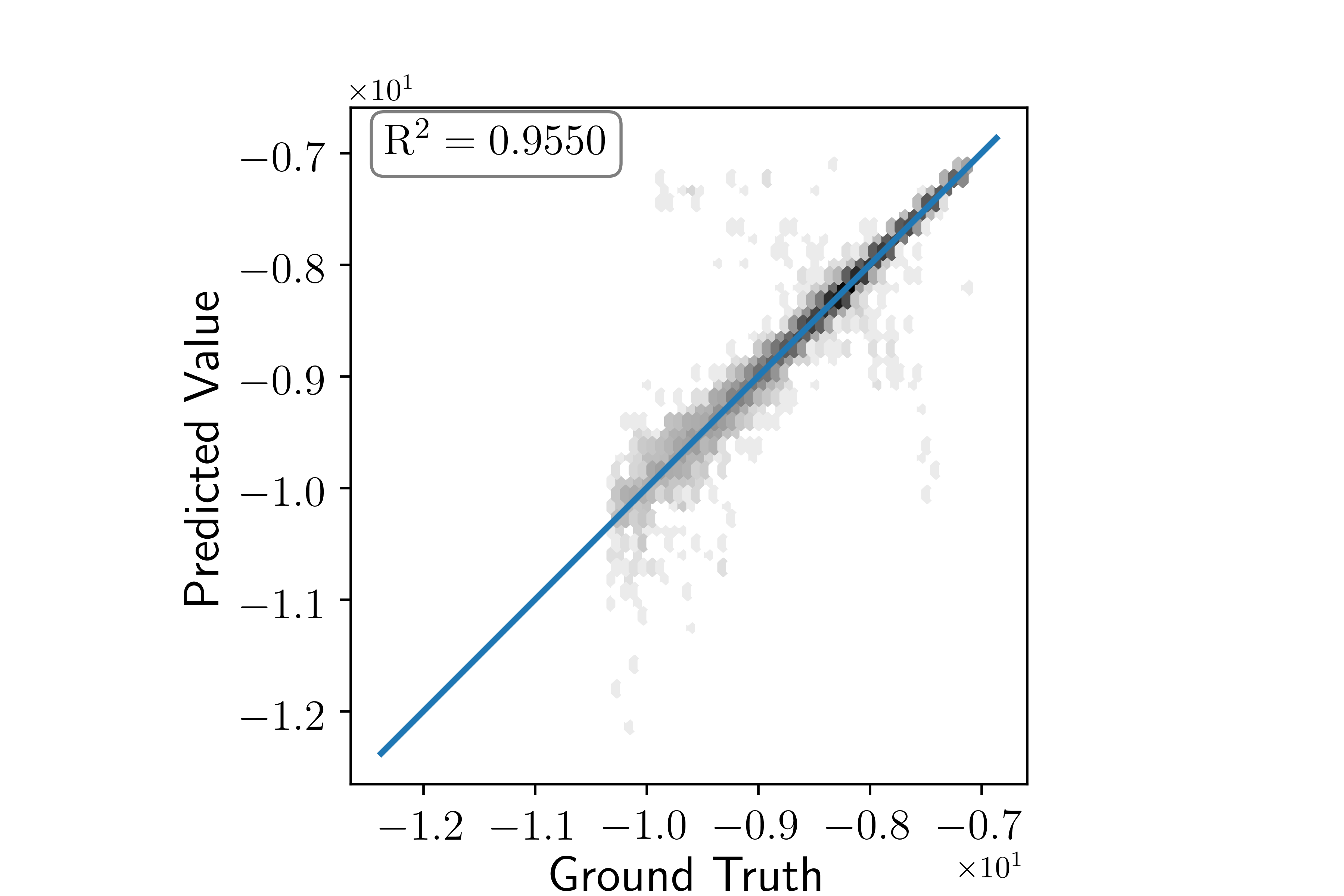} }
	\end{subfigure}
	\caption{Cross-plots of type 1. Predicted value vs. ground truth of attenuation for iteration \num{7} and the final. First column corresponds to the last added measurement and second column is associated with the measurement with the worst $R^2$ value (after averaging the $R^2$ values of attenuation and phase).}
	\label{fig:cross-plots14}
\end{figure}

\Cref{fig:cross-plot21,fig:cross-plot22} compare the ground truth vs. predicted values of the inversion variables and show the improvements of predictions over different iterations. The bottom row in \Cref{fig:cross-plot22} shows the final comparison between estimated and real values of subsurface properties. We observe from different rows in \Cref{fig:cross-plot21,fig:cross-plot22} that although the resistivity predictions improve rapidly, the distance prediction $d_u$ takes a larger number of iterations to reach a high correlation with the ground truth. This is consistent with the physics of the problem: it is more challenging to determine bed boundary locations than resistivities.

\vspace{-3.5mm}

\subsection{Inversion of realistic synthetic model}
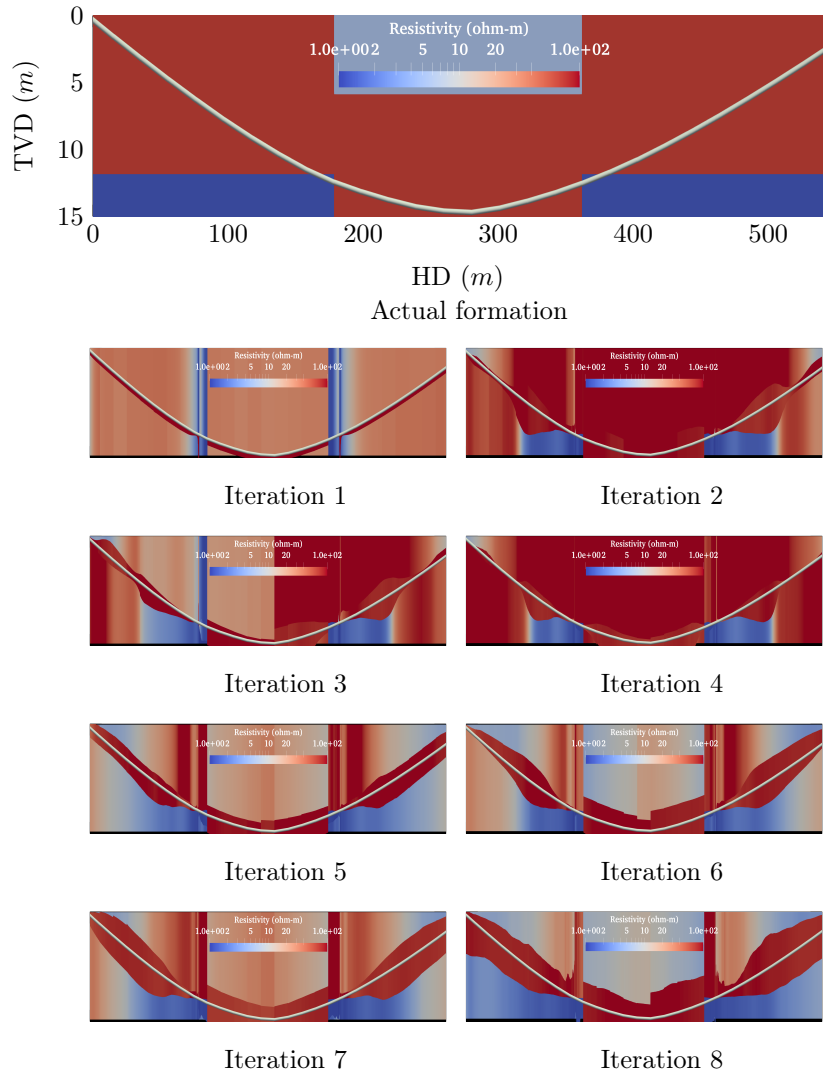
\begin{figure*}[ht]
	\centering
	\begin{tikzpicture}
	\node at (2.05,27.5)[scale=1]{\input{Figures/Syn1/Real/real_v1.tex}};
	\node at (0,24.0)[scale=1]{\input{Figures/Syn1/exp0/exp0_v1.tex}};
	\node at (5,24.0)[scale=1]{\input{Figures/Syn1/exp1/exp1_v1.tex}};
	\node at (0,21.5)[scale=1]{\input{Figures/Syn1/exp2/exp2_v1.tex}};
	\node at (5,21.5)[scale=1]{\input{Figures/Syn1/exp3/exp3_v1.tex}};
	\node at (0,19.0)[scale=1]{\input{Figures/Syn1/exp4/exp4_v1.tex}};
	\node at (5,19.0)[scale=1]{\input{Figures/Syn1/exp5/exp5_v1.tex}};
	\node at (0,16.5)[scale=1]{\input{Figures/Syn1/exp6/exp6_v1.tex}};
	\node at (5,16.5)[scale=1]{\input{Figures/Syn1/exp7/exp7_v1.tex}};
	\end{tikzpicture}
	\caption{Model problem 1. The evolution of the inverted formation at each iteration compared to the actual formation. The gray line indicates the well trajectory.}
	\label{fig:syn}
\end{figure*}

We assess the performance of the inversion process with two realistic synthetic test cases:

\paragraph{Model problem 1.} We create a model with realistic geological features as depicted on the top panel of \Cref{fig:syn}. The model has a resistive layer with a water-saturated layer underneath in the bottom left and right and it exhibits two geological faults. The trajectory dip angle varies from \ang{82} to \ang{96}. 
	
	
	In subsequent rows of \Cref{fig:syn},  we observe the improvements in predictions with different iterations. We conclude that the final prediction with the final upgraded measurement acquisition system and larger dataset can properly predict the formation resisitivity at a distance of up to \SIrange[range-phrase=--,range-units=single]{5}{7}{\m} from the logging instrument.

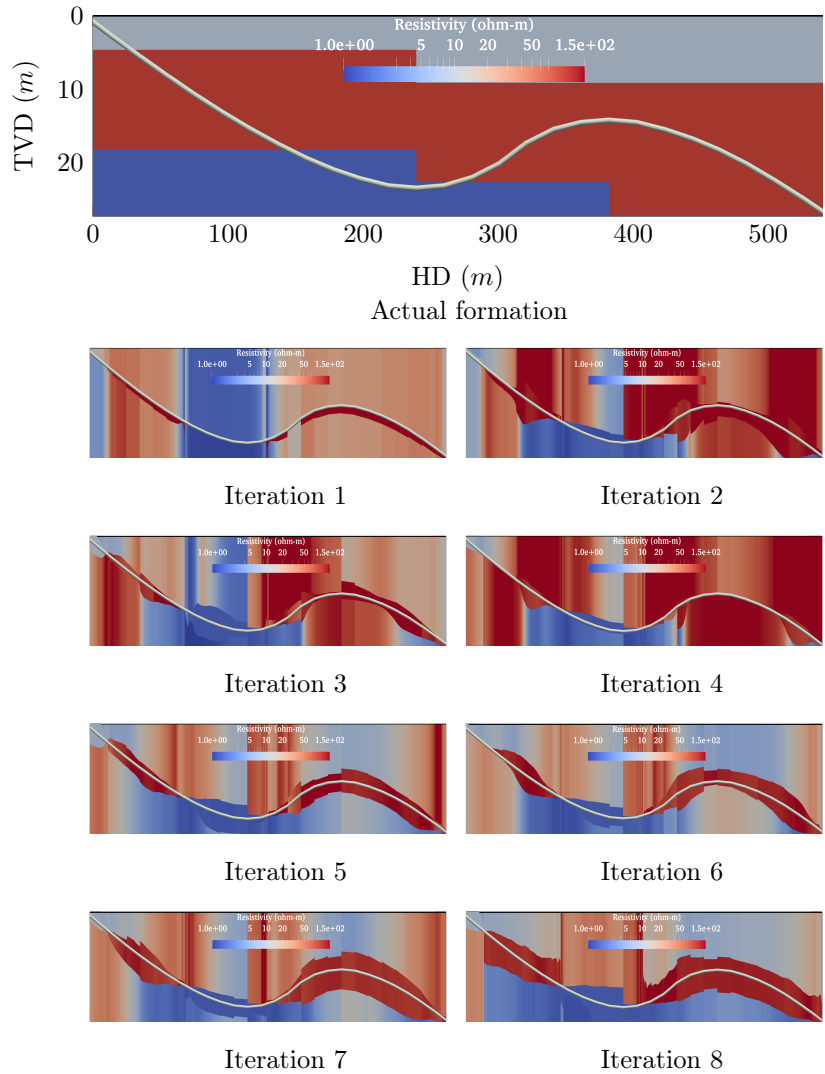
\begin{figure*}[ht]
	\centering
	\begin{tikzpicture}
	\node at (2.05,27.5)[scale=1]{\input{Figures/Syn2/Real/real_v1.tex}};
	\node at (0,24.0)[scale=1]{\input{Figures/Syn2/exp0/exp0_v1.tex}};
	\node at (5,24.0)[scale=1]{\input{Figures/Syn2/exp1/exp1_v1.tex}};
	\node at (0,21.5)[scale=1]{\input{Figures/Syn2/exp2/exp2_v1.tex}};
	\node at (5,21.5)[scale=1]{\input{Figures/Syn2/exp3/exp3_v1.tex}};
	\node at (0,19.0)[scale=1]{\input{Figures/Syn2/exp4/exp4_v1.tex}};
	\node at (5,19.0)[scale=1]{\input{Figures/syn2/exp5/exp5_v1.tex}};
	\node at (0,16.5)[scale=1]{\input{Figures/Syn2/exp6/exp6_v1.tex}};
	\node at (5,16.5)[scale=1]{\input{Figures/Syn2/exp7/exp7_v1.tex}};
	\end{tikzpicture}
	\caption{Model problem 2. The evolution of the inverted formation at each iteration compared to the actual formation. The gray line indicates the well trajectory.}
	\label{fig:syn2}
\end{figure*}

\paragraph{Model problem 2.} We consider a geological formation with a conductive layer surrounded by two different resistive layers on top and bottom. The top panel of \Cref{fig:syn2} displays the original model. The trajectory dip angle varies from \ang{82} to \ang{96}.

We conclude that the final prediction with the final upgraded measurement acquisition system can predict the targeted resistive layer from a few meters distance.

\section{Discussion and Conclusion} \label{sec:discon}
This paper focuses on the use of deep neural networks (DNNs) for the
inversion of borehole resistivity measurements for geosteering applications. 
Specifically, we propose an iterative method for designing a measurement acquisition system and illustrate the effectiveness of the method with several benchmark examples.

Prior work \cite{Shahriari_loss} shows that a loss function based on data misfit is unsuitable for inversion due to the non-uniqueness of the inverse problem. As a remedy to that, it is possible to use a loss function specifically designed for \textit{encoder-decoder} architectures with \textit{regularization}. However, regularization terms hide some of the possibly existing solutions of the inverse problem. Here, we avoid regularization and employ a \textit{two-step} loss function.

We then design a measurement acquisition system. We start with a single measurement and iteratively select a minimal set of measurements from a large set. By analyzing the inversion of borehole resistivity measurements, we encounter that seven carefully selected measurements is enough to uniquely determine the earth subsurface with the selected parameterization. Numerical results show that the resulting measurement acquisition system is sufficient to identify both resistive and conductive layers above and below the logging instrument. It also allows to anticipate the presence of a water-bearing layer a few vertical meters in advance, as well as to provide a proper distance estimation from that layer to the logging instrument.

Future work includes: 
\begin{enumerate*} [label=(\alph*), itemjoin={{, }}, itemjoin*={{, and }}]
\item to consider noisy measurements
\item to use more general earth subsurface parameterizations 
\item to consider statistics other than $R^2$ \cite{kvaalseth1985cautionary} for driving the algorithm that selects an optimal measurement acquisition system
\item to use transfer learning for inversion of 2D and 3D geometries
\item to automate different tasks in the ML pipeline using AutoML \cite{he2021automl, hutter2019automated} techniques for data processing, DNN architecture optimization, feature selection and hyperparameter tuning. 

\end{enumerate*}

\section*{Acknowledgments}
Mostafa Shahriari has been supported by the Austrian Ministry for Transport, Innovation and Technology (BMVIT), the Federal Ministry for Digital and Economic Affairs (BMDW), the Province of Upper Austria in the frame of the COMET - Competence Centers for Excellent Technologies Program managed by Austrian Research Promotion Agency FFG, the COMET Module S3AI. 

David Pardo has received funding from: the European Union's Horizon 2020 research and innovation program under the Marie Sklodowska-Curie grant agreement No 777778 (MATHROCKS); the European Regional Development Fund (ERDF) through the Interreg V-A Spain-France-Andorra program POCTEFA 2014-2020 Project PIXIL (EFA362/19); the Spanish Ministry of Science and Innovation with references PID2019-108111RB-I00 (FEDER/AEI) and the “BCAM Severo Ochoa” accreditation of excellence (SEV-2017-0718); and the Basque Government through the BERC 2018-2021 program, the two Elkartek projects 3KIA (KK-2020/00049) and MATHEO (KK-2019-00085), the grant "Artificial Intelligence in BCAM number EXP. 2019/00432", and the Consolidated Research Group MATHMODE (IT1294-19) given by the Department of Education. 

 \clearpage
\printbibliography


\end{document}

%% file: Figures/LWD_schem.tex
\begin{tikzpicture}

\fill[gray!80!white] (-1.5*5,0) -- (1.5*5,0.) --  
 (1.5*5, 0.15*3) -- (-1.5*5,0.15*3) -- cycle;


\fill[black!80!white] (-0.4064*5-0.02*3,0) -- (-0.4064*5+0.02*3,0.) node[below] {\footnotesize \bf \textcolor{black}{Tx\textsubscript{1,1}}} -- (-0.4064*5+0.02*3, 0.15*3)  -- (-0.4064*5-0.02*3,0.15*3) -- cycle;

\fill[black!80!white] (0.4064*5-0.02*3,0) -- (0.4064*5+0.02*3,0.) node[below] {\footnotesize \bf \textcolor{black}{Tx\textsubscript{1,2}}}  -- (0.4064*5+0.02*3, 0.15*3)  -- (0.4064*5-0.02*3,0.15*3) -- cycle;

\fill[black!80!white] (-0.8128*5-0.02*3,0) -- (-0.8128*5+0.02*3,0.) -- (-0.8128*5+0.02*3, 0.15*3) node[above] {\footnotesize \bf \textcolor{black}{Tx\textsubscript{2,1}}}  -- (-0.8128*5-0.02*3,0.15*3) -- cycle;
\fill[black!80!white] (0.8128*5-0.02*3,0) -- (0.8128*5+0.02*3,0.)  -- (0.8128*5+0.02*3, 0.15*3) node[above] {\footnotesize \bf \textcolor{black}{Tx\textsubscript{2,2}}} -- (0.8128*5-0.02*3,0.15*3) -- cycle;

\fill[black!80!white] (-0.4064*15-0.02*3,0) -- (-0.4064*15+0.02*3,0.) node[below] {\footnotesize \bf \textcolor{black}{Tx\textsubscript{3,1}}} -- (-0.4064*15+0.02*3, 0.15*3)  -- (-0.4064*15-0.02*3,0.15*3) -- cycle;

\fill[black!80!white] (0.4064*15-0.02*3,0) -- (0.4064*15+0.02*3,0.) node[below] {\footnotesize \bf \textcolor{black}{Tx\textsubscript{3,2}}} -- (0.4064*15+0.02*3, 0.15*3)  -- (0.4064*15-0.02*3,0.15*3) -- cycle;

\fill[red!80!white] (-0.1016*5-0.02*3,0) -- (-0.1016*5+0.02*3,0.)  -- (-0.1016*5+0.02*3, 0.15*3) node[above] {\footnotesize \bf \textcolor{black}{Rx$_1$}} -- (-0.1016*5-0.02*3,0.15*3) -- cycle;
\fill[red!80!white] ( 0.1016*5-0.02*3,0) -- ( 0.1016*5+0.02*3,0.) -- ( 0.1016*5+0.02*3, 0.15*3)  node[above] {\footnotesize \bf \textcolor{black}{Rx$_2$}} -- ( 0.1016*5-0.02*3,0.15*3) -- cycle;

\draw[black, line width=1pt,<->] (-0.1016*5, -0.6)      -- (0.1016*5, -0.6) node[pos=0.5, below] {\footnotesize \bf \textcolor{black}{\SI{0.2032}{\m}}}  ;
\draw[gray, dashed] (-0.1016*5, -0.6)      -- (-0.1016*5, 0)  ;
\draw[gray, dashed] ( 0.1016*5, -0.6)      -- ( 0.1016*5, 0)  ;

\draw[black, line width=1pt,<->] (-0.1016*20, 1.05)      -- (0.1016*20, 1.05) node[pos=0.5, above] {\footnotesize \bf \textcolor{black}{\SI{0.8128}{\m},~\SI{2}{\mega \hertz}}};
\draw[gray, dashed] (-0.1016*20, 1.05)      -- (-0.1016*20, 0.45)  ;
\draw[gray, dashed] ( 0.1016*20, 1.05)      -- ( 0.1016*20, 0.45)  ;

\draw[black, line width=1pt,<->] (-0.1016*40, -1.2)      -- (0.1016*40, -1.2) node[pos=0.5, below] {\footnotesize \bf \textcolor{black}{\SI{1.6256}{\m},~\SI{0.5}{\mega \hertz}}}  ;
\draw[gray, dashed] (-0.1016*40, -1.2)      -- (-0.1016*40, 0)  ;
\draw[gray, dashed] ( 0.1016*40, -1.2)      -- ( 0.1016*40, 0)  ;

\draw[black, line width=1pt,<->] (-0.1016*60, 1.65)      -- (0.1016*60, 1.65) node[pos=0.5, above] {\footnotesize \bf \textcolor{black}{\SI{2.4384}{\m},~\SI{0.25}{\mega \hertz}}};
\draw[gray, dashed] (-0.1016*60, 1.65)      -- (-0.1016*60, 0.45)  ;
\draw[gray, dashed] ( 0.1016*60, 1.65)      -- ( 0.1016*60, 0.45)  ;

\end{tikzpicture}

%% file: Figures/deep_azim_schem.tex
\begin{tikzpicture}

\fill[gray!80!white] (-1.5*5,0) -- (1.5*5,0.) --  
(1.5*5, 0.15*3) -- (-1.5*5,0.15*3) -- cycle;


\fill[black!80!white] (-0.6096*10-0.02*3,0) -- (-0.6096*10+0.02*3,0.)
node[below] {\footnotesize \bf \textcolor{black}{Tx}} -- (-0.6096*10+0.02*3, 0.15*3)   -- (-0.6096*10-0.02*3,0.15*3) -- cycle;
%
\fill[red!80!white] (-0.02*3,0) -- (0.02*3,0.)  -- (0.02*3, 0.15*3) node[above] {\footnotesize \bf \textcolor{black}{Rx\textsubscript{1}}} -- (-0.02*3,0.15*3) -- cycle;

\fill[red!80!white] (0.6096*10-0.02*3,0) -- (0.6096*10+0.02*3,0.)node[below] {\footnotesize \bf \textcolor{black}{Rx\textsubscript{2}}}  -- (0.6096*10+0.02*3, 0.15*3)  -- (0.6096*10-0.02*3,0.15*3) -- cycle;

\draw[black, line width=1pt,<->] (-0.6096*10, -0.6)      -- (0, -0.6) node[pos=0.5, below] {\footnotesize \bf \textcolor{black}{ \SI{12}{\m},~\SI{24}{\kilo \hertz}}}  ;
\draw[gray, dashed] (-0.6096*10,-0.6)      -- (-0.6096*10, 0)  ;
\draw[gray, dashed] ( 0.0,-0.6)      -- ( 0.0, 0)  ;

\draw[black, line width=1pt,<->] (-0.6096*10, 1.05)      -- (0.6096*10, 1.05) node[pos=0.5, above] {\footnotesize \bf \textcolor{black}{\SI{25}{\m},~\SI{2}{\kilo \hertz}}}  ;
\draw[gray, dashed] (-0.6096*10,1.05)      -- (-0.6096*10, 0.45)  ;
\draw[gray, dashed] ( 0.6096*10,1.05)      -- ( 0.6096*10, 0.45)  ;

\end{tikzpicture}

%% file: Figures/Syn1/Real/real_v1.tex
\pgfplotsset{every axis legend/.append style={
		at={(0.5,1.03)},
		anchor=south},
	every axis plot/.append style={line width=1.8pt},
}
\begin{tikzpicture}
\begin{axis}[
xmin=0,
xmax=540,
legend columns=2,
ymin=0.0,
 ymax=15,
height=0.35\textwidth,
width=0.93\textwidth,
 y dir=reverse,
xlabel={HD ($m$)},
ylabel near ticks,
ylabel={TVD ($m$)},
enlargelimits=false,
]



\addplot graphics[xmin=0,xmax=540,ymin=0,ymax=15] {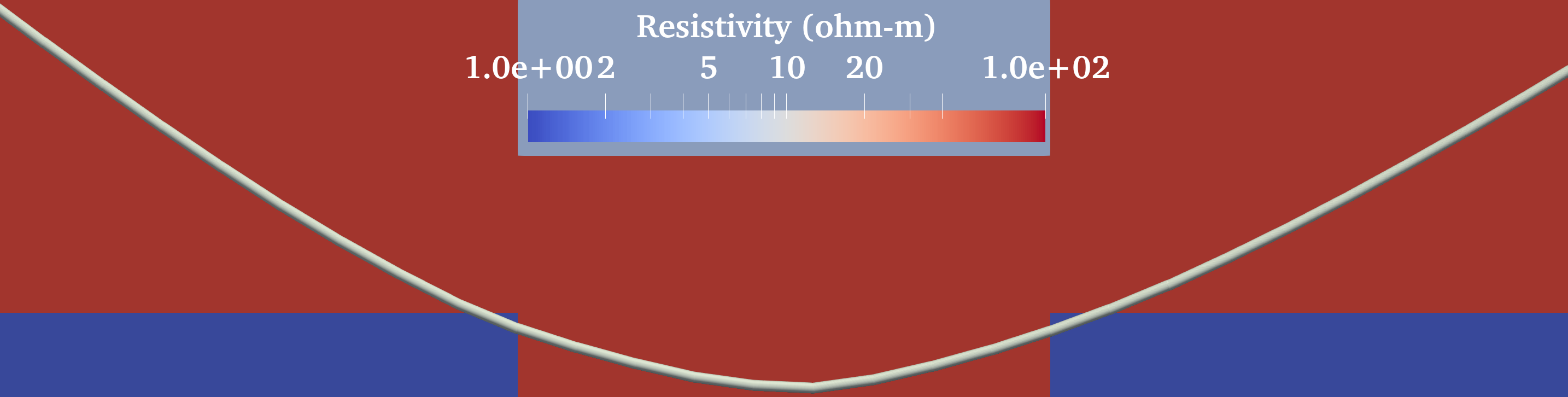};

\end{axis}	
\node[rotate=0] (I_h) at (5,-1.25) {Actual formation};
\end{tikzpicture}

%% file: Figures/Syn1/exp0/exp0_v1.tex
\pgfplotsset{every axis legend/.append style={
		at={(0.5,1.03)},
		anchor=south},
	every axis plot/.append style={line width=1.8pt},
}
\begin{tikzpicture}
\begin{axis}[
legend columns=2,
height=0.25\textwidth,
width=0.52\textwidth,
%
enlargelimits=false,
yticklabels={,,},
xticklabels={,,}
]



\addplot graphics[xmin=0,xmax=540,ymin=0,ymax=15] {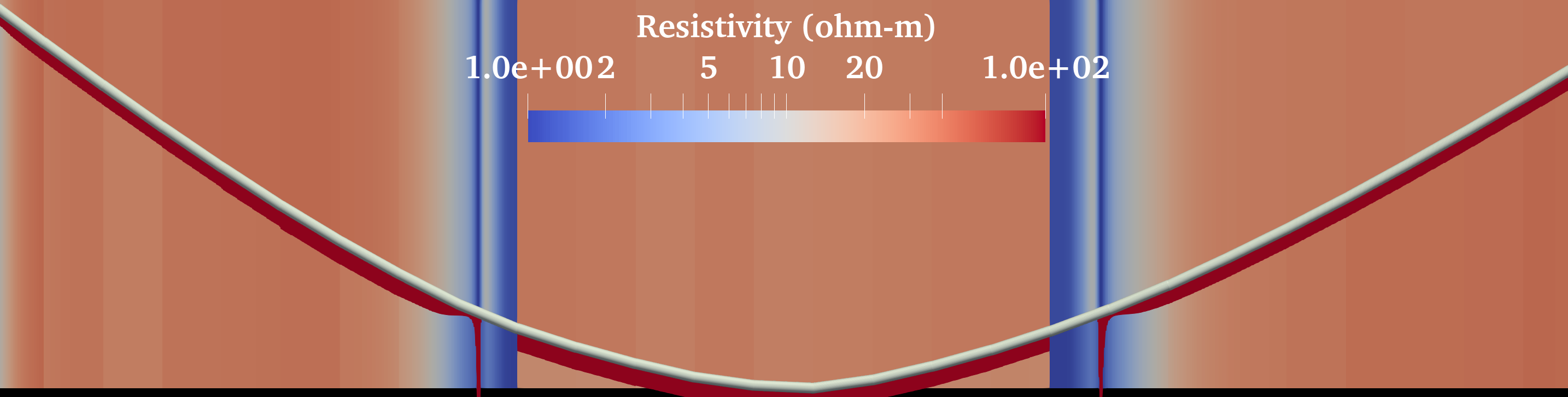};

\end{axis}	

\node[rotate=0] (I_h) at (2.6,-.50) {Iteration 1};
\end{tikzpicture}

%% file: Figures/Syn1/exp1/exp1_v1.tex
\pgfplotsset{every axis legend/.append style={
		at={(0.5,1.03)},
		anchor=south},
	every axis plot/.append style={line width=1.8pt},
}
\begin{tikzpicture}
\begin{axis}[
legend columns=2,
height=0.25\textwidth,
width=0.52\textwidth,
%
enlargelimits=false,
yticklabels={,,},
xticklabels={,,}
]



\addplot graphics[xmin=0,xmax=540,ymin=0,ymax=15] {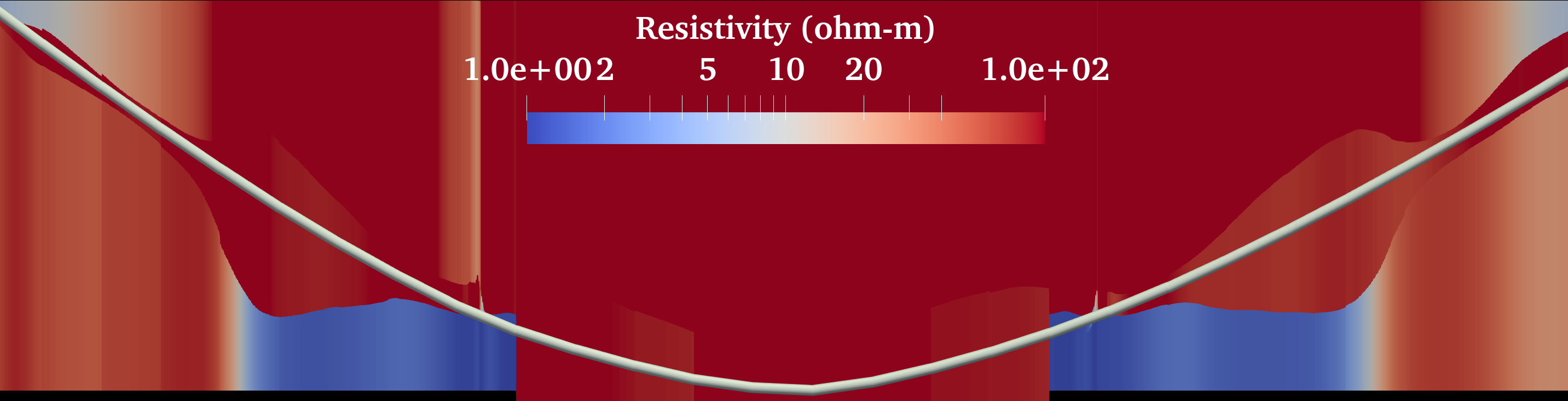};

\end{axis}	
\node[rotate=0] (I_h) at (2.6,-.50) {Iteration 2};
\end{tikzpicture}

%% file: Figures/Syn1/exp2/exp2_v1.tex
\pgfplotsset{every axis legend/.append style={
		at={(0.5,1.03)},
		anchor=south},
	every axis plot/.append style={line width=1.8pt},
}
\begin{tikzpicture}
\begin{axis}[
legend columns=2,
height=0.25\textwidth,
width=0.52\textwidth,
%
enlargelimits=false,
yticklabels={,,},
xticklabels={,,}
]



\addplot graphics[xmin=0,xmax=540,ymin=0,ymax=15] {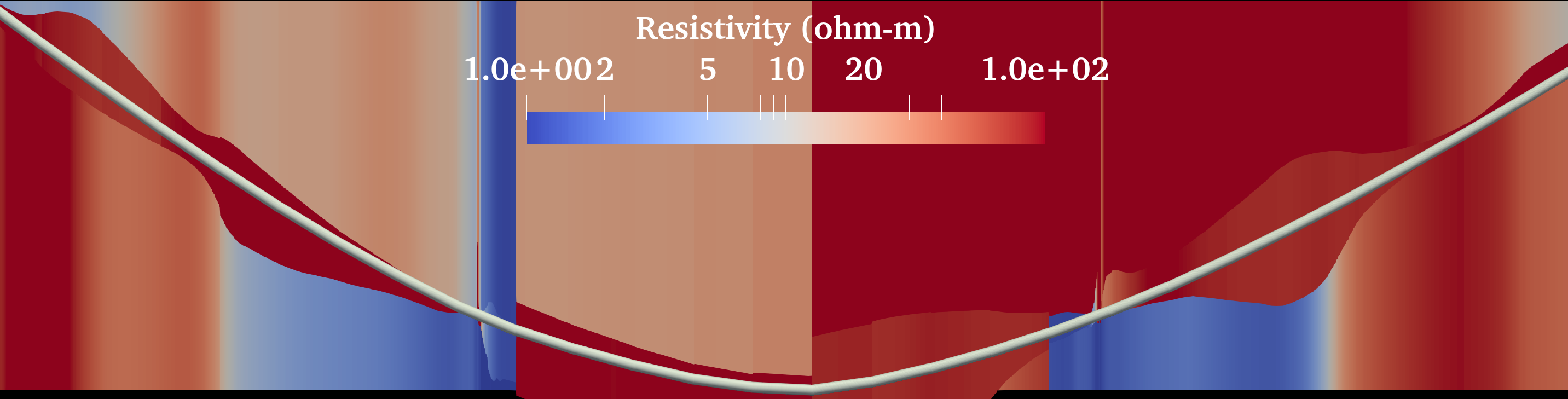};

\end{axis}	
\node[rotate=0] (I_h) at (2.6,-.50) {Iteration 3};
\end{tikzpicture}

%% file: Figures/Syn1/exp3/exp3_v1.tex
\pgfplotsset{every axis legend/.append style={
		at={(0.5,1.03)},
		anchor=south},
	every axis plot/.append style={line width=1.8pt},
}
\begin{tikzpicture}
\begin{axis}[
legend columns=2,
height=0.25\textwidth,
width=0.52\textwidth,
%
enlargelimits=false,
yticklabels={,,},
xticklabels={,,}
]



\addplot graphics[xmin=0,xmax=540,ymin=0,ymax=15] {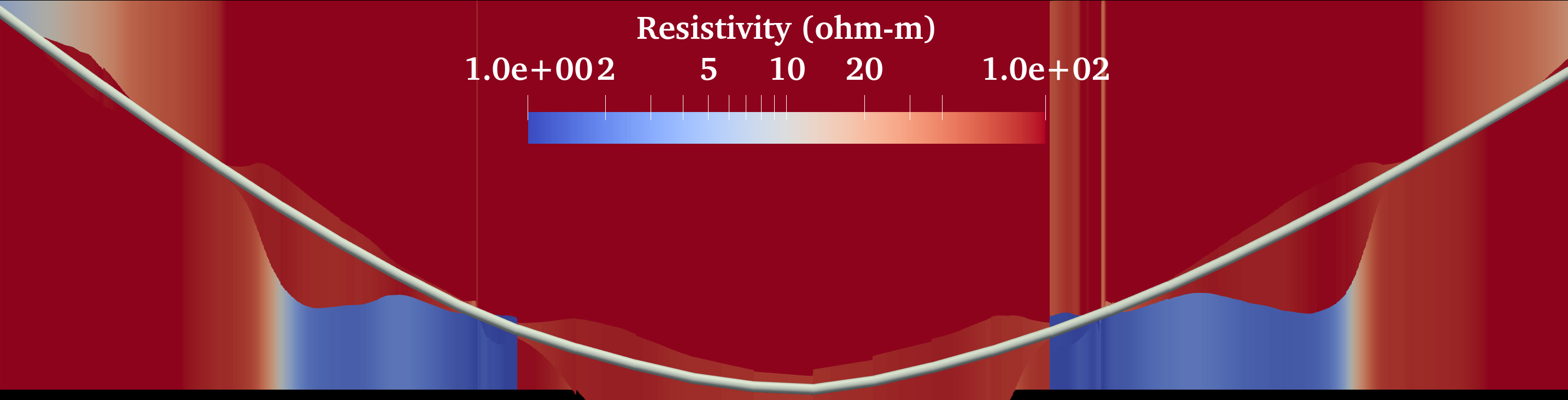};

\end{axis}	
\node[rotate=0] (I_h) at (2.6,-.50) {Iteration 4};
\end{tikzpicture}

%% file: Figures/Syn1/exp4/exp4_v1.tex
\pgfplotsset{every axis legend/.append style={
		at={(0.5,1.03)},
		anchor=south},
	every axis plot/.append style={line width=1.8pt},
}
\begin{tikzpicture}
\begin{axis}[
legend columns=2,
height=0.25\textwidth,
width=0.52\textwidth,
%
enlargelimits=false,
yticklabels={,,},
xticklabels={,,}
]



\addplot graphics[xmin=0,xmax=540,ymin=0,ymax=15] {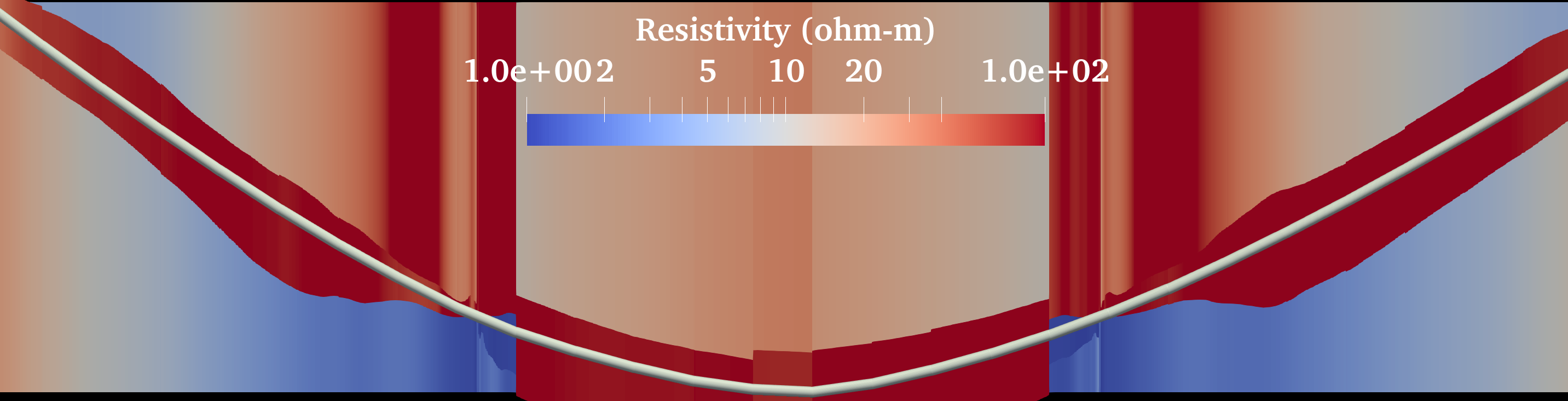};

\end{axis}	
\node[rotate=0] (I_h) at (2.6,-.50) {Iteration 5};
\end{tikzpicture}

%% file: Figures/Syn1/exp5/exp5_v1.tex
\pgfplotsset{every axis legend/.append style={
		at={(0.5,1.03)},
		anchor=south},
	every axis plot/.append style={line width=1.8pt},
}
\begin{tikzpicture}
\begin{axis}[
legend columns=2,
height=0.25\textwidth,
width=0.52\textwidth,
%
enlargelimits=false,
yticklabels={,,},
xticklabels={,,}
]



\addplot graphics[xmin=0,xmax=540,ymin=0,ymax=15] {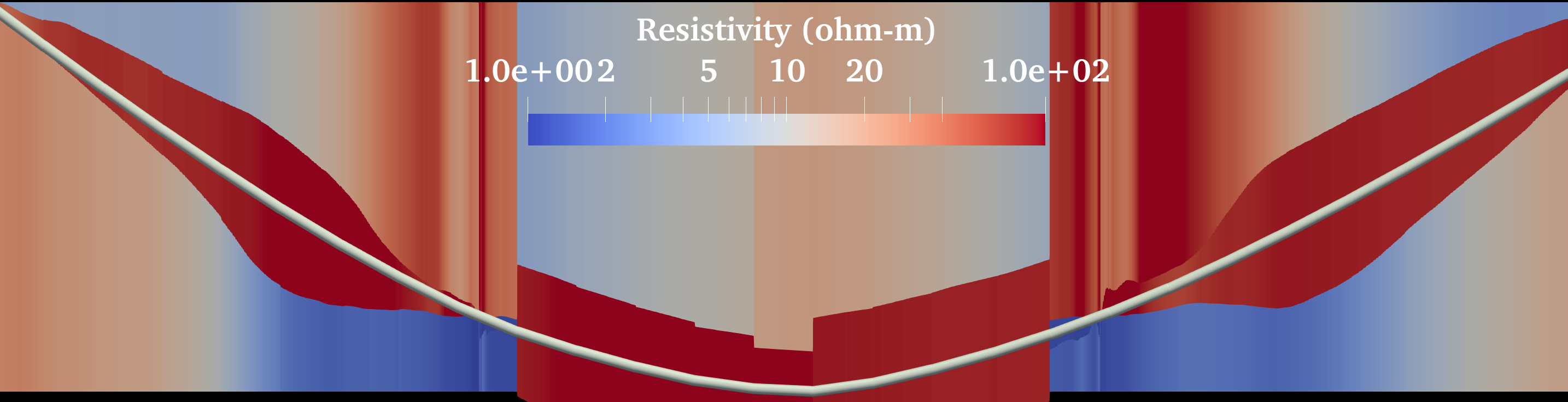};

\end{axis}	
\node[rotate=0] (I_h) at (2.6,-.50) {Iteration 6};
\end{tikzpicture}

%% file: Figures/Syn1/exp6/exp6_v1.tex
\pgfplotsset{every axis legend/.append style={
		at={(0.5,1.03)},
		anchor=south},
	every axis plot/.append style={line width=1.8pt},
}
\begin{tikzpicture}
\begin{axis}[
legend columns=2,
height=0.25\textwidth,
width=0.52\textwidth,
%
enlargelimits=false,
yticklabels={,,},
xticklabels={,,}
]



\addplot graphics[xmin=0,xmax=540,ymin=0,ymax=15] {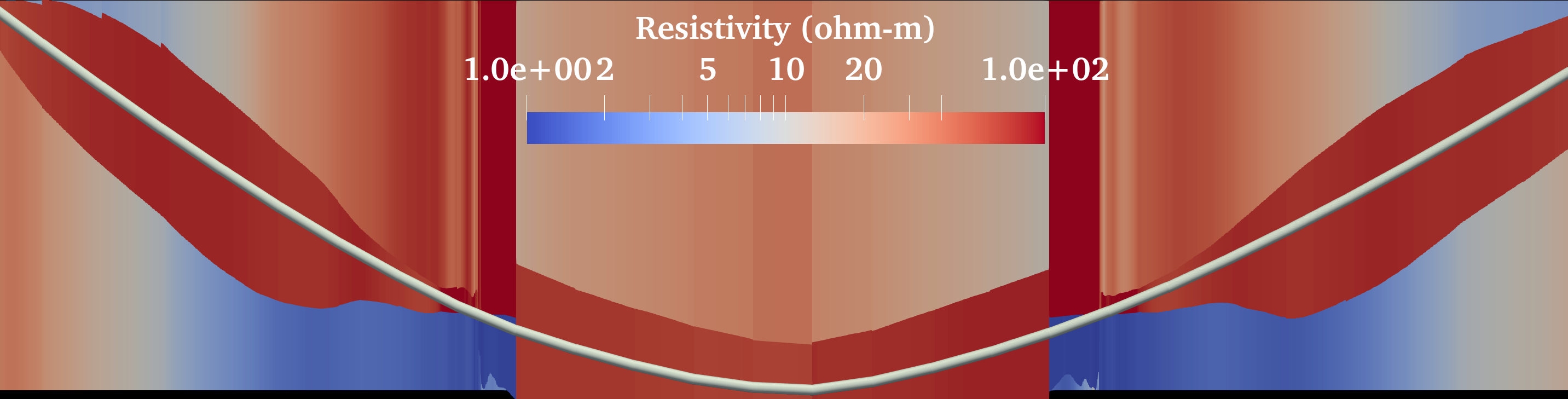};

\end{axis}	
\node[rotate=0] (I_h) at (2.6,-.50) {Iteration 7};
\end{tikzpicture}

%% file: Figures/Syn1/exp7/exp7_v1.tex
\pgfplotsset{every axis legend/.append style={
		at={(0.5,1.03)},
		anchor=south},
	every axis plot/.append style={line width=1.8pt},
}
\begin{tikzpicture}
\begin{axis}[
legend columns=2,
height=0.25\textwidth,
width=0.52\textwidth,
%
enlargelimits=false,
yticklabels={,,},
xticklabels={,,}
]



\addplot graphics[xmin=0,xmax=540,ymin=0,ymax=15] {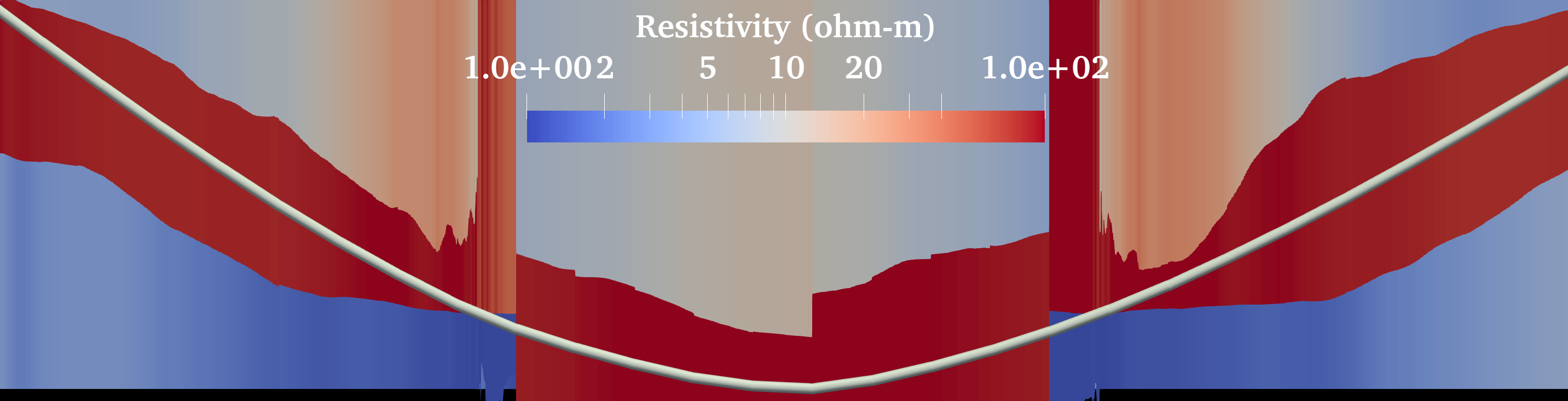};

\end{axis}	
\node[rotate=0] (I_h) at (2.6,-.50) {Iteration 8};
\end{tikzpicture}

%% file: Figures/Syn2/Real/real_v1.tex
\pgfplotsset{every axis legend/.append style={
		at={(0.5,1.03)},
		anchor=south},
	every axis plot/.append style={line width=1.8pt},
}
\begin{tikzpicture}
\begin{axis}[
xmin=0,
xmax=540,
legend columns=2,
 ymin=0.0,
ymax=27.3,
height=0.35\textwidth,
width=0.93\textwidth,
 y dir=reverse,
xlabel={HD ($m$)},
ylabel near ticks,
ylabel={TVD ($m$)},
enlargelimits=false,
extra x ticks={9.3},
extra x tick labels={$540$}
]



\addplot graphics[xmin=0,xmax=540,ymin=0,ymax=27.3] {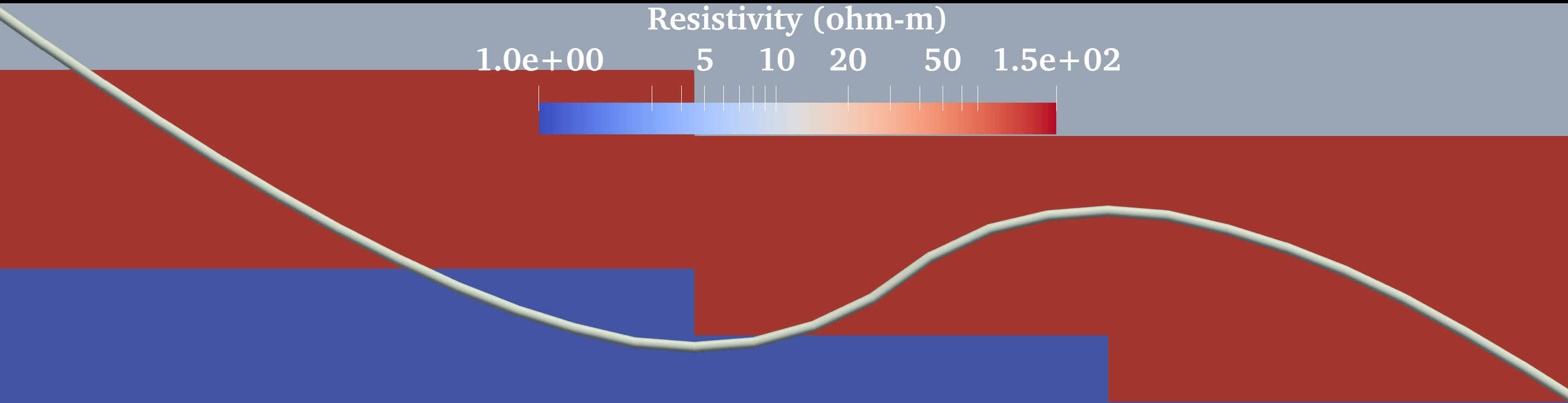};

\end{axis}	
\node[rotate=0] (I_h) at (5,-1.25) {Actual formation};
\end{tikzpicture}

%% file: Figures/Syn2/exp0/exp0_v1.tex
\pgfplotsset{every axis legend/.append style={
		at={(0.5,1.03)},
		anchor=south},
	every axis plot/.append style={line width=1.8pt},
}
\begin{tikzpicture}
\begin{axis}[
legend columns=2,
height=0.25\textwidth,
width=0.52\textwidth,
%
enlargelimits=false,
yticklabels={,,},
xticklabels={,,}
]



\addplot graphics[xmin=0,xmax=540,ymin=0,ymax=15] {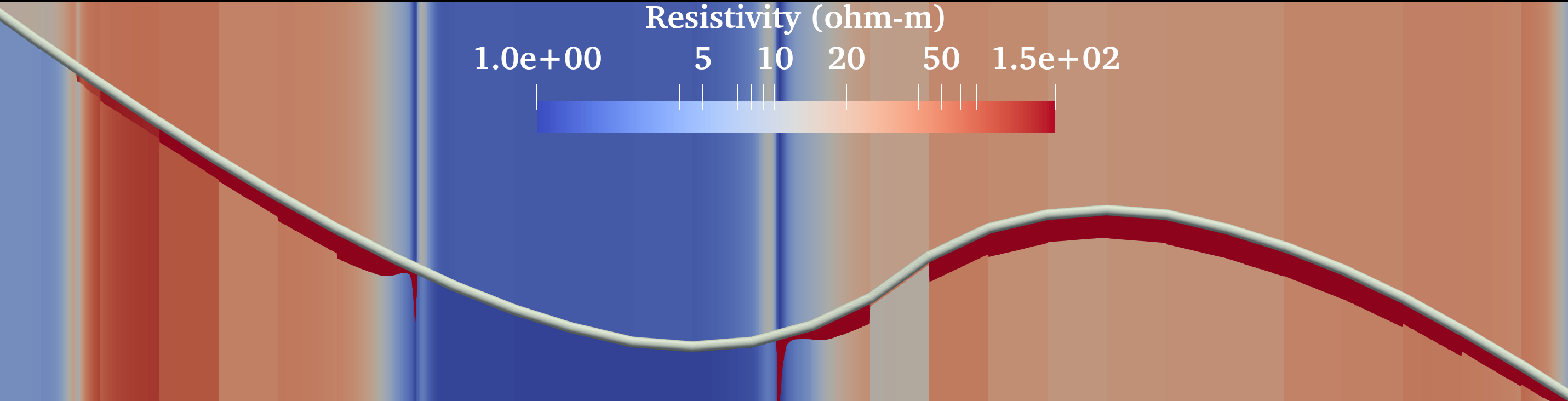};

\end{axis}	

\node[rotate=0] (I_h) at (2.6,-.50) {Iteration 1};
\end{tikzpicture}

%% file: Figures/Syn2/exp1/exp1_v1.tex
\pgfplotsset{every axis legend/.append style={
		at={(0.5,1.03)},
		anchor=south},
	every axis plot/.append style={line width=1.8pt},
}
\begin{tikzpicture}
\begin{axis}[
legend columns=2,
height=0.25\textwidth,
width=0.52\textwidth,
%
enlargelimits=false,
yticklabels={,,},
xticklabels={,,}
]



\addplot graphics[xmin=0,xmax=540,ymin=0,ymax=15] {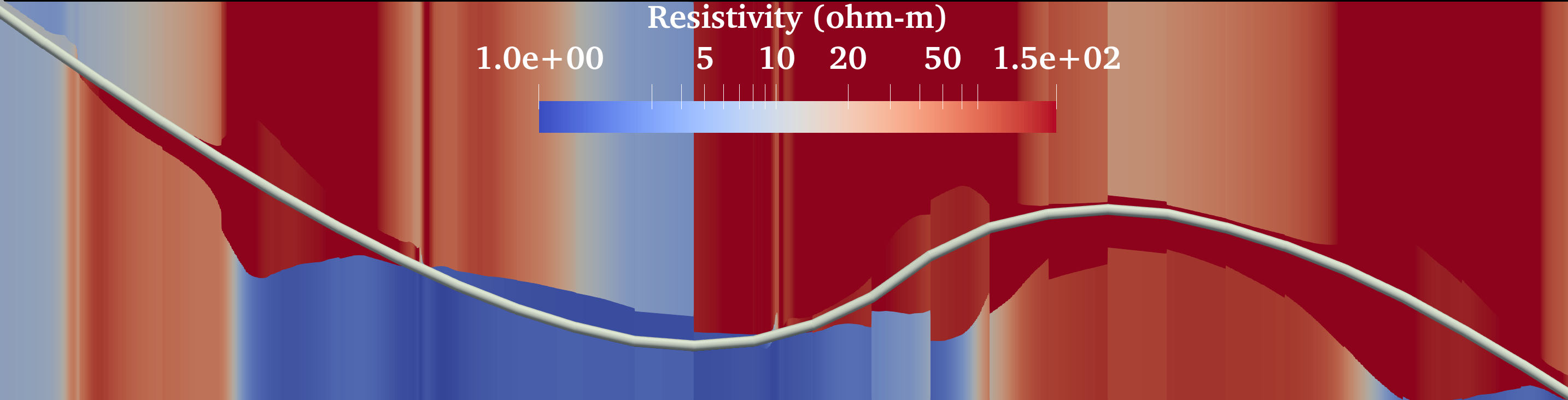};

\end{axis}	
\node[rotate=0] (I_h) at (2.6,-.50) {Iteration 2};
\end{tikzpicture}

%% file: Figures/Syn2/exp2/exp2_v1.tex
\pgfplotsset{every axis legend/.append style={
		at={(0.5,1.03)},
		anchor=south},
	every axis plot/.append style={line width=1.8pt},
}
\begin{tikzpicture}
\begin{axis}[
legend columns=2,
height=0.25\textwidth,
width=0.52\textwidth,
%
enlargelimits=false,
yticklabels={,,},
xticklabels={,,}
]



\addplot graphics[xmin=0,xmax=540,ymin=0,ymax=15] {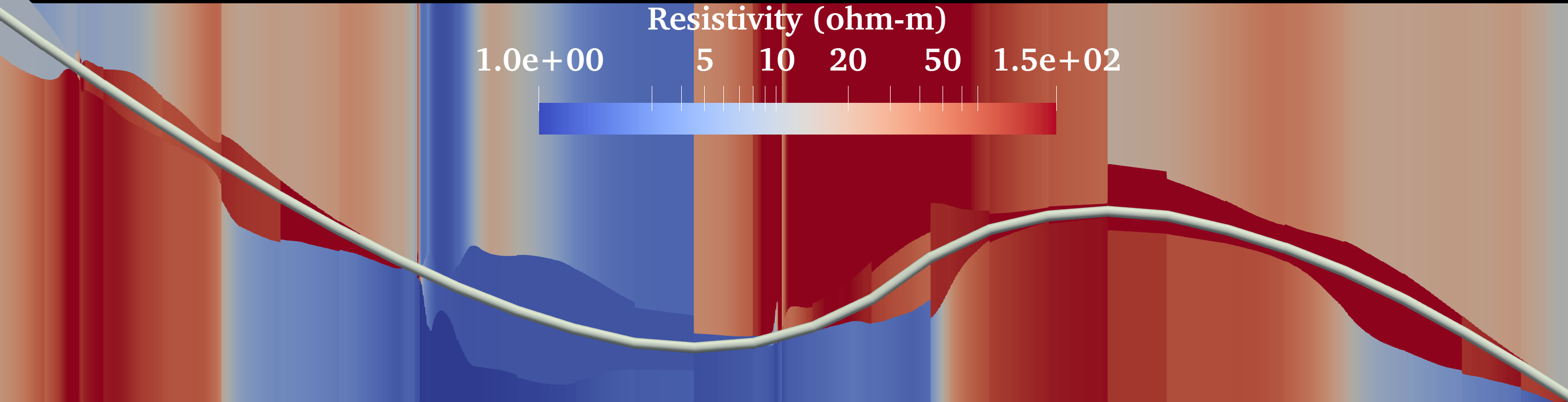};

\end{axis}	
\node[rotate=0] (I_h) at (2.6,-.50) {Iteration 3};
\end{tikzpicture}

%% file: Figures/Syn2/exp3/exp3_v1.tex
\pgfplotsset{every axis legend/.append style={
		at={(0.5,1.03)},
		anchor=south},
	every axis plot/.append style={line width=1.8pt},
}
\begin{tikzpicture}
\begin{axis}[
legend columns=2,
height=0.25\textwidth,
width=0.52\textwidth,
%
enlargelimits=false,
yticklabels={,,},
xticklabels={,,}
]



\addplot graphics[xmin=0,xmax=540,ymin=0,ymax=15] {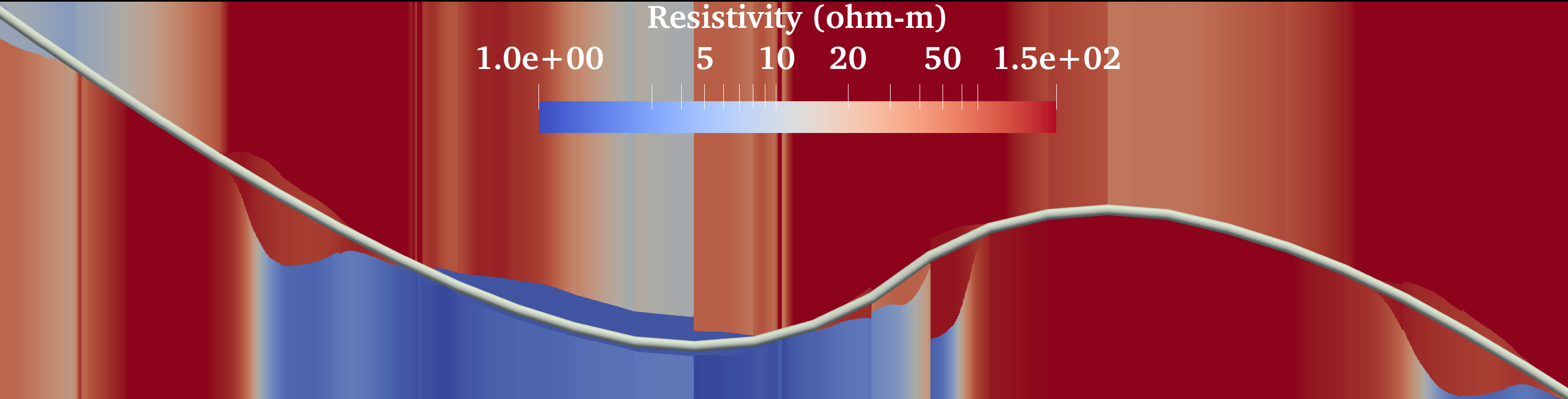};

\end{axis}	
\node[rotate=0] (I_h) at (2.6,-.50) {Iteration 4};
\end{tikzpicture}

%% file: Figures/Syn2/exp4/exp4_v1.tex
\pgfplotsset{every axis legend/.append style={
		at={(0.5,1.03)},
		anchor=south},
	every axis plot/.append style={line width=1.8pt},
}
\begin{tikzpicture}
\begin{axis}[
legend columns=2,
height=0.25\textwidth,
width=0.52\textwidth,
%
enlargelimits=false,
yticklabels={,,},
xticklabels={,,}
]



\addplot graphics[xmin=0,xmax=540,ymin=0,ymax=15] {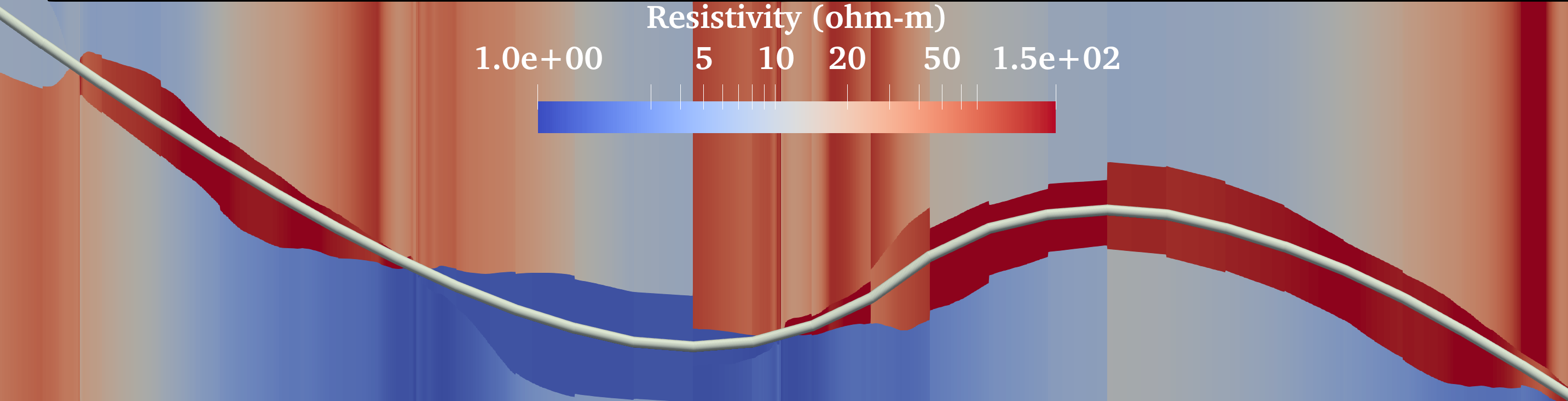};

\end{axis}	
\node[rotate=0] (I_h) at (2.6,-.50) {Iteration 5};
\end{tikzpicture}

%% file: Figures/Syn2/exp6/exp6_v1.tex
\pgfplotsset{every axis legend/.append style={
		at={(0.5,1.03)},
		anchor=south},
	every axis plot/.append style={line width=1.8pt},
}
\begin{tikzpicture}
\begin{axis}[
legend columns=2,
height=0.25\textwidth,
width=0.52\textwidth,
%
enlargelimits=false,
yticklabels={,,},
xticklabels={,,}
]



\addplot graphics[xmin=0,xmax=540,ymin=0,ymax=15] {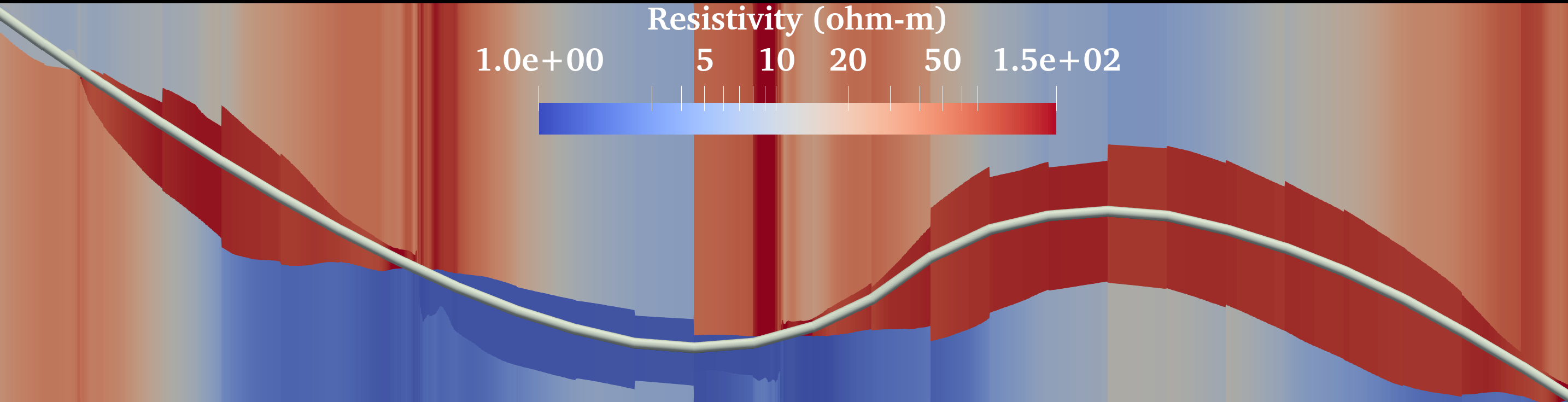};

\end{axis}	
\node[rotate=0] (I_h) at (2.6,-.50) {Iteration 7};
\end{tikzpicture}

%% file: Figures/Syn2/exp7/exp7_v1.tex
\pgfplotsset{every axis legend/.append style={
		at={(0.5,1.03)},
		anchor=south},
	every axis plot/.append style={line width=1.8pt},
}
\begin{tikzpicture}
\begin{axis}[
legend columns=2,
height=0.25\textwidth,
width=0.52\textwidth,
%
enlargelimits=false,
yticklabels={,,},
xticklabels={,,}
]



\addplot graphics[xmin=0,xmax=540,ymin=0,ymax=15] {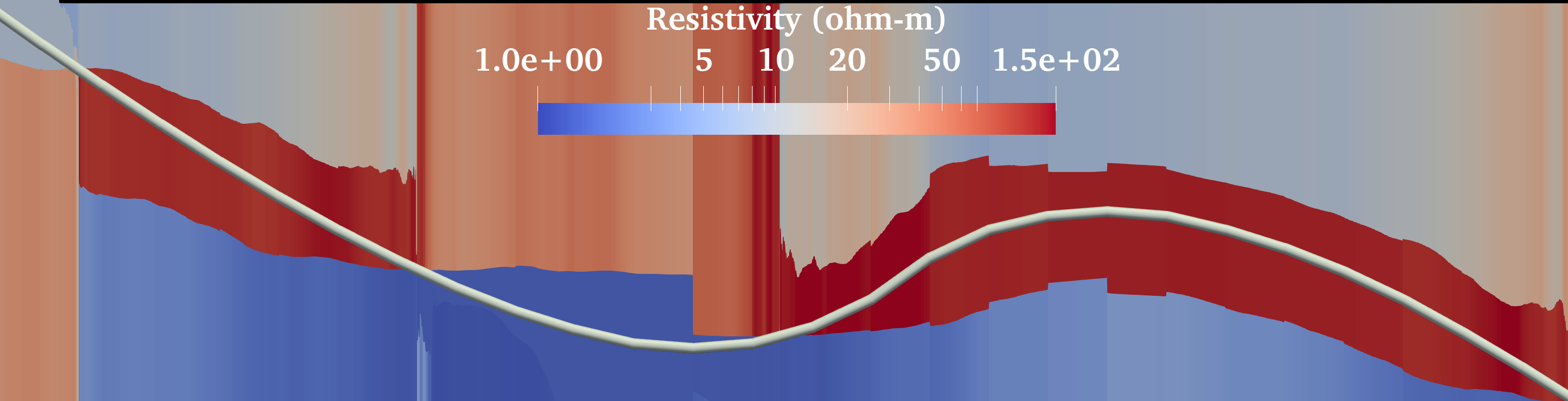};

\end{axis}	
\node[rotate=0] (I_h) at (2.6,-.50) {Iteration 8};
\end{tikzpicture}